\documentclass{report}
\usepackage[final,nonatbib]{nips_2018}
\usepackage{natbib}
\usepackage[utf8]{inputenc}
\usepackage{paralist}
\usepackage{hyperref}
\usepackage[dvipsnames]{xcolor}
\usepackage{url}
\usepackage{graphicx}
\graphicspath{{fig/}}
\title{Neural Approaches to Conversational AI
\\[0.2cm]
\large Question Answering, Task-Oriented Dialogues and Social Chatbots}

\author{
{\bf Jianfeng Gao}\\
Microsoft Research\\
{\tt jfgao@microsoft.com}
\and
{\bf Michel Galley}\\
Microsoft Research\\
{\tt mgalley@microsoft.com}
\and
{\bf Lihong Li}\\
Google Brain\\
{\tt lihong@google.com}
}

\usepackage{mwe}
\usepackage{amsmath}
\usepackage{amssymb}
\usepackage{wasysym}
\usepackage{bbm}

\begin{document}

\maketitle

\vspace{4cm}
\begin{abstract}
The present paper surveys neural approaches to conversational AI that have been developed in the last few years. We group conversational systems into three categories: (1) question answering agents, (2) task-oriented dialogue agents, and (3) chatbots. For each category, we present a review of state-of-the-art neural approaches, draw the connection between them and traditional 
approaches, and discuss the progress that has been made and challenges still being faced, using specific systems and models as case studies.\footnote{We are grateful to the anonymous reviewers,
Chris Brockett,
Asli Celikyilmaz,
Yu Cheng,
Bill Dolan,
Pascale Fung,
Zhe Gan,
Sungjin Lee,
Jinchao Li,
Xiujun Li,
Bing Liu,
Andrea Madotto,
Rangan Majumder,
Alexandros Papangelis,
Olivier Pietquin,
Chris Quirk,
Alan Ritter,
Paul Smolensky,
Alessandro Sordoni,
Yang Song,
Hisami Suzuki, 
Wei Wei,
Tal Weiss,
Kun Yuan,
and
Yizhe Zhang
for their helpful comments and suggestions on earlier versions of this paper.}
\end{abstract}

\newcommand{\todo}[1]{{\color{red}{[{\bf TODO}: #1]}}}
\newcommand{\todoJG}[1]{{\color{blue}{[{\bf JG}: #1]}}}
\newcommand{\todoMG}[1]{{\color{purple}{[{\bf MG}: #1]}}}
\newcommand{\todoLL}[1]{{\color{cyan}{[{\bf LL}: #1]}}}

\newcommand{\figref}[1]{Fig.~\ref{#1}}
\newcommand{\eqnref}[1]{Eqn.~\ref{#1}}
\newcommand{\chref}[1]{Chapter~\ref{#1}}
\newcommand{\secref}[1]{Sec.~\ref{#1}}
\newcommand{\tabref}[1]{Table~\ref{#1}}
\newcommand{\ie}{{i.e.}}
\newcommand{\eg}{{e.g.}}
\newcommand{\etc}{{etc.}}
\newcommand{\cf}{{cf.}}

\newcommand{\dnfont}[1]{{\texttt{#1}}}  
\newcommand{\dafont}[1]{{\texttt{#1}}}  
\newcommand{\slotfont}[1]{\texttt{#1}}  
\newcommand{\valuefont}[1]{\textcolor{gray}{\texttt{#1}}}
\newcommand{\vecb}[1]{\mathbf{#1}}

\newcommand{\exbox}[1]{ 
{\begin{center}\fbox{%
    \begin{minipage}{.95\textwidth} 
      \centering #1
    \end{minipage}%
  }\end{center}} 
}

\newcommand{\defeq}{:=}
\newcommand{\E}{\mathbb{E}}
\newcommand{\Rset}{\mathbb{R}}
\newcommand{\mt}{{\operatorname{T}}}  			
\newcommand{\mi}{{-1}}  											
\newcommand{\argmin}{\operatorname{argmin}}
\newcommand{\argmax}{\operatorname{argmax}}

\newcommand{\Sset}{\mathcal{S}}
\newcommand{\Aset}{\mathcal{A}}
\newcommand{\Dset}{\mathcal{D}}

\newcommand{\mMSE}{\operatorname{MSE}}
\newcommand{\mFone}{\operatorname{F1}}
\newcommand{\mREC}{\operatorname{RECALL}}
\newcommand{\mACC}{\operatorname{ACCURACY}}
\newcommand{\mPRE}{\operatorname{PRECISION}}
\newcommand{\mAUC}{\operatorname{AUC}}
\newcommand{\1}{\mathbf{1}}

\tableofcontents
\newpage

\chapter{Introduction}
\label{sec:intro} 

Developing an intelligent dialogue system\footnote{``Dialogue systems'' and ``conversational AI'' are often used interchangeably in the scientific literature. The difference is reflective of different traditions. The former term is more general in that a dialogue system might be purely rule-based rather than AI-based.} that not only emulates human conversation, but also answers questions on topics ranging from latest news about a movie star to Einstein's theory of relativity, and fulfills complex tasks such as travel planning, has been one of the longest running goals in AI. The goal has remained elusive until recently. We are now observing promising results both in academia sindustry, as large amounts of conversational data become available for training, and the breakthroughs in deep learning (DL) and reinforcement learning (RL) are applied to conversational AI.  

Conversational AI is fundamental to natural user interfaces. It is a rapidly growing field, attracting many researchers in the Natural Language Processing (NLP), Information Retrieval (IR) and Machine Learning (ML) communities. For example, SIGIR 2018 has created a new track of Artificial Intelligence, Semantics, and Dialog to bridge research in AI and IR, especially targeting Question Answering (QA), deep semantics and dialogue with intelligent agents. 

Recent years have seen the rise of a small industry of tutorials and survey papers on deep learning and dialogue systems. \citet{yih2015deep,yih2016deep,Gao2017introduction} reviewed deep learning approaches for a wide range of IR and NLP tasks, including dialogues. \citet{chen17deep} presented a tutorial on dialogues, with a focus on task-oriented agents.  \citeauthor{serban2015survey}~(\citeyear{serban2015survey}; \citeyear{serban18survey}) surveyed public dialogue datasets that can be used to develop conversational agents. \citet{chen2017survey} reviewed popular deep neural network models for dialogues, focusing on supervised learning approaches. 
The present work substantially expands the scope of \citet{chen2017survey,serban2015survey} by going beyond data and supervised learning to provide what we believe is the first survey of neural approaches to conversational AI, targeting NLP and IR audiences.\footnote{One important topic of conversational AI that we do not cover is Spoken Language Understanding (SLU). SLU systems are designed to extract the meaning from speech utterances and their application are vast, ranging from voice search in mobile devices to meeting summarization. The present work does encompass many Spoken Dialogue Systems -- for example \citet{young2013pomdp} -- but does not focus on components related to speech. 
We refer readers to \citet{tur2011spoken} for a survey of SLU.} 
Its contributions are:
\begin{itemize}
\item We provide a comprehensive survey of the neural approaches to conversational AI that have been developed in the last few years, covering QA, task-oriented and social bots with a unified view of optimal decision making. 
\item We draw connections between modern neural approaches and traditional 
approaches, allowing us to better understand why and how the research has evolved and to shed light on how we can move forward. 
\item We present state-of-the-art approaches to training dialogue agents using both supervised and reinforcement learning.
\item We sketch out the landscape of conversational systems developed in the research community and released in industry, demonstrating via case studies the progress that has been made and the challenges that we are still facing.
\end{itemize}

\section{Who Should Read this Paper?}
This paper is based on tutorials given at the SIGIR and ACL conferences in 2018 \citep{gao2018neural,gao2018neural-acl}, with the IR and NLP communities as the primary target audience.  However, audiences with other backgrounds (such as machine learning) will also find it an accessible introduction to conversational AI with numerous pointers, especially to recently developed neural approaches.

We hope that this paper will prove a valuable resource for students, researchers, and software developers. It provides a unified view, as well as a detailed presentation of the important ideas and insights needed to understand and create modern dialogue agents that will be instrumental to making world knowledge and services accessible to millions of users in ways that seem natural and intuitive.

This survey is structured as follows:
\begin{itemize}
\item The rest of this chapter introduces dialogue tasks and presents a unified view in which open-domain dialogue is formulated as an optimal decision making process. 
\item \chref{sec:neural-background} introduces basic mathematical tools and machine learning concepts, and reviews recent progress in the deep learning and reinforcement learning techniques that are fundamental to developing neural dialogue agents.
\item \chref{sec:qa-bot} describes question answering (QA) agents, focusing on neural models for knowledge-base QA and machine reading comprehension (MRC).
\item \chref{sec:dialogue} describes task-oriented dialogue agents, focusing on applying deep reinforcement learning to dialogue management.
\item \chref{sec:chitchat} describes social chatbots, focusing on fully data-driven neural approaches to end-to-end generation of conversational responses.
\item \chref{sec:commercial} gives a brief review of several conversational systems in industry.
\item \chref{sec:conclusion} concludes the paper with a discussion of research trends.
\end{itemize}



\section{Dialogue: What Kinds of Problems?}


\begin{figure}[t] 
\centering 
\includegraphics[width=1.0\linewidth]{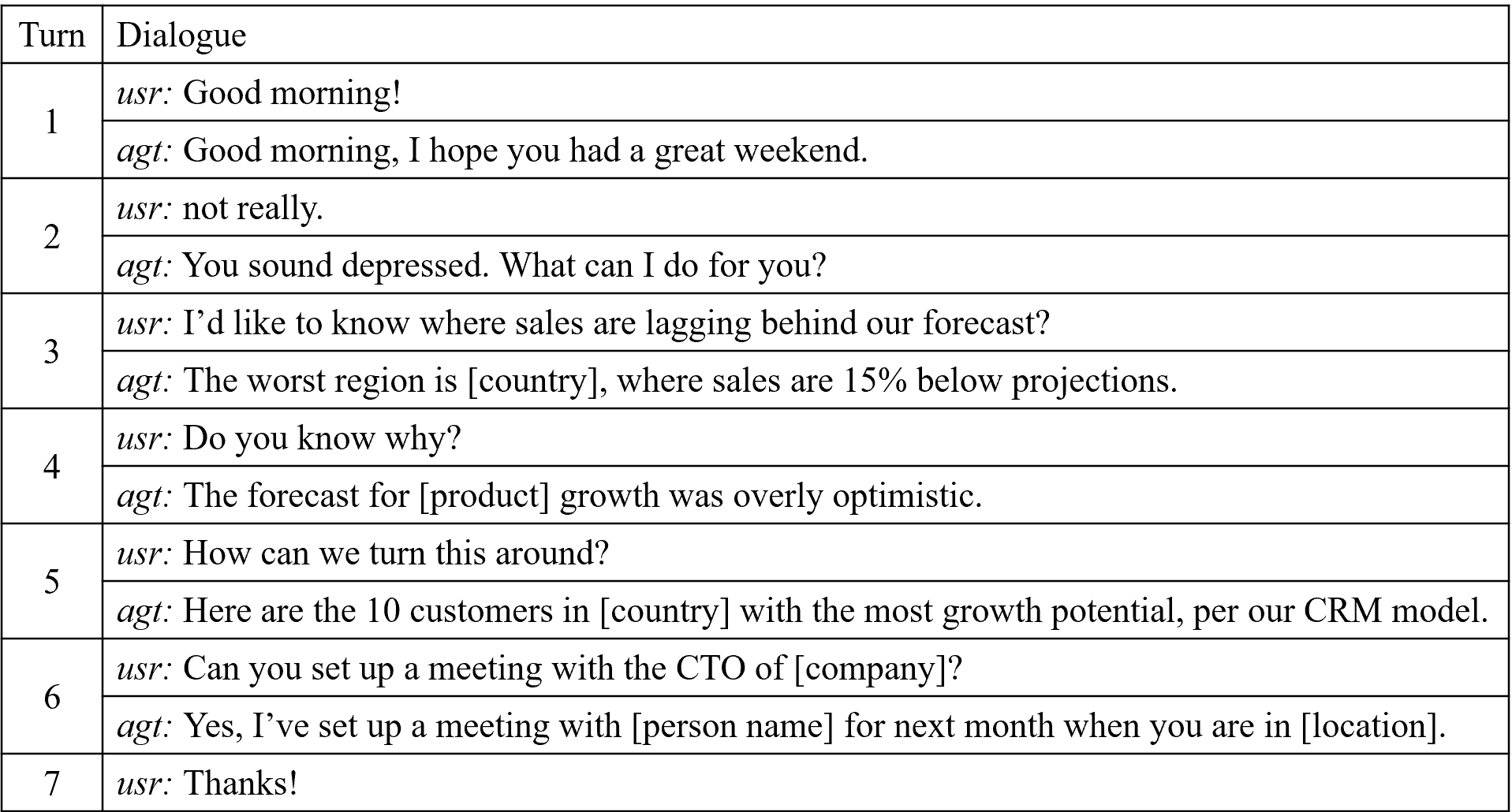}
\vspace{-2mm}
\caption{A human-agent dialogue during the process of making a business decision. (\textit{usr}: user, \textit{agt}: agent) The dialogue consists of multiple segments of different types. Turns 1 and 2 are a social chat segment. Turns 3 to 5 are a QA segment. Turns 6 and 7 are a task-completion segment.} 
\label{fig:sample_dialogue} 
\vspace{0mm}
\end{figure}

\figref{fig:sample_dialogue} shows a human-agent dialogue during the process of making a business decision. The example illustrates the kinds of problems a dialogue system is expected to solve:

\begin{itemize}
\item \textbf{question answering}: the agent needs to provide concise, direct answers to user queries based on rich knowledge drawn from various data sources including text collections such as Web documents and pre-compiled knowledge bases such as sales and marketing datasets, as the example shown in Turns 3 to 5 in \figref{fig:sample_dialogue}.
\item \textbf{task completion}: the agent needs to accomplish user tasks ranging from restaurant reservation to meeting scheduling (e.g., Turns 6 to 7 in \figref{fig:sample_dialogue}), and to business trip planning.
\item \textbf{social chat}: the agent needs to converse seamlessly and appropriately with users --- like a human as in the Turing test --- and provide useful recommendations (\eg, Turns 1 to 2 in \figref{fig:sample_dialogue}).
\end{itemize}

One may envision that the above dialogue can be collectively accomplished by a set of agents, also known as \emph{bots}, each of which is designed for solving a particular type of task, \eg, QA bots, task-completion bots, social chatbots. These bots can be grouped into two categories, \emph{task-oriented} and \emph{chitchat}, depending on whether the dialogue is conducted to assist users to achieve specific tasks, \eg, obtain an answer to a query or have a meeting scheduled.

Most of the popular personal assistants in today’s market, such as Amazon Alexa, Apple Siri, Google Home, and Microsoft Cortana, are task-oriented bots. These can only handle relatively simple tasks, such as reporting weather and requesting songs. An example of a chitchat dialogue bot is Microsoft XiaoIce. Building a dialogue agent to fulfill complex tasks as in \figref{fig:sample_dialogue} remains one of the most fundamental challenges for the IR and NLP communities, and AI in general.

\begin{figure}[t] 
\centering 
\includegraphics[width=1.0\linewidth]{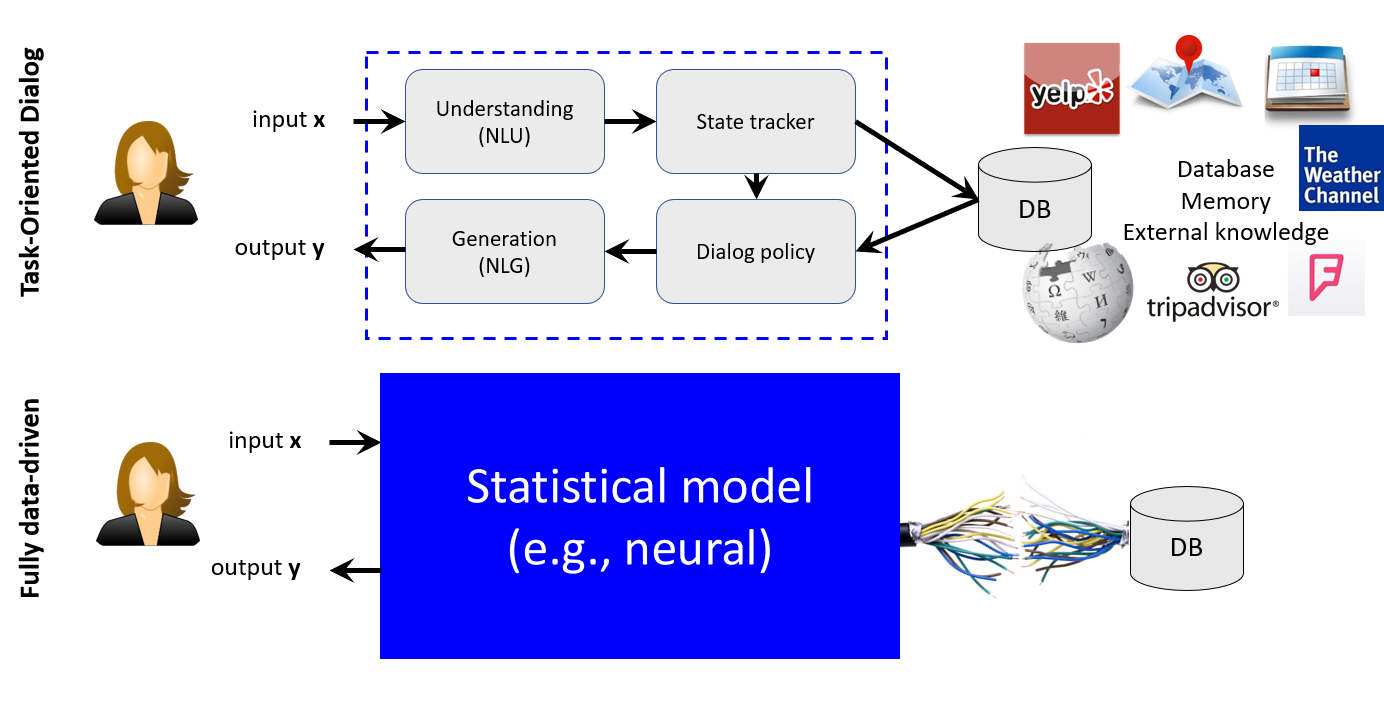}
\vspace{-2mm}
\caption{Two architectures of dialogue systems for (Top) traditional task-oriented dialogue and (Bottom) fully data-driven dialogue. } 
\label{fig:two-dialogue-system} 
\vspace{0mm}
\end{figure}

A typical task-oriented dialogue agent is composed of four modules, as illustrated in  \figref{fig:two-dialogue-system} (Top): (1) a Natural Language Understanding (NLU) module for identifying user intents and extracting associated information; (2) a state tracker for tracking the dialogue state that captures all essential information in the conversation so far; (3) a dialogue policy that selects the next action based on the current state; and (4) a Natural Language Generation (NLG) module for converting agent actions to natural language responses. In recent years there has been a trend towards developing fully data-driven systems by unifying these modules using a deep neural network that maps the user input to the agent output directly, as shown in \figref{fig:two-dialogue-system} (Bottom). 

Most task-oriented bots are implemented using a modular system, where the bot often has access to an external database on which to inquire about information to accomplish the task \citep{young2013pomdp,tur2011spoken}. Social chatbots, on the other hand, are often implemented using a unitary  (non-modular) system. 
Since the primary goal of social chatbots is to be AI companions to humans with an emotional connection rather than completing specific tasks, they are often developed to mimic human conversations by training DNN-based response generation models on large amounts of human-human conversational data \citep{ritter2011data,sordoni2015neural,vinyals2015neural,shang2015neural}. Only recently have researchers begun to explore how to ground the chitchat in world knowledge \citep{ghazvininejad2017knowledge} and images \citep{mostafazadeh2017image} so as to make the conversation more contentful and interesting.

\section{A Unified View: Dialogue as Optimal Decision Making}
The example dialogue in \figref{fig:sample_dialogue} can be formulated as a decision making process. It has a natural hierarchy: a top-level process selects what agent to activate for a particular subtask (\eg, answering a question, scheduling a meeting, providing a recommendation or just having a casual chat), and a low-level process, controlled by the selected agent, chooses primitive actions to complete the subtask.

Such hierarchical decision making processes can be cast in the mathematical framework of \emph{options} over Markov Decision Processes (MDPs)~\citep{sutton99between}, where options generalize primitive actions to higher-level actions. In a traditional MDP setting, an agent chooses a primitive action at each time step.  With options, the agent can choose a ``multi-step'' action which for example could be a sequence of primitive actions for completing a subtask. 

If we view each option as an action, both top- and low-level processes can be naturally captured by the reinforcement learning framework. The dialogue agent navigates in a MDP, interacting with its environment over a sequence of discrete steps. At each step, the agent observes the current state, and chooses an action according to a policy. The agent then receives a reward and observes a new state, continuing the cycle until the episode terminates. The goal of dialogue learning is to find optimal policies to maximize expected rewards.  \tabref{rl-dialogue} formulates an sample of dialogue agents using this unified view of RL, where the state-action spaces characterize the complexity of the problems, and the rewards are the objective functions to be optimized. 

The unified view of hierarchical MDPs has already been applied to guide the development of some large-scale open-domain dialogue systems. Recent examples include Sounding Board \footnote{\url{https://sounding-board.github.io/}}, a social chatbot that won the 2017 Amazon Alexa Prize, and Microsoft XiaoIce \footnote{\url{https://www.msxiaobing.com/}}, arguably the most popular social chatbot that has attracted more than 660 million users worldwide since its release in 2014. Both systems use a hierarchical dialogue manager: a master (top-level) that manages the overall conversation process, and a collection of skills (low-level) that handle different types of conversation segments (subtasks).

The reward functions in \tabref{rl-dialogue}, which seem contradictory in CPS (e.g., we need to minimize CPS for efficient task completion but maximize CPS for improving user engagement), suggest that we have to balance the long-term and short-term gains when developing a dialogue system. 
For example, XiaoIce is a social chatbot optimized for user engagement, but is also equipped with more than 230 skills, most of which are QA and task-oriented. XiaoIce is optimized for \emph{expected} CPS which corresponds a long-term, rather than a short-term, engagement. Although incorporating many task-oriented and QA skills can reduce CPS in the short term since these skills help users accomplish tasks \emph{more efficiently} by minimizing CPS, these new skills establish XiaoIce as an efficient and trustworthy personal assistant, thus strengthening the emotional bond with human users in the long run.


Although RL provides a unified ML framework for building dialogue agents, applying RL requires training the agents by interacting with real users, which can be expensive in many domains. Hence, in practice, we often use a hybrid approach that combines the strengths of different ML methods. For example, we might use imitation and/or supervised learning methods (if there is a large amount of human-human conversational corpus) to obtain a reasonably good agent before applying RL to continue improving it.
In the paper, we will survey these ML approaches and their use for training dialogue systems.

\begin{table}[t]
\footnotesize
\centering
\caption{Reinforcement Learning for Dialogue. CPS stands for Conversation-turns Per Session, and is defined as the average number of conversation-turns between the bot and the user in a conversational session.}
\label{rl-dialogue}
\begin{tabular}{|l|l|l|l|}
\hline
\textbf{dialogue} & \textbf{state} & \textbf{action} & \textbf {reward} \\ \hline
QA & \begin{tabular}[c]{@{}l@{}}understanding of \\ user query intent\end{tabular} & \begin{tabular}[c]{@{}l@{}}clarification\\ questions \\ or answers\end{tabular} & \begin{tabular}[c]{@{}l@{}} relevance of answer,\\ (min) CPS\end{tabular} \\ \hline
task-oriented 
& \begin{tabular}[c]{@{}l@{}}understanding of \\ user goal\end{tabular}           
& \begin{tabular}[c]{@{}l@{}}dialogue-act and \\ slot/value\end{tabular} 
& \begin{tabular}[c]{@{}l@{}}task success rate,\\ (min) CPS\end{tabular} \\ \hline
chitchat                  
& \begin{tabular}[c]{@{}l@{}}conversation history\\ and user intent\end{tabular}   
& responses 
& \begin{tabular}[c]{@{}l@{}}user engagement,\\ measured in CPS\end{tabular} \\ \hline
top-level bot
& \begin{tabular}[c]{@{}l@{}}understanding of \\ user top-level intent\end{tabular} 
& options 
& \begin{tabular}[c]{@{}l@{}}user engagement,\\ measured in CPS\end{tabular} \\ \hline
\end{tabular}
\end{table}

\section{The Transition of NLP to Neural Approaches}

\begin{figure}[t]
\centering 
\includegraphics[width=0.96\linewidth]{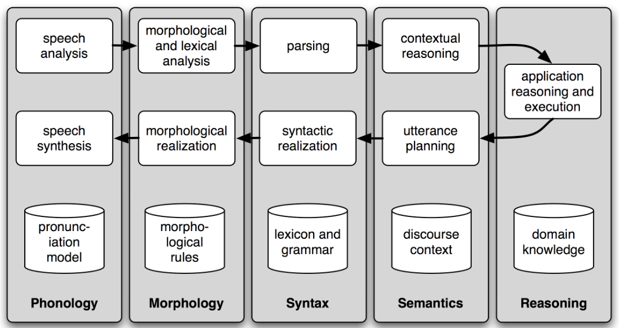}
\vspace{0mm}
\caption{Traditional NLP Component Stack. Figure credit: \citet{pythonbook}.} 
\label{fig:traditional-nlp-component-stack} 
\vspace{0mm}
\end{figure}

Neural approaches are now transforming the field of NLP and IR, where symbolic  approaches have been dominating for decades.

NLP applications differ from other data processing systems in their use of language knowledge of various levels, including phonology, morphology, syntax, semantics and discourse \citep{jurafsky2009speech}.  Historically, much of the NLP field has organized itself around the architecture of \figref{fig:traditional-nlp-component-stack}, with  researchers aligning their work with one component task, such as morphological analysis or parsing.
These tasks can be viewed as resolving (or realizing) natural language ambiguity (or diversity) at different levels by mapping (or generating) a natural language sentence to (or from) a series of human-defined, unambiguous, symbolic representations, such as Part-Of-Speech (POS) tags, context free grammar, first-order predicate calculus. With the rise of data-driven and statistical approaches, these components have remained and have been adapted as a rich source of engineered features to be fed into a variety of machine learning models \citep{manning2014stanford}.

\begin{figure}[t]
\centering 
\includegraphics[width=1.0\linewidth]{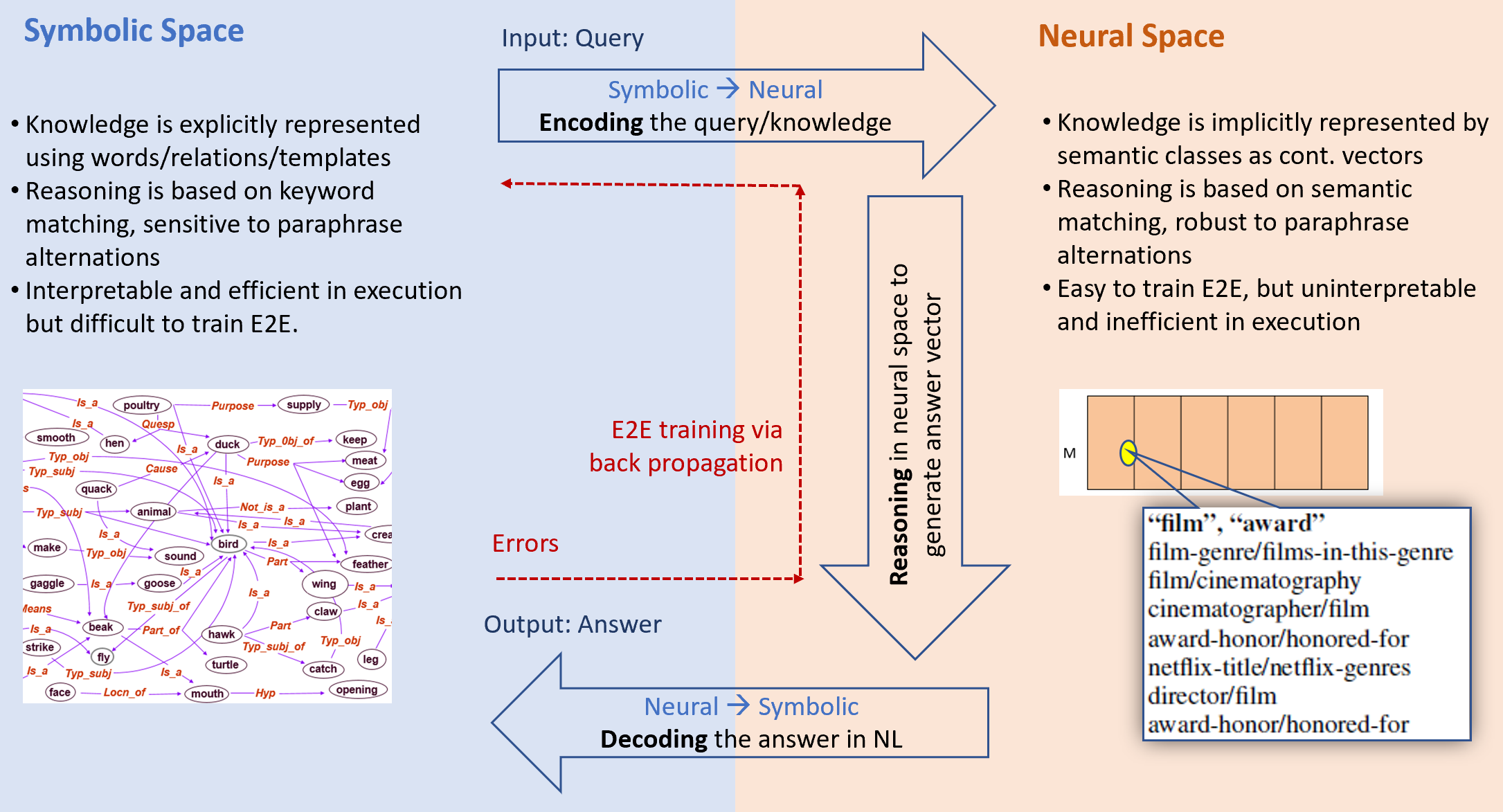}
\caption{Symbolic and Neural Computation.} 
\label{fig:symbolic-to-neural-shift} 
\vspace{0mm}
\end{figure}

Neural approaches do not rely on any human-defined symbolic representations but learn in a \emph{task-specific} neural space where task-specific knowledge is \emph{implicitly} represented as semantic concepts using low-dimensional continuous vectors. As \figref{fig:symbolic-to-neural-shift} illustrates, neural methods in NLP tasks (\eg, machine reading comprehension and dialogue) often consist of three steps: 
(1) \emph{encoding} symbolic user input and knowledge into their neural semantic representations, where semantically related or similar concepts are represented as vectors that are close to each other; (2) \emph{reasoning} in the neural space to generate a system response based on input and system state; and (3) \emph{decoding} the system response into a natural language output in a symbolic space. Encoding, reasoning and decoding are implemented using neural networks of different architectures, all of which may be stacked into a deep neural network trained in an end-to-end fashion via back propagation.

End-to-end training results in tighter coupling between the end application and the neural network architecture, lessening the need for traditional NLP component boundaries like morphological analysis and parsing. This drastically flattens the technology stack of \figref{fig:traditional-nlp-component-stack}, and substantially reduces the need for feature engineering. Instead, the focus has shifted to carefully tailoring the increasingly complex architecture of neural networks to the end application.

Although neural approaches have already been widely adopted in many AI tasks, including image processing, speech recognition and machine translation~\citep[e.g.,][]{goodfellow2016deep}, their impact on conversational AI has come somewhat more slowly. Only recently have we begun to observe neural approaches establish state-of-the-art results on an array of conversation benchmarks for both component tasks and end applications and, in the process, sweep aside the traditional component-based boundaries that have defined research areas for decades. 
This symbolic-to-neural shift is also reshaping the conversational AI landscape by opening up new tasks and user experiences that were not possible with older techniques.  One reason for this is that neural approaches provide a consistent representation for many modalities, capturing linguistic and non-linguistic (\eg, image and video \citep{mostafazadeh2017image}) features in the same modeling framework. 

There are also works on hybrid methods that combine the strengths of both neural and symbolic approaches \eg, \citep{mou2016coupling,liang2016neural}. 
As summarized in Fig.~\ref{fig:symbolic-to-neural-shift}, neural approaches can be trained in an end-to-end fashion and are robust to paraphrase alternations, but are weak in terms of execution efficiency and explicit interpretability. Symbolic approaches, on the other hand, are difficult to train and sensitive to paraphrase alternations, but are more interpretable and efficient in execution.

\chapter{Machine Learning Background}
\label{sec:neural-background}


This chapter presents a brief review of the deep learning and reinforcement learning technologies that are most relevant to conversational AI in later chapters. 

\section{Machine Learning Basics}

\citet{mitchell1997machine} defines machine learning broadly to include any computer program that improves its performance at some task $T$, measured by $P$, through experiences $E$. 

Dialogue, as summarized in \tabref{rl-dialogue}, is a well-defined learning problem with $T$, $P$, and $E$ specified as follows:

\begin{itemize}
\item{$T$: perform conversations with a user to fulfill the user's goal.}
\item{$P$: cumulative reward defined in~\tabref{rl-dialogue}.}
\item{$E$: a set of dialogues, each of which is a sequence of user-agent interactions.}
\end{itemize}

As a simple example, a single-turn QA dialogue agent might improve its performance \emph{as measured by accuracy or relevance of its generated answers} at \emph{the QA task}, through experiences of \emph{human-labeled question-answer pairs}.

A common recipe of building an ML agent using \emph{supervised learning} (SL) consists of a dataset, a model, a cost function (a.k.a. loss function) and an optimization procedure.

\begin{itemize}
\item{The dataset consists of $(x, y^*)$ pairs, where for each input $x$, there is a ground-truth output $y^*$. In QA, $x$ consists of an input question and the documents from which an answer is generated,  and $y^*$ is the desired answer provided by a knowledgeable external supervisor.}
\item{The model is typically of the form $y=f(x;\theta)$, where $f$ is a function (\eg, a neural network) parameterized by $\theta$ that maps input $x$ to output $y$.}
\item{The cost function is of the form $L(y^*, f(x;\theta))$. $L(.)$ is often designed as a smooth function of error, and is differentiable w.r.t. $\theta$. A commonly used cost function that meets these criteria is the \emph{mean squared error} (MSE), defined as
\[
\frac{1}{M} \sum_{i=1}^M (y^*_i-f(x_i;\theta))^2\,.
\]}
\item{The optimization can be viewed as a search algorithm to identify the best $\theta$ that minimize $L(.)$. Given that $L$ is differentiable, the most widely used optimization procedure for deep learning is mini-batch Stochastic Gradient Descent (SGD) which updates $\theta$ after each batch as
\begin{equation}
\theta \leftarrow \theta - \dfrac{\alpha}{N} \sum_{i=1}^N\nabla_\theta L(y^*_i, f(x_i;\theta))\,,  \label{eqn:sgd}
\end{equation}
where $N$ is the batch size and $\alpha$ the learning rate.
}
\end{itemize}

\paragraph{Common Supervised Learning Metrics.}
Once a model is trained, it can be tested on a \emph{hold-out} dataset to have an estimate of its generalization performance.  Suppose the model is $f(\cdot;\theta)$, and the hold-out set contains $N$ data points: $\Dset = \{(x_1, y_1^*), (x_2, y_2^*), \ldots, (x_N, y_N^*)\}$.

The first metric is the aforementioned mean squared error that is appropriate for regression problems (i.e., $y_i^*$ is considered real-values):
\[
\mMSE(f) \defeq \frac{1}{N} \sum_{i=1}^N (y_i^* - f(x_i;\theta))^2\,. 
\]

For classification problems, $y_i^*$ takes values from a finite set viewed as categories.  For simplicity, assume $y_i^*\in\{+1,-1\}$ here, so that an example $(x_i,y_i^*)$ is called positive (or negative) if $y_i^*$ is $+1$ (or $-1$).  The following metrics are often used:
\begin{itemize}
    \item Accuracy: the fraction of examples for which $f$ predicts correctly:
    \[
    \mACC(f) \defeq \frac{1}{N} \sum_{i=1}^N \1(f(x_i;\theta) = y_i^*)\,,
    \]
    where $\1(E)$ is $1$ if expression $E$ is true and $0$ otherwise.
    \item Precision: the fraction of correct predictions among examples that are predicted by $f$ to be positive:
    \[
    \mPRE(f) \defeq \frac{\sum_{i=1}^N \1(f(x_i;\theta) = y_i^* ~\text{AND}~ y_i^*=+1)}{\sum_{i=1}^N \1(f(x_i;\theta) = +1)}\,.
    \]
    \item Recall: the fraction of positive examples that are correctly predicted by $f$:
    \[
    \mREC(f) \defeq  \frac{\sum_{i=1}^N \1(f(x_i;\theta) = y_i^* ~\text{AND}~ y_i^*=+1)}{\sum_{i=1}^N \1(y_i^*=+1)}\,.
    \]
    \item F1 Score: the harmonic mean of precision and recall:
    \[
    \mFone(f) \defeq \frac{2 \times \mACC(f) \times \mREC(f)}{\mACC(f) + \mREC(f)}\,.
    \]
\end{itemize}
Other metrics are also widely used, especially for complex tasks beyond binary classification, such as the BLEU score~\citep{papineni2002bleu}.


\paragraph{Reinforcement Learning.} The above SL recipe applies to prediction tasks on a fixed dataset. However, in interactive problems such as dialogues\footnote{As shown in \tabref{rl-dialogue}, dialogue learning is formulated as RL where the agent learns a policy $\pi$ that in each dialogue turn chooses an appropriate action $a$ from the set $\Aset$, based on dialogue state $s$, so as to achieve the greatest cumulative reward.}, 
it can be challenging to obtain examples of desired behaviors that are both correct and representative of all the states in which the agent has to act. In unexplored territories, the agent has to learn how to act by interacting with an unknown environment on its own.  This learning paradigm is known as \emph{reinforcement learning} (RL), where there is a feedback loop between the agent and the external environment. 
In other words, while \emph{SL learns from previous experiences} provided by a knowledgeable external supervisor, \emph{RL learns by experiencing} on its own. RL differs from SL in several important respects \citep{sutton18reinforcement,mitchell1997machine}

\begin{itemize}
\item{\textbf{Exploration-exploitation tradeoff.} In RL, the agent needs to collect reward signals from the environment. This raises the question of which experimentation strategy results in more effective learning. The agent has to \emph{exploit} what it already knows in order to obtain high rewards, while having to \emph{explore} unknown states and actions in order to make better action selections in the future.}
\item{\textbf{Delayed reward and temporal credit assignment.} In RL, training information is not available in the form of $(x,y^*)$ as in SL. Instead, the environment provides only delayed rewards as the agent executes a sequence of actions. For example, we do not know whether a dialogue succeeds in completing a task until the end of the session. The agent, therefore, has to determine which of the actions in its sequence are to be credited with producing the eventual reward, a problem known as \emph{temporal credit assignment}.}
\item{\textbf{Partially observed states.} In many RL problems, the observation perceived from the environment at each step, \eg, user input in each dialogue turn, provides only partial information about the entire state of the environment based on which the agent selects the next action. Neural approaches learn a deep neural network to represent the state by encoding all information observed at the current and past steps, \eg, all the previous dialogue turns and the retrieval results from external databases.}
\end{itemize}

A central challenge in both SL and RL is \emph{generalization}, the ability to perform well on unseen inputs. Many learning theories and algorithms have been proposed to address the challenge with some success by, \eg, seeking a good tradeoff between the amount of available training data and the model capacity to avoid underfitting and overfitting. Compared to previous techniques, neural approaches provide a 
potentially more effective solution by leveraging the representation learning power of deep neural networks, as we will review in the next section.

\section{Deep Learning}

Deep learning (DL) involves training neural networks, which in their original form consisted of a single layer (\ie, the perceptron) \citep{rosenblatt1957perceptron}. The perceptron is incapable of learning even simple functions such as the logical XOR, so subsequent work explored the use of ``deep'' architectures, which added hidden layers between input and output~\citep{rosenblatt1962principles,minsky1969perceptrons}, a form of neural network that is commonly called the multi-layer perceptron (MLP), or deep neural networks (DNNs). This section introduces some commonly used DNNs for NLP and IR. Interested readers are referred to \citet{goodfellow2016deep} for a comprehensive discussion.

\subsection{Foundations}
\label{sec:deep-learning-foundations}

\begin{figure}[t]
\centering 
\includegraphics[width=1.00\linewidth]{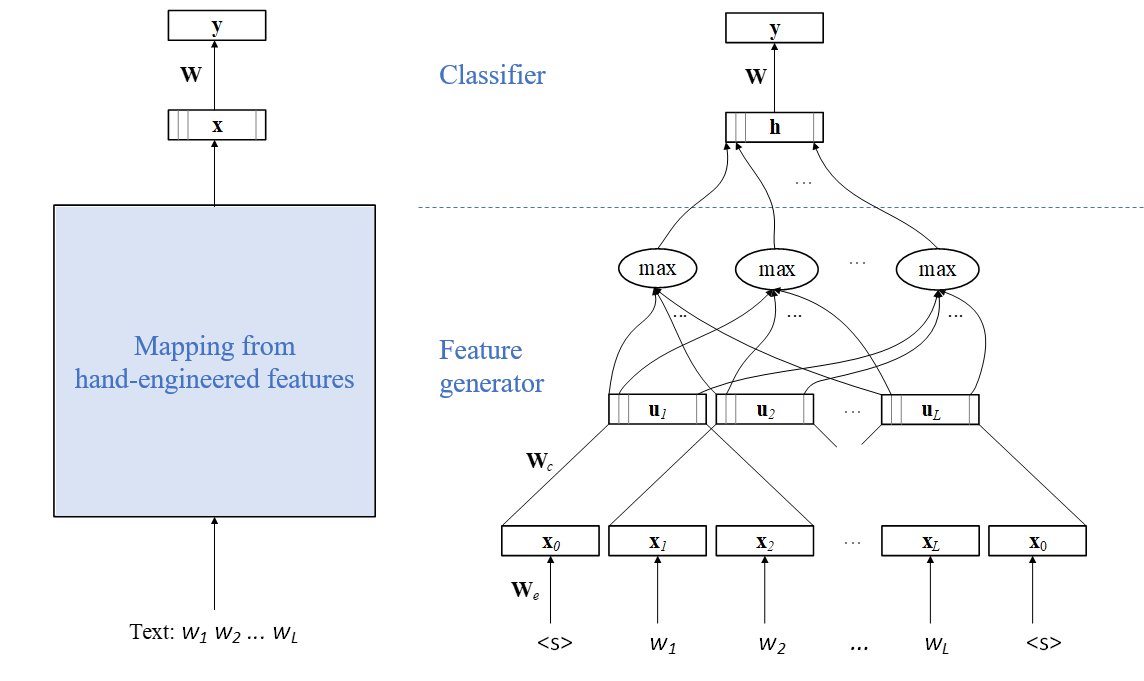}
\vspace{-2mm}
\caption{Flowcharts of classic machine learning (Left) and deep learning (Right). A convolutional neural network is used as an example for deep learning.}
\label{fig:classic-ml-dl} 
\vspace{-2mm}
\end{figure}

Consider a text classification problem: labeling a text string (\eg, a document or a query) by a domain name such as ``sport'' and ``politics''. As illustrated in \figref{fig:classic-ml-dl} (Left), a classical ML algorithm first maps a text string to a vector representation $\mathbf{x}$ using a set of hand-engineered features (\eg, word and character $n$-grams, entities, and phrases etc.), then learns a linear classifier with a softmax layer to compute the distribution of the domain labels $\mathbf{y}=f(\mathbf{x};\mathbf{W})$, where $\mathbf{W}$ is a matrix learned from training data using SGD to minimize the misclassification error. The design effort is focused mainly on feature engineering. 

Instead of using hand-designed features for $\mathbf{x}$, DL approaches jointly optimize the feature representation and classification using a DNN, as exemplified in \figref{fig:classic-ml-dl} (Right). We see that the DNN consists of two halves. The top half can be viewed as a linear classifier, similar to that in the classical ML model in \figref{fig:classic-ml-dl} (Left), except that its input vector $\mathbf{h}$ is not based on hand-engineered features but is learned using the bottom half of the DNN, which can be viewed as a feature generator optimized jointly with the classifier in an end-to-end fashion. Unlike classical ML, the effort of designing a DL classifier is mainly on optimizing DNN architectures for effective representation learning. 

For NLP tasks, depending on the type of linguistic structures that we hope to capture in the text, we may apply different types of neural network (NN) layer structures, such as convolutional layers for local word dependencies and recurrent layers for global word sequences. These layers can be combined and stacked to form a deep architecture to capture different semantic and context information at different abstract levels.  Several widely used NN layers are described below:

\paragraph{Word Embedding Layers.} In a symbolic space each word is represented as a one-hot vector whose dimensionality $n$ is the size of a pre-defined vocabulary.  The vocabulary is often large; \eg, $n>100K$. We apply a 
word embedding model, which is parameterized by a linear projection matrix $\mathbf{W}_e \in \Rset^{n \times m}$, to map each one-hot vector to a $m$-dimensional real-valued vector ($m \ll n$) in a neural space where the embedding vectors of the words that are more semantically similar are closer to each other.

\paragraph{Fully Connected Layers.} They perform linear projections as $\mathbf{W}^{\intercal} \mathbf{x}$.\footnote{We often omit the bias terms for simplifying notations in this paper.} 
We can stack multiple fully connected layers to form a deep feed-forward NN (FFNN) by introducing a nonlinear \emph{activation function} $g$ after each linear projection. If we view a text as a Bag-Of-Words (BOW) and let $\mathbf{x}$ be the sum of the embedding vectors of all words in the text, a deep FFNN can extract highly nonlinear features to represent hidden semantic topics of the text at different layers, \eg, 
$\mathbf{h}^{(1)}=g \left( \mathbf{W}^{(1)\intercal} \mathbf{x} \right)$ 
at the first layer, and 
$\mathbf{h}^{(2)}=g \left( \mathbf{W}^{(2)\intercal} \mathbf{h}^{(1)} \right)$
at the second layer, and so on, where $\mathbf{W}$'s are trainable matrices.

\paragraph{Convolutional-Pooling Layers.} An example of convolutional neural networks (CNNs) is shown in \figref{fig:classic-ml-dl} (Right).
A convolutional layer forms a local feature vector, denoted $\mathbf{u}_i$, of word $w_i$ in two steps.  It first generates a contextual vector $\mathbf{c}_i$ by concatenating the word embedding vectors of $w_i$ and its surrounding words defined by a fixed-length window.  It then performs a projection to obtain $\mathbf{u}_i = g \left( \mathbf{W}^{\intercal}_{c} \mathbf{c}_i \right)$, where $\mathbf{W}_{c}$ is a trainable matrix and $g$ is an activation function. Then, a pooling layer combines the outputs $\mathbf{u}_i, i=1...L$ into a single global feature vector $\mathbf{h}$. For example, in \figref{fig:classic-ml-dl}, the \emph{max pooling} operation is applied over each ``time'' $i$ of the sequence of the vectors computed by the convolutional layer to obtain $\mathbf{h}$, where each element is computed as $h_j=\max_{1 \le i \le L} u_{i,j}$. Another popular pooling function is \emph{average pooling}.

\begin{figure}[t]
\centering 
\includegraphics[width=0.45\linewidth]{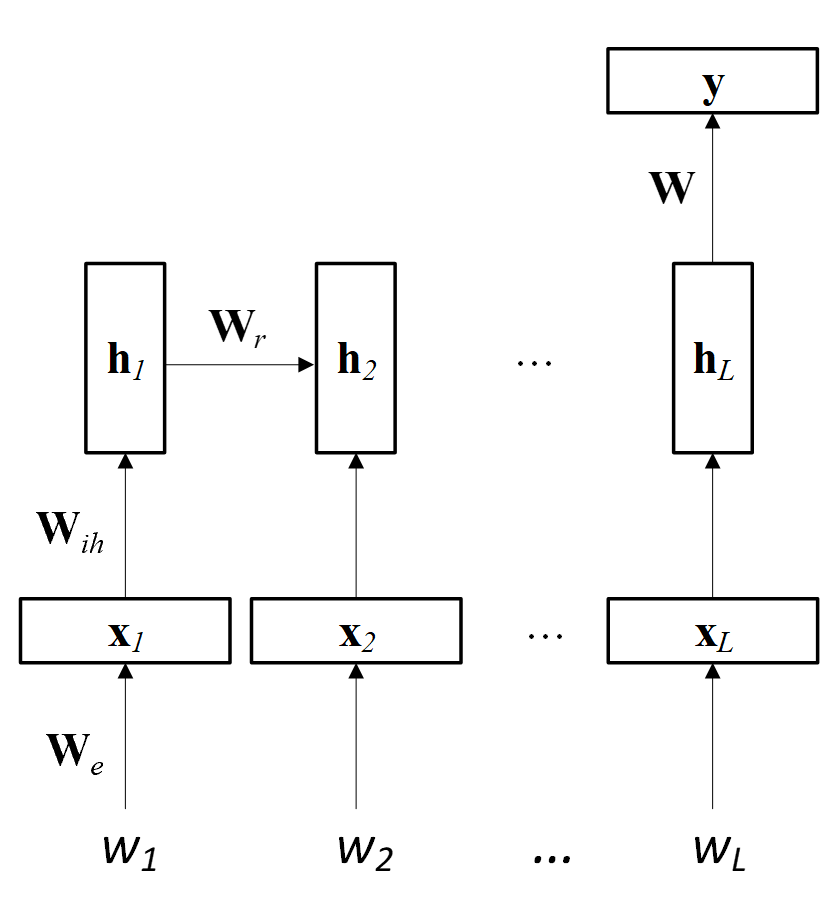}
\vspace{-2mm}
\caption{An example of recurrent neural networks.} 
\label{fig:rnn-example} 
\vspace{-2mm}
\end{figure}

\paragraph{Recurrent Layers.} An example of recurrent neural networks (RNNs) is shown in \figref{fig:rnn-example}. RNNs are commonly used for sentence embedding where we view a text as a sequence of words rather than a BOW. They map the text to a dense and low-dimensional semantic vector by sequentially and recurrently processing each word, and mapping the subsequence up to the current word into a low-dimensional vector as 
$\mathbf{h}_i = \text{RNN} (\mathbf{x}_i,\mathbf{h}_{i-1}) \defeq g\left( \mathbf{W}^{\intercal}_{ih} \mathbf{x}_i + \mathbf{W}^{\intercal}_{r} \mathbf{h}_{i-1} \right)$, 
where $\mathbf{x}_i$ is the word embedding of the $i$-th word in the text,
$\mathbf{W}_{ih}$ and $\mathbf{W}_{r}$ are trainable matrices, and
$\mathbf{h}_i$ is the semantic representation of the word sequence up to the $i$-th word.  

\subsection{Two Examples}
\label{case-study-dssm}

This section gives a brief description of two examples of DNN models, designed for the ranking and text generation tasks, respectively. They are composed of the NN layers described in the last section. 

\paragraph{DSSM for Ranking.}

In a ranking task, given an input query $x$, we want to rank all its candidate answers $y \in \mathcal{Y}$, based on a similarity scoring function $\text{sim}(x,y)$. The task is fundamental to many IR and NLP applications, such as query-document ranking, answer selection in QA, and dialogue response selection.

\begin{figure}[t]
\centering 
\includegraphics[width=0.40\linewidth]{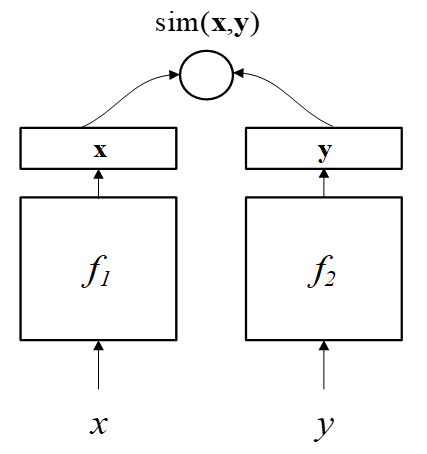}
\vspace{-2mm}
\caption{The architecture of DSSM.} 
\label{fig:dssm} 
\vspace{-2mm}
\end{figure}

DSSM stands for Deep Structured Semantic Models ~\citep{huang2013learning,shen2014latent}, or more generally, Deep Semantic Similarity Model~\citep{gao2014modeling}. DSSM is a deep learning model for measuring the semantic similarity of a pair of inputs $(x, y)$ \footnote{DSSM can be applied to a wide range of tasks depending on the definition of $(x, y)$. For example, $(x, y)$ is a query-document pair for Web search ranking \citep{huang2013learning,shen2014latent}, a document pair in recommendation \citep{gao2014modeling}, a question-answer pair in QA \citep{yih2015semantic}, a sentence pair of different languages in machine translation \citep{gao2014learning}, and an image-text pair in image captioning \citep{fang2015captions} and so on.}. 
As illustrated in \figref{fig:dssm}, a DSSM consists of a pair of DNNs, $f_1$ and $f_2$, which map inputs $x$ and $y$ into corresponding vectors in a common low-dimensional semantic space. Then the similarity of $x$ and $y$ is measured by the cosine distance of the two vectors. $f_1$ and $f_2$ can be of different architectures depending on $x$ and $y$. For example, to compute the similarity of an image-text pair, $f_1$ can be a deep convolutional NN and $f_2$ an RNN.

Let $\theta$ be the parameters of $f_1$ and $f_2$. $\theta$ is learned to identify the most effective feature representations of $x$ and $y$, optimized directly for end tasks. In other words, we learn a hidden semantic space, parameterized by $\theta$, where the semantics of distance between vectors in the space is defined by the task or, more specifically, the training data of the task. 
For example, in Web document ranking, the distance measures the query-document relevance, and $\theta$ is optimized using a pair-wise rank loss. Consider a query $x$ and two candidate documents $y^+$ and $y^-$, where $y^+$ is more relevant than $y^-$ to $x$. 
Let $\text{sim}_{\theta} (x,y)$ be the similarity of $x$ and $y$ in the semantic space parameterized by $\theta$ as
\[
\text{sim}_{\theta}(x,y) = \cos(f_1(x), f_2(y)).
\]
We want to maximize $\Delta = \text{sim}_{\theta}(x,y^+) - \text{sim}_{\theta}(x,y^-)$. We do so by optimizing a smooth loss function
\begin{equation}
L(\Delta; \theta)=\log \left( 1 + \exp{(-\gamma \Delta)} \right), 
\label{eqn:dssm-loss}
\end{equation}
where $\gamma$ is a scaling factor, using SGD of \eqnref{eqn:sgd}.

\paragraph{Seq2Seq for Text Generation.}

\begin{figure}[t]
\centering 
\includegraphics[width=0.75\linewidth]{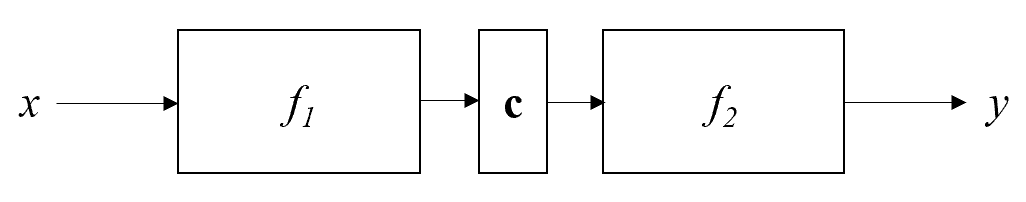}
\vspace{-2mm}
\caption{The architecture of seq2seq.} 
\label{fig:s2s} 
\vspace{-2mm}
\end{figure}

In a text generation task, given an input text $x$, we want to generate an output text $y$. This task is fundamental to applications such as machine translation and dialogue response generation. 

Seq2seq stands for the sequence-to-sequence architecture \citep{sutskever2014sequence}, which is also known as the encoder-decoder architecture~\citep{cho14learning}. Seq2Seq is typically implemented based on sequence models such as RNNs or gated RNNs. Gate RNNs, such as Long-Short Term Memory (LSTM)~\citep{hochreiter1997long} and the networks based on Gated Recurrent Unit (GRU)~\citep{cho14learning}, are the extensions of RNN in \figref{fig:rnn-example}, and are often more effective in capturing long-term dependencies due to the use of gated cells that have paths through time that have derivatives neither vanishing nor exploding. We will illustrate in detail how LSTM is applied to end-to-end conversation models in \secref{sec:e2econvo}.

Seq2seq defines the probability of generating $y$ conditioned on $x$ as $P(y|x)$ \footnote{Similar to DSSM, seq2seq can be applied to a variety of generation tasks depending on the definition of $(x,y)$. For example, $(x,y)$ is a sentence pair of different languages in machine translation \citep{sutskever2014sequence,cho14learning}, an image-text pairs in image captioning~\citep{vinyals2015show} (where $f_1$ is a CNN), and message-response pairs in dialogue~\citep{vinyals2015neural,li2015diversity}.}. As illustrated in \figref{fig:s2s}, a seq2seq model consists of (1) an input RNN or encoder $f_1$ that encodes input sequence $x$ into context vector $\mathbf{c}$, usually as a simple function of its final hidden state; and (2) an output RNN or decoder $f_2$ that generates output sequence $y$ conditioned on $\mathbf{c}$. $x$ and $y$ can be of different lengths. The two RNNs, parameterized by $\theta$, are trained jointly to minimize the loss function over all the pairs of $(x,y)$ in training data
\begin{equation}
L(\theta)=\frac{1}{M} \sum_{i=1...M} \log -P_{\theta}(y_i|x_i)\,.
\label{eqn:s2s-loss}
\end{equation}

\section{Reinforcement Learning}
\label{sec:basics:rl}

This section reviews reinforcement learning to facilitate discussions in later chapters.  For a comprehensive treatment of this topic, interested readers are referred to existing textbooks and reviews, such as \citet{sutton18reinforcement,
kaelbling96reinforcement,
bertsekas96neuro,
szepesvari10algorithms,
wiering12reinforcement,
li18deep}.

\subsection{Foundations} 
\label{sec:basics:rl:foundations}


Reinforcement learning (RL) is a learning paradigm where an intelligent agent learns to make optimal decisions by interacting with an initially unknown environment~\citep{sutton18reinforcement}.  Compared to supervised learning, a distinctive challenge in RL is to learn without a teacher (that is, without supervisory labels).  As we will see, this will lead to algorithmic considerations that are often unique to RL.

\begin{figure}
\centering
\includegraphics[width=0.6\textwidth]{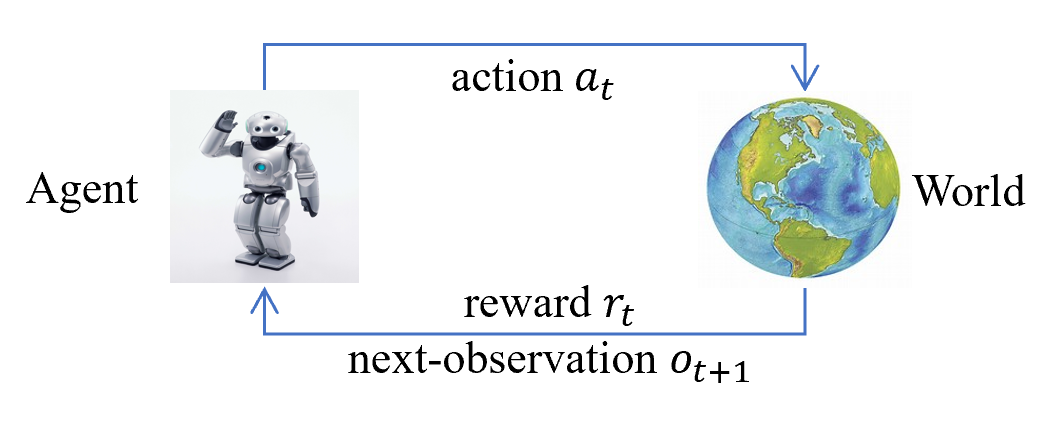}
\caption{Interaction between an RL agent and the external environment. 
} \label{fig:agent-env}
\end{figure}

As illustrated in \figref{fig:agent-env}, the agent-environment interaction is often modeled as a discrete-time Markov decision process, or MDP~\citep{puterman94markov}, described by a five-tuple $M = \langle \Sset, \Aset, P, R, \gamma\rangle$:
\begin{itemize}
\item{$\Sset$ is a possibly infinite set of states the environment can be in;}
\item{$\Aset$ is a possibly infinite set of actions the agent can take in a state;}
\item{$P(s'|s,a)$ gives the transition probability of the environment landing in a new state $s'$ after action $a$ is taken in state $s$;}
\item{$R(s,a)$ is the average reward immediately received by the agent after taking action $a$ in state $s$; and}
\item{$\gamma\in(0,1]$ is a discount factor.}
\end{itemize}

The intersection can be recorded as a trajectory $(s_1, a_1, r_1, \ldots)$, generated as follows: at step $t=1,2,\ldots$,
\begin{itemize}
\item{the agent observes the environment's current state $s_t\in\Sset$, and takes an action $a_t\in\Aset$;}
\item{the environment transitions to a next-state $s_{t+1}$, distributed according to the transition probabilities $P(\cdot|s_t,a_t)$;}
\item{associated with the transition is an immediate reward $r_t \in \Rset$, whose average is $R(s_t,a_t)$.}
\end{itemize}
Omitting the subscript, each step results in a tuple $(s,a,r,s')$ that is called a \emph{transition}. 
The goal of an RL agent is to maximize the long-term reward by taking optimal actions (to be defined soon).  Its action-selection policy, denoted by $\pi$, can be deterministic or stochastic.  In either case, we use $a \sim \pi(s)$ to denote selection of action by following $\pi$ in state $s$.  Given a policy $\pi$, the value of a state $s$ is the average discounted long-term reward from that state:
\[
V^\pi(s) \defeq \E[r_1 + \gamma r_2 + \gamma^2 r_3 + \cdots | s_1 = s, a_i \sim \pi(s_i), \forall i \geq 1]\,.
\]
We are interested in optimizing the policy so that $V^\pi$ is maximized for \emph{all} states.  Denote by $\pi^*$ an optimal policy, and $V^*$ its corresponding value function (also known as the optimal value function).  In many cases, it is more convenient to use another form of value function called the Q-function:
\[
Q^\pi(s,a) \defeq \E[r_1 + \gamma r_2 + \gamma^2 r_3 + \cdots | s_1 = s, a_1=a, a_i \sim \pi(s_i), \forall i>1]\,,
\]
which measures the average discounted long-term reward by first selecting $a$ in state $s$ and then following policy $\pi$ thereafter.  The optimal Q-function, corresponding to an optimal policy, is denoted by $Q^*$.

\subsection{Basic Algorithms}
\label{sec:basics:rl:algorithms}

We now describe two popular classes of algorithms, exemplified by Q-learning and policy gradient, respectively.

\paragraph{Q-learning.}

The first family is based on the observation that an optimal policy can be immediately retrieved if the optimal Q-function is available.  Specifically, the optimal policy can be determined by
\[
\pi^*(s) = \arg\max_a Q^*(s,a)\,.
\]
Therefore, a large family of RL algorithms focuses on learning $Q^*(s,a)$, and are collectively called \emph{value function-based} methods.

In practice, it is expensive to represent $Q(s,a)$ by a table, one entry for each distinct $(s,a)$, when the problem at hand is large.  For instance, the number of states in the game of Go is larger than $2 \times 10^{170}$~\citep{tromp06combinatorics}.  Hence, we often use compact forms to represent~$Q$.  In particular, we assume the $Q$-function has a predefined parametric form, parameterized by some vector $\theta\in\Rset^d$.  An example is linear approximation:
\[
Q(s,a;\theta) = \phi(s,a)^\mt \theta\,, 
\]
where $\phi(s,a)$ is a $d$-dimensional hand-coded feature vector for state-action pair $(s,a)$, and $\theta$ is the corresponding coefficient vector to be learned from data.  In general, $Q(s,a;\theta)$ may take different parametric forms. For example, in the case of Deep Q-Network (DQN),  $Q(s,a;\theta)$ takes the form of deep neural networks,
such as multi-layer perceptrons and convolutional networks~\citep{tesauro95temporal,mnih15human}, recurrent network~\citep{hausknecht15deep,li15recurrent}, etc.  More examples will be seen in later chapters. 
Furthermore, it is possible to represent the Q-function in a non-parametric way, using decision trees~\citep{ernst05tree} or Gaussian processes~\citep{engel05reinforcement}, which is outside of the scope of this introductory section.

To learn the Q-function, we modify the parameter $\theta$ using the following update rule, after observing a state transition $(s,a,r,s')$:
\begin{equation}
\theta \leftarrow \theta + \alpha\underbrace{\left(r + \gamma \max_{a'} Q(s',a';\theta) - Q(s,a;\theta)\right)}_{\text{``temporal difference''}} \nabla_\theta Q(s,a;\theta)\,.  \label{eqn:qlearn}
\end{equation}
The above update is known as Q-learning~\citep{watkins89learning}, which applies a small change to $\theta$, controlled by the step-size parameter $\alpha$ and computed from the \emph{temporal difference}~\citep{sutton88learning}.

While popular, in practice, Q-learning can be quite unstable and requires many samples before reaching a good approximation of $Q^*$. Two modifications are often helpful.  The first is \emph{experience replay}~\citep{lin92self}, popularized by \citet{mnih15human}. 
Instead of using an observed transition to update $\theta$ just \emph{once} using \eqnref{eqn:qlearn}, one may store it in a {replay buffer}, and periodically sample transitions from it to perform Q-learning updates.  This way, every transition can be used multiple times, thus increasing sample efficiency.  Furthermore, it helps stabilize learning by preventing the data distribution from changing too quickly over time when updating parameter $\theta$.

The second is a two-network implementation~\citep{mnih15human}, an instance of the more general fitted value iteration algorithm~\citep{munos08finite}.  Here, the learner maintains an extra copy of the Q-function, called the \emph{target network}, parameterized by $\theta_{\operatorname{target}}$.  During learning, $\theta_{\operatorname{target}}$ is fixed and is used to compute temporal difference to update $\theta$. 
Specifically, \eqnref{eqn:qlearn} now becomes:
\begin{equation}
\theta \leftarrow \theta + \alpha\underbrace{\left(r + \gamma \max_{a'} Q(s',a';\theta_{\operatorname{target}}) - Q(s,a;\theta)\right)}_{\text{temporal difference with a target network}} \nabla_\theta Q(s,a;\theta)\,.  \label{eqn:qlearn2}
\end{equation}
Periodically, $\theta_{\operatorname{target}}$ is updated to be $\theta$, and the process continues.

There have been a number of recent improvements to the basic Q-learning described above, such as dueling Q-network~\citep{wang16dueling}, double Q-learning~\citep{vanhasselt16deep}, and a provably convergent SBEED algorithm~\citep{dai18sbeed}. 

\paragraph{Policy Gradient.}

The other family of algorithms tries to optimize the policy directly, without having to learn the Q-function.  Here, the policy itself is directly parameterized by $\theta\in\Rset^d$, and $\pi(s;\theta)$ is often a distribution over actions.  Given any $\theta$, the policy is naturally evaluated by the average long-term reward it gets in a trajectory of length $H$, $\tau=(s_1, a_1,r_1, \ldots, s_H, a_H, r_H)$:\footnote{We describe policy gradient in the simpler bounded-length trajectory case, although it can be extended to problems when the trajectory length is unbounded~\citep{baxter01infinite,baxter01experiments}.}
\[
J(\theta) \defeq \E\left[\sum_{t=1}^H \gamma^{t-1} r_t | a_t \sim \pi(s_t;\theta)\right]\,.
\]
If it is possible to estimate the gradient $\nabla_\theta J$ from sampled trajectories, one can do stochastic gradient ascent\footnote{Stochastic gradient \emph{ascent} is simply stochastic gradient \emph{descent} on the negated objective function.} to maximize $J$:
\begin{equation}
\theta \leftarrow \theta + \alpha \nabla_\theta J(\theta)\,, \label{eqn:generic-pg}
\end{equation}
where $\alpha$ is again a stepsize parameter.

One such algorithm, known as REINFORCE~\citep{williams92simple}, estimates the gradient as follows.  Let $\tau$ be a length-$H$ trajectory generated by $\pi(\cdot;\theta)$; that is, $a_t \sim \pi(s_t;\theta)$ for every $t$.  Then, a stochastic gradient based on this single trajectory is given by
\begin{equation}
\nabla_\theta J(\theta) = \sum_{t=1}^{H-1} \gamma^{t-1} \left(\nabla_\theta \log \pi(a_t|s_t;\theta) \sum_{h=t}^H \gamma^{h-t}r_h\right)\,. \label{eqn:reinforce}
\end{equation}

REINFORCE may suffer high variance in practice, as its gradient estimate depends directly on the sum of rewards along the entire trajectory.  Its variance may be reduced by the use of an estimated value function of the current policy, often referred to as the critic in actor-critic algorithms~\citep{sutton00policy,konda00actor}:
\begin{equation}
\nabla_\theta J(\theta) = \sum_{t=1}^{H-1} \gamma^{t-1} \left(\nabla_\theta \log \pi(a_t|s_t;\theta) \hat{Q}(s_t,a_t,h) \right)\,, \label{eqn:actor-critic}
\end{equation}
where $\hat{Q}(s,a,h)$ is an estimated value function for the current policy $\pi(s;\theta)$ that is used to approximate $\sum_{h=t}^H \gamma^{h-t}r_h$ in \eqnref{eqn:reinforce}.  
$\hat{Q}(s,a,h)$ may be learned by standard temporal difference methods (similar to Q-learning), but many variants exist. 
Moreover, there has been much work on methods to compute the gradient $\nabla_\theta J$ more effectively than 
\eqnref{eqn:actor-critic}. Interested readers can refer to a few related works and the references therein for further details~\citep{kakade02natural,peters05natural,
schulman15trust,schulman15high,
mnih16asynchronous,gu17qprop,
dai18boosting,liu18action}.

\subsection{Exploration}

So far we have described basic algorithms for updating either the value function or the policy, when transitions are given as input.  Typically, an RL agent also has to determine how to select actions to collect desired transitions for learning.  Always selecting the action (``exploitation'') that seems best is problematic, as not selecting a novel action (that is, underrepresented, or even absent, in data collected so far), known as ``exploration'', may result in the risk of not seeing outcomes that are potentially better.  Balancing exploration and exploitation efficiently is one of the unique challenges in reinforcement learning.

A basic exploration strategy is known as $\epsilon$-greedy.  The idea is to choose the action that looks best with high probability (for exploitation), and a random action with small probability (for exploration).  In the case of DQN, suppose $\theta$ is the current parameter of the Q-function, then the action-selection rule for state $s$ is given as follows:
\begin{eqnarray*}
a_t = \begin{cases}
\arg\max_a Q(s_t,a;\theta) & \text{with probability $1-\epsilon$} \\
\text{random action} & \text{with probability $\epsilon$\,.}
\end{cases}
\end{eqnarray*}
In many problems this simple approach is effective (although not necessarily optimal).  A further discussion is found in \secref{sec:dialogue:exploration}.

\chapter{Question Answering and Machine Reading Comprehension}
\label{sec:qa-bot}

Recent years have witnessed an increasing demand for conversational Question Answering (QA) agents that allow users to query a large-scale Knowledge Base (KB) or a document collection in natural language. The former is known as KB-QA agents and the latter text-QA agents. KB-QA agents are more flexible and user-friendly than traditional SQL-like systems in that users can query a KB interactively without composing complicated SQL-like queries. Text-QA agents are much easier to use in mobile devices than traditional search engines, such as Bing and Google, in that they provide concise, direct answers to user queries, as opposed to a ranked list of relevant documents. 

It is worth noting that multi-turn, conversational QA is an emerging research topic, and is not as well-studied as single-turn QA. Many papers reviewed in this chapter are focused on the latter. However, single-turn QA is an indispensable building block for all sorts of dialogues (\eg, chitchat and task-oriented), deserving our full attention if we are to develop real-world dialogue systems.

In this chapter, we start with a review of KB and symbolic approaches to KB-QA based on semantic parsing. We show that a symbolic system is hard to scale 
because the keyword-matching-based, query-to-answer inference used by the system is inefficient for a very large KB, and is not robust to paraphrasing. To address these issues, neural approaches are developed to represent queries and KB using continuous semantic vectors so that the inference can be performed at the semantic level in a compact neural space. 
We also describe the typical architecture of multi-turn, conversational KB-QA agents, using a movie-on-demand agent as an example, and review several conversational KB-QA datasets developed recently. 
 
We then discuss neural text-QA agents. The heart of these systems is a neural Machine Reading Comprehension (MRC) model that generates an answer to an input question based on a (set of) passage(s). After reviewing popular MRC datasets and TREC text-QA open benchmarks, we describe the technologies developed for state-of-the-art MRC models along two dimensions: (1) the methods of encoding questions and passages as vectors in a neural space, and (2) the methods of performing reasoning in the neural space to generate the answer. We also describe the architecture of multi-turn, conversational text-QA agents, and the way MRC tasks and models are extended to conversational QA. 


\section{Knowledge Base}
\label{sec:knowledge-base}

Organizing the world's facts and storing them in a structured database, large scale Knowledge Bases (KB) like DBPedia \citep{auer2007dbpedia}, Freebase \citep{bollacker2008freebase} and Yago \citep{suchanek2007yago} have become important resources for supporting open-domain QA. 

A typical KB consists of a collection of subject-predicate-object triples $(s,r,t)$ where $s, t\in\mathcal{E}$ are entities and $r\in\mathcal{R}$ is a predicate or relation. A KB in this form is often called a Knowledge Graph (KG) due to its graphical representation, \ie, the entities are nodes and the relations the directed edges that link the nodes.

\begin{figure}[t]
\centering 
\includegraphics[width=1\linewidth]{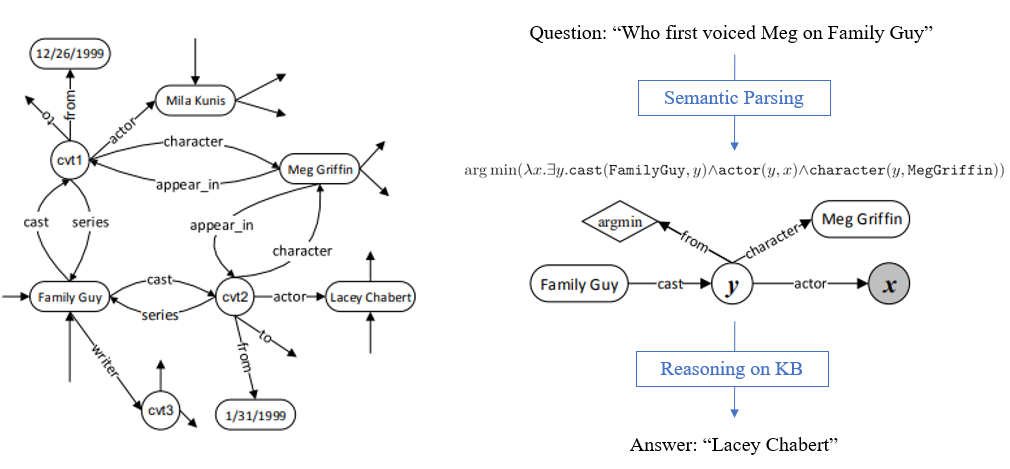}
\vspace{-2mm}
\caption{An example of semantic parsing for KB-QA. (Left) A subgraph of Freebase related to the TV show \texttt{Family Guy}. (Right) A question, its logical form in $\lambda$-calculus and query graph, and the answer. Figures adapted from \citet{yih2015semantic}.} 
\label{fig:freebase-semantic-parsing-example} 
\vspace{0mm}
\end{figure}

\figref{fig:freebase-semantic-parsing-example} (Left) shows a small subgraph of Freebase related to the TV show \texttt{Family Guy}. Nodes include some names, dates and special Compound Value Type (CVT) entities.\footnote{CVT is not a real-world entity, but is used to collect multiple fields of an event or a special relationship.} A directed edge describes the relation between two entities, labeled by a predicate.


\section{Semantic Parsing for KB-QA}
\label{sec:semantic-parsing-kb-qa}

Most state-of-the-art symbolic approaches to KB-QA are based on semantic parsing, where a question is mapped to its formal meaning representation (\eg, logical form) and then translated to a KB query. The answers to the question can then be obtained by finding a set of paths in the KB that match the query and retrieving the end nodes of these paths \citep{richardson1998mindnet,berant2013semantic,yao2014information,bao2014knowledge,yih2015deep}.

We take the example used in \citet{yih2015semantic} to illustrate the QA process. \figref{fig:freebase-semantic-parsing-example} (Right) shows the logical form in $\lambda$-calculus and its equivalent graph representation, known as \emph{query graph}, of the question ``Who first voiced Meg on Family Guy?''. Note that the query graph is grounded in Freebase. The two entities, \texttt{MegGriffin} and \texttt{FamilyGuy}, are represented by two rounded rectangle nodes. The circle node $y$ means that there should exist an entity describing some casting relations like the character, actor and the time she started the role. $y$ is grounded in a CVT entity in this case. The shaded circle node $x$ is also called the \emph{answer node}, and is used to map entities retrieved by the query. The diamond node $\arg\min$ constrains that the answer needs to be the earliest actor for this role. Running the query graph without the aggregation function against the graph as in \figref{fig:freebase-semantic-parsing-example} (Left) will match both \texttt{LaceyChabert} and \texttt{MilaKunis}. But only \texttt{LaceyChabert} is the correct answer as she started this role earlier (by checking the from property of the grounded CVT node).


Applying a symbolic KB-QA system to a very large KB is challenging for two reasons:
\begin{itemize}
\item  \textbf{Paraphrasing in natural language}: This leads to a wide variety of semantically equivalent ways of stating the same question, and in the KB-QA setting, this may cause mismatches between the natural language questions and the label names (\eg, predicates) of the nodes and edges used in the KB. As in the example of \figref{fig:freebase-semantic-parsing-example}, we need to measure how likely the predicate used in the question matches that in Freebase, such as ``Who first \emph{voiced} Meg on Family Guy?'' vs. \texttt{cast-actor}. \citet{yih2015semantic} proposed to use a learned DSSM described in \secref{case-study-dssm}, which is conceptually an embedding-based method we will review in \secref{embedding-based-methods}.
\item \textbf{Search complexity}: Searching all possible multi-step (compositional) relation paths that match complex queries is prohibitively expensive because the number of candidate paths grows exponentially with the path length. We will review symbolic and neural approaches to multi-step reasoning in \secref{sec:multi-step-reasoning-kb}.
\end{itemize}


\section{Embedding-based Methods}
\label{embedding-based-methods}

To address the paraphrasing problem, embedding-based methods map entities and relations in a KB to continuous vectors in a neural space; see, \eg, \citet{bordes2013translating,socher2013reasoning,yang2015embedding,yih2015deep}. This space can be viewed as a hidden semantic space where various expressions with the same semantic meaning map to the same continuous vector. 

Most KB embedding models are developed for the Knowledge Base Completion (KBC) task: predicting the existence of a triple $(s,r,t)$ that is not seen in the KB. This is a simpler task than KB-QA since it only needs to predict whether a fact is true or not, and thus does not suffer from the search complexity problem.

The bilinear model is one of the basic KB embedding models \citep{yang2015embedding,nguyen2017overview}. It learns a vector $\mathbf{x}_e \in \Rset^d$ for each entity $e\in \mathcal{E}$ and a matrix $\mathbf{W}_r \in \Rset^{d \times d}$ for each relation $r \in \mathcal{R}$. The model scores how likely a triple $(s,r,t)$ holds using 
\begin{equation}
\text{score}(s,r,t;\theta)=\mathbf{x}_s^\top \mathbf{W}_r \mathbf{x}_t.
\label{eqn:kb-embedding}
\end{equation}
The model parameters $\theta$ (\ie, the embedding vectors and matrices) are trained on pair-wise training samples in a similar way to that of the DSSM described in \secref{case-study-dssm}. For each positive triple $(s,r,t)$ in the KB, denoted by $x^+$, we construct a set of negative triples $x^-$ by corrupting $s$, $t$, or $r$. The training objective is to minimize the pair-wise rank loss of \eqnref{eqn:dssm-loss}, or more commonly the margin-based loss defined as
\[
L(\theta)= \sum_{(x^+,x^-) \in \mathcal{D}} \left[ \gamma + \text{score} (x^-;\theta)-\text{score} (x^+;\theta) \right]_+, 
\]
where $[x]_+ \defeq \max(0,x)$, $\gamma$ is the margin hyperparameter, and $\mathcal{D}$ the training set of triples.

These basic KB models have been extended to answer multi-step relation queries, as known as \emph{path queries}, \eg, ``Where did Tad Lincoln's parents live?'' \citep{toutanova2016compositional,guu2015traversing,neelakantan2015compositional}. A path query consists of an initial anchor entity $s$ (\eg, \texttt{TadLincoln}), followed by a sequence of relations to be traversed $(r_1,...,r_k)$ (\eg, \texttt{(parents, location)}). We can use vector space compositions to combine the embeddings of individual relations $r_i$ into an embedding of the path $(r_1,...,r_k)$.
The natural composition of the bilinear model of \eqnref{eqn:kb-embedding} is matrix multiplication. Thus, to answer how likely a path query $(q,t)$ holds, where $q=(s,r_1,...,r_k)$, we would compute
\begin{equation}
\text{score}(q,t)=\mathbf{x}_s^\top \mathbf{W}_{r_1}...\mathbf{W}_{r_k} \mathbf{x}_t.
\label{eqn:kb-embedding-composition}
\end{equation}

These KB embedding methods are shown to have good generalization performance in terms of validating unseen facts (\eg, triples and path queries) given an existing KB. 
Interested users are referred to \citet{nguyen2017overview} for a detailed survey of embedding models for KBC.



\section{Multi-Step Reasoning on KB}
\label{sec:multi-step-reasoning-kb}

\begin{figure}[t]
\centering 
\includegraphics[width=0.76\linewidth]{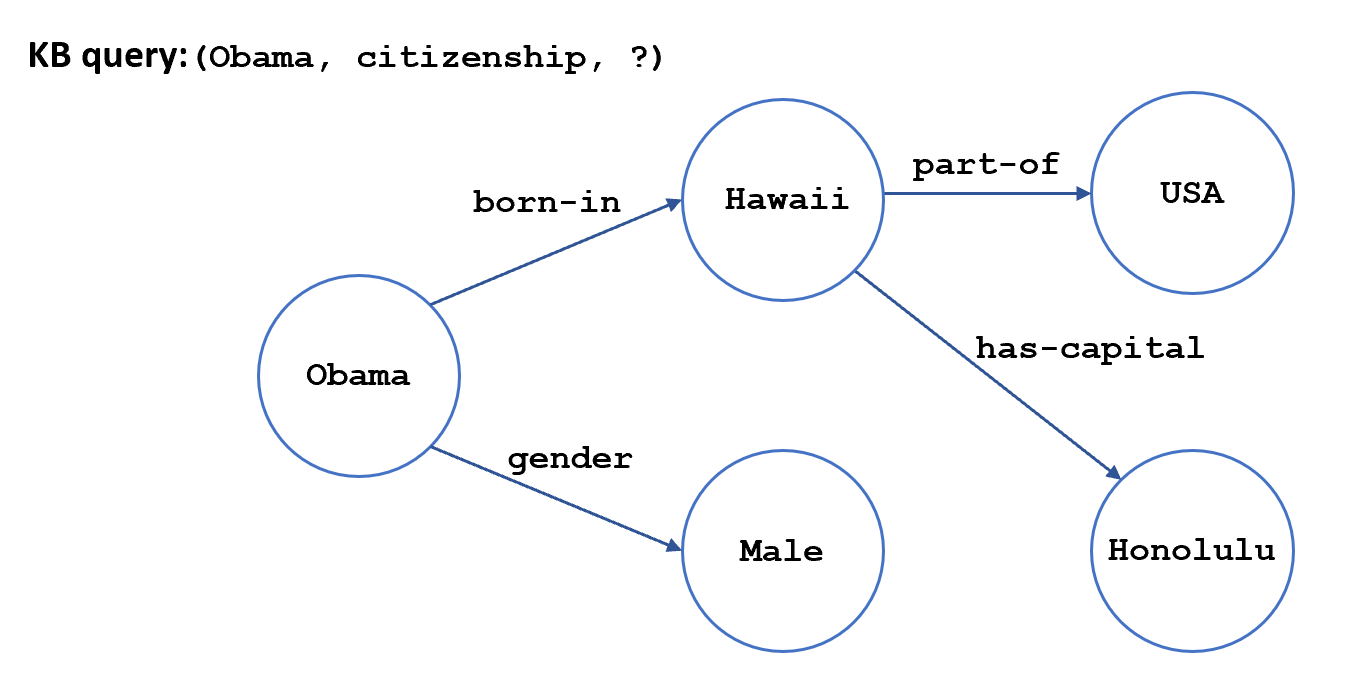}
\caption{An example of knowledge base reasoning (KBR). We want to identify the answer node \texttt{USA} for a KB query \texttt{(Obama, citizenship, ?)}. Figure adapted from \citet{shen2018reinforcewalk}.} 
\label{fig:kbr-example} 
\end{figure}

Knowledge Base Reasoning (KBR) is a subtask of KB-QA. As described in \secref{sec:semantic-parsing-kb-qa}, KB-QA is performed in two steps: (1) semantic parsing translates a question into a KB query, then (2) KBR traverses the query-matched paths in a KB to find the answers.

To reason over a KB, for each relation $r \in \mathcal{R}$, we are interested in learning a set of first-order logical rules in the form of \emph{relational paths}, $\pi = (r_1,...,r_k)$. 
For the KBR example in \figref{fig:kbr-example}, given the question ``What is the citizenship of Obama?'', its translated KB query in the form of subject-predicate-object triple is \texttt{(Obama, citizenship, ?)}. Unless the triple \texttt{(Obama, citizenship, USA)} is explicitly stored in the KB,\footnote{As pointed out by \citet{nguyen2017overview}, even very large KBs, such as Freebase and DBpedia, which contain billions of fact triples about the world, are still far from complete.} a multi-step reasoning procedure is needed to deduce the answer from the paths that contain relevant triples, such as \texttt{(Obama, born-in, Hawaii)} and \texttt{(Hawaii, part-of, USA)}, using the learned relational paths such as \texttt{(born-in, part-of)}. 

Below, we describe three categories of multi-step KBR methods. They differ in whether reasoning is performed in a discrete symbolic space or a continuous neural space.

\subsection{Symbolic Methods}

\begin{table}[t]
\footnotesize
\caption{A sample of relational paths learned by PRA. For each relation, its top-2 PRA paths are presented, adapted from \citet{lao2011random}.}
\begin{tabular}{ll}
\hline
ID & PRA Path \# Comment  \\ \hline
   & \textbf{\texttt{athlete-plays-for-team}} \\
1  & \begin{tabular}[c]{@{}l@{}} \texttt{(athlete-plays-in-league, league-players,} \\ \end{tabular} \\
   & \begin{tabular}[c]{@{}l@{}} \texttt{~athlete-plays-for-team)} \\ \end{tabular} \\
   & \begin{tabular}[c]{@{}l@{}}
   \# teams with many players in the athlete's league
   \end{tabular} \\
2  & \begin{tabular}[c]{@{}l@{}} \texttt{(athlete-plays-in-league, league-teams, team-against-team)} \\ \end{tabular} \\
   & \begin{tabular}[c]{@{}l@{}}
   \# teams that play against many teams in the athlete's league
   \end{tabular} \\
   & \textbf{\texttt{stadium-located-in-city}}  \\
1  & \begin{tabular}[c]{@{}l@{}} \texttt{(stadium-home-team,team-home-stadium,stadium-located-in-city)}  \\  \end{tabular}          \\
   & \begin{tabular}[c]{@{}l@{}}
   \# city of the stadium with the same team
   \end{tabular} \\
2  & \begin{tabular}[c]{@{}l@{}} \texttt{(latitude-longitude,latitude-longitude-of,}  \\  \end{tabular}          \\
   & \begin{tabular}[c]{@{}l@{}} \texttt{~stadium-located-in-city)}  \\  \end{tabular}          \\
   & \begin{tabular}[c]{@{}l@{}}
   \# city of the stadium with the same location
   \end{tabular} \\
   & \textbf{\texttt{team-home-stadium}}  \\
1  & \begin{tabular}[c]{@{}l@{}} \texttt{(team-plays-in-city,city-stadium)} \\ \end{tabular} \\
   & \begin{tabular}[c]{@{}l@{}}
   \# stadium located in the same city with the query team
   \end{tabular} \\
2  & \begin{tabular}[c]{@{}l@{}} \texttt{(team-member,athlete-plays-for-team,team-home-stadium)}  \\ \end{tabular} \\
   & \begin{tabular}[c]{@{}l@{}}
   \# home stadium of teams which share players with the query
   \end{tabular} \\
   & \textbf{\texttt{team-plays-in-league}}  \\
1  & \begin{tabular}[c]{@{}l@{}} \texttt{(team-plays-sport,players,athlete-players-in-league)}  \\  \end{tabular} \\
   & \begin{tabular}[c]{@{}l@{}}
   \# the league that the query team's members belong to
   \end{tabular} \\
2  & \begin{tabular}[c]{@{}l@{}} \texttt{(team-plays-against-team,team-players-in-league)}  \\ \end{tabular} \\ 
   & \begin{tabular}[c]{@{}l@{}}
   \# the league that query team's competing team belongs to
   \end{tabular} \\ \\ \hline
\end{tabular}
\label{tab:pra-path}
\end{table}

The Path Ranking Algorithm (PRA) \citep{lao2010relational,lao2011random} is one of the primary symbolic approaches to learning relational paths in large KBs. PRA uses random walks with restarts to perform multiple bounded depth-first search to find relational paths.  \tabref{tab:pra-path} shows a sample of relational paths learned by PRA. A relational path is a sequence $\pi = (r_1,...,r_k)$. An instance of the relational path is a sequence of nodes $e_1,...,e_{k+1}$ such that $(e_i,r_i,e_{i+1})$ is a valid triple.

During KBR, given a query $q=(s,r,?)$, PRA selects the set of relational paths for $r$, denoted by $\mathcal{B}_r = \{\pi_1,\pi_2,...\}$, then traverses the KB according to the query and $\mathcal{B}_r$, and scores each candidate answer $t$ using a linear model
\begin{equation} \label{eqn:pra}
\text{score}(q,t)= \sum_{\pi \in \mathcal{B}_r} \lambda_\pi P(t|s,\pi)\,,
\end{equation}
where $\lambda_\pi$'s are the learned weights, and $P(t|s,\pi)$ is the probability of reaching $t$ from $s$ by a random walk that instantiates the relational path $\pi$, also known as a \emph{path constrained random walk}.

Because PRA operates in a fully discrete space, it does not take into account semantic similarities among relations. As a result, PRA can easily produce millions of categorically distinct paths even for a small path length, which not only hurts generalization but makes reasoning prohibitively expensive. 
To reduce the number of relational paths that need to be considered in KBR, \citet{lao2011random} used heuristics (e.g., requiring that a path be included in PRA only if it retrieves at least one target entity in the training data) and added an $L_1$ regularization term in the loss function for training the linear model of \eqnref{eqn:pra}. 
%
\citet{gardner2014incorporating} proposed a modification to PRA that leverages the KB embedding methods, as described in \secref{embedding-based-methods}, to collapse and cluster PRA paths according to their relation embeddings.

\subsection{Neural Methods}

\begin{figure}[t]
\centering 
\includegraphics[width=0.96\linewidth]{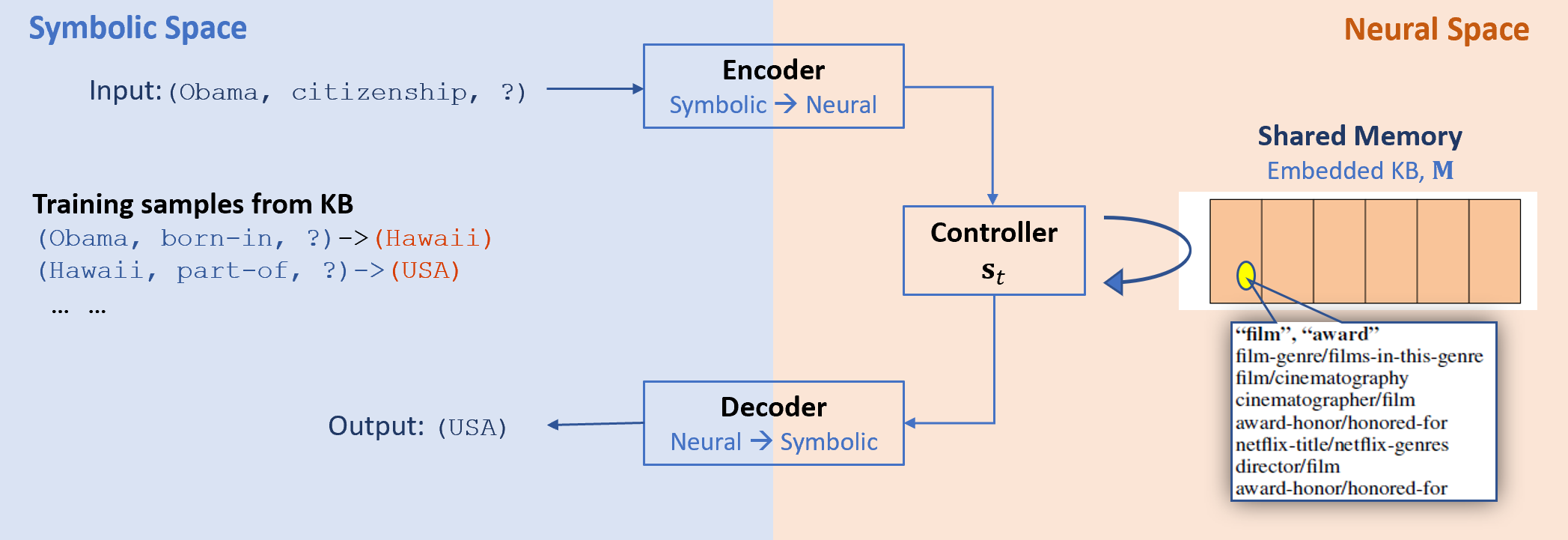}
\vspace{0mm}
\caption{An overview of the neural methods for KBR \citep{shen2017traversing,yang2017differentiable}. The KB is embedded in neural space as matrix $\mathbf{M}$ that is learned to store compactly the connections between related triples (\eg, the relations that are semantically similar are stored as a cluster). The controller is designed to adaptively produce lookup sequences in $\mathbf{M}$ and decide when to stop, and the encoder and decoder are responsible for the mapping between the symbolic and neural spaces.} 
\label{fig:irn-architecture} 
\vspace{0mm}
\end{figure}

Implicit ReasoNet (IRN) \citep{shen2016implicit,shen2017traversing} and Neural Logic Programming (Neural LP) \citep{yang2017differentiable} are proposed to perform multi-step KBR in a neural space and achieve state-of-the-art results on popular benchmarks. The overall architecture of these methods is shown in \figref{fig:irn-architecture}, which can be viewed as an instance of the neural approaches illustrated in \figref{fig:symbolic-to-neural-shift}  (Right). In what follows, we use IRN as an example to illustrate how these neural methods work. IRN consists of four modules: encoder, decoder, shared memory, and controller, as in \figref{fig:irn-architecture}.

\paragraph{Encoder and Decoder}

These two modules are task-dependent. Given an input query $(s,r,?)$, the encoder maps $s$ and $r$, respectively, into their embedding vectors and then concatenates the two vectors to form the initial hidden state vector $\mathbf{s}_0$ of the controller.  The use of vectors rather than matrices for relation representations is inspired by the bilinear-diag model \citep{yang2015embedding}, which restricts the relation representations to the class of diagonal matrices.

The decoder outputs a prediction vector $\mathbf{o}=\tanh(\mathbf{W}_o^\top \mathbf{s}_t + \mathbf{b}_o)$, a nonlinear projection of state $\mathbf{s}$ at time $t$, where $\mathbf{W}_o$ and $\mathbf{b}_o$ are the weight matrix and bias vector, respectively. In KBR, we can map the answer vector $\mathbf{o}$ to its answer node (entity) $o$ in the symbolic space based on $L_1$ distance as $o = \arg \min_{e \in \mathcal{E}} \|\mathbf{o}- \mathbf{x}_e\|_1$, where 
$\mathbf{x}_e$ is the embedding vector of entity $e$.  

\paragraph{Shared Memory}

The shared memory $\mathbf{M}$ is differentiable, and consists of a list of vectors $\{\mathbf{m}_i\}_{1 \le i \le |\mathbf{M}|}$ that are randomly initialized and updated through back-propagation in training.  $\mathbf{M}$ stores a compact version of KB optimized for the KBR task. That is, each vector represents a concept (a cluster of relations or entities) and the distance between vectors represents the semantic relatedness of these concepts. For example, the system may fail to answer the question \texttt{(Obama, citizenship, ?)} even if it finds the relevant facts in $\mathbf{M}$, such as \texttt{(Obama, born-in, Hawaii)} and \texttt{(Hawaii, part-of, USA)}, because it does not know that \texttt{bore-in} and \texttt{citizenship} are semantically related relations. In order to correct the error, 
$\mathbf{M}$ needs to be updated using the gradient to encode the piece of new information by moving the two relation vectors closer to each other in the neural space. 

\paragraph{Controller}

The controller is implemented as an RNN. Given initial state $\mathbf{s}_0$, it uses attention to iteratively lookup and fetch information from $\mathbf{M}$ to update the state $\mathbf{s}_t$ at time $t$ according to \eqnref{eqn:controller}, until it decides to terminate the reasoning process and calls the decoder to generate the output.
\begin{equation} \label{eqn:controller}
\begin{aligned}
a_{t,i} &=  \frac{\exp \left( \lambda \cos(\mathbf{W}_1^\top \mathbf{m}_i, \mathbf{W}_2^\top \mathbf{s}_t) \right) }{\sum_k \exp \left( \lambda \cos(\mathbf{W}_1^\top \mathbf{m}_k, \mathbf{W}_2^\top \mathbf{s}_t) \right) } , \\
\mathbf{x}_t &= \sum_i^{|\mathbf{M}|} a_{t,i} \mathbf{m}_i, \\
\mathbf{s}_{t+1} &= g(\mathbf{W}_3^\top \mathbf{s}_t + \mathbf{W}_4^\top \mathbf{x}_t),
\end{aligned}
\end{equation}
where $\mathbf{W}$'s are learned projection matrices, $\lambda$ a scaling factor and $g$ a nonlinear activation function.

The reasoning process of IRN can be viewed as a Markov Decision Process (MDP), as illustrated in \secref{sec:basics:rl:foundations}. The step size in the information lookup and fetching sequence of \eqnref{eqn:controller} is not given by training data, but is decided by the controller on the fly. More complex queries need more steps. Thus, IRN learns a stochastic policy to get a distribution over termination and prediction actions by the REINFORCE algorithm \citep{williams92simple}, which is described in \secref{sec:basics:rl:algorithms} and \eqnref{eqn:reinforce}. Since all the modules of IRN are differentiable, IRN is an end-to-end differentiable neural model whose parameters, including the embedded KB matrix $\mathbf{M}$, can be jointly optimized using SGD on the training samples derived from a KB, as shown in \figref{fig:irn-architecture}.

As outlined in \figref{fig:symbolic-to-neural-shift}, neural methods operate in a continuous neural space, and do not suffer from the problems associated with symbolic methods. They are robust to paraphrase alternations because knowledge is implicitly represented by semantic classes via continuous vectors and matrices. 
They also do not suffer from the search complexity issue even with complex queries (\eg path queries) and a very large KB 
because they reason over a compact representation of a KB (\eg, the matrix $\mathbf{M}$ in the shared memory in IRN) rather than the KB itself. 

One of the major limitations of these methods is the lack of interpretability. Unlike PRA which traverses the paths in the graph explicitly as \eqnref{eqn:pra}, IRN does not follow explicitly any path in the KB during reasoning but performs lookup operations over the shared memory iteratively using the RNN controller with attention, each time using the revised internal state $\mathbf{s}$ as a query for lookup. It remains challenging to recover the symbolic representations of queries and paths (or first-order logical rules) from the neural controller. See \citep{shen2017traversing,yang2017differentiable} for some interesting preliminary results of interpretation of neural methods.

\subsection{Reinforcement Learning based Methods}

DeepPath \citep{xiong2017deeppath}, MINERVA \citep{das2017go} and M-Walk \citep{shen2018reinforcewalk} are among the state-of-the-art methods that use RL for multi-step reasoning over a KB. They use a policy-based agent with continuous states based on KB embeddings to traverse the knowledge graph to identify the answer node (entity) for an input query. The RL-based methods are as robust as the neural methods due to the use of continuous vectors for state representation, and are as interpretable as symbolic methods because the agents explicitly traverse the paths in the graph.

We formulate KBR as an MDP defined by the tuple $(\mathcal{S}, \mathcal{A}, R, \mathcal{P})$, where $\mathcal{S}$ is the continuous state space, $\mathcal{A}$ the set of available actions, $\mathcal{P}$ the state transition probability matrix, and $R$ the reward function. Below, we follow M-Walk 
to describe these components in detail. We denote a KB as graph $\mathcal{G}(\mathcal{E}, \mathcal{R})$ which consists a collection of entity nodes $\mathcal{E}$ and the relation edges $\mathcal{R}$ that link the nodes. We denote a KB query as 
$q=(e_0,r,?)$, where $e_0$ and $r$ are the given source node and relation, respectively, and $?$ the answer node to be identified.

\paragraph{States} 

Let $s_t$ denote the state at time $t$, which encodes information of all traversed nodes up to $t$, all the previous selected actions and the initial query $q$. $s_t$ can be defined recursively as follows:
\begin{equation}
\begin{aligned}
s_0 &\defeq \{ q,\mathcal{R}_{e_0},\mathcal{E}_{e_0} \} , \\
s_t &= s_{t-1} \cup \{ a_{t-1},e_t,\mathcal{R}_{e_t},\mathcal{E}_{e_t} \}, 
\end{aligned}
\end{equation}
where 
$a_t \in \mathcal{A}$ is the action selected by the agent at time $t$, $e_t$ is the currently visited node, $\mathcal{R}_{e_t} \in \mathcal{R}$ is the set of all the edges connected to $e_t$, and $\mathcal{E}_{e_t} \in \mathcal{E}$ is the set of all the nodes connected to $e_t$. Note that in RL-based methods, $s_t$ is represented as a continuous vector using \eg, a RNN in M-Walk and MINERVA or a MLP in DeepPath.

\paragraph{Actions}

Based on $s_t$, the agent selects one of the following actions: (1) choosing an edge in $\mathcal{E}_{e_t}$ and moving to the next node $e_{t+1} \in \mathcal{E}$, or (2) terminating the reasoning process and outputting the current node $e_t$ as a prediction of the answer node $e_T$. 

\paragraph{Transitions}

The transitions are deterministic. As shown in \figref{fig:kbr-example}, once action $a_t$ is selected, the next node $e_{t+1}$ and its associated $\mathcal{E}_{e_{t+1}}$ and $\mathcal{R}_{e_{t+1}}$ are known.

\paragraph{Rewards}

We only have the terminal reward of $+1$ if $e_T$ is the correct answer, and $0$ otherwise. 

\paragraph{Policy Network}

The policy $\pi_\theta(a|s)$ denotes the probability of selecting action $a$ given state $s$, and is implemented as a neural network parameterized by $\theta$. The policy network is optimized to maximize $\mathbb{E}[V_\theta(s_0)]$, which is the long-term reward of starting from $s_0$ and following the policy $\pi_\theta$ afterwards.  In KBR, the policy network can be trained using RL, such as the REINFORCE method, from the training samples in the form of triples ${(e_s, r, e_t)}$ extracted from a KB. To address the reward sparsity issue (\ie, the reward is only available at the end of a path), \citet{shen2018reinforcewalk} proposed to use Monte Carlo Tree Search to generate a set of simulated paths with more positive terminal rewards by exploiting the fact that all the transitions are deterministic for a given knowledge graph.   

\section{Conversational KB-QA Agents}
\label{sec:conversational-kbqa}

All of the KB-QA methods we have described so far are based on \emph{single-turn} agents which assume that users can compose in one shot a complicated, compositional natural language query that can uniquely identify the answer in the KB. 

However, in many cases, it is unreasonable to assume that users can construct compositional queries without prior knowledge of the structure of the KB to be queried. Thus, conversational KB-QA agents are more desirable because they allow users to query a KB interactively without composing complicated queries. 



\begin{figure}[t] 
\centering 
\includegraphics[width=0.66\linewidth]{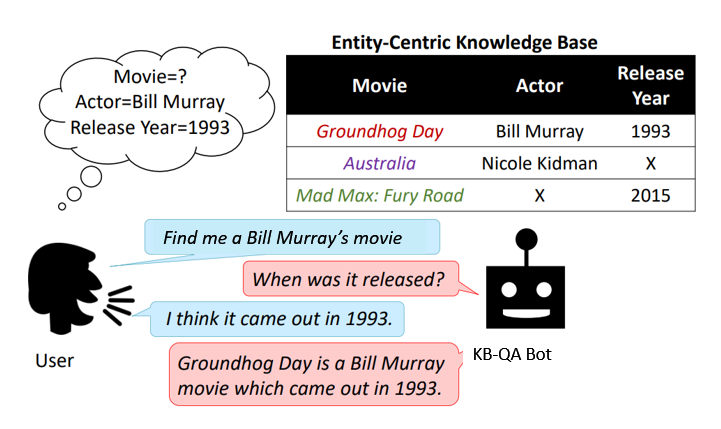}
\vspace{-2mm}
\caption{An interaction between a user and a multi-turn KB-QA agent for the movie-on-demand task. Figure credit: \citet{dhingra17towards}.} 
\label{fig:movie-on-demand-example} 
\vspace{0mm}
\end{figure}

A conversational KB-QA agent is useful for many interactive KB-QA tasks such as movie-on-demand, where a user attempts to find a movie based on certain attributes of that movie, as illustrated by the example in \figref{fig:movie-on-demand-example}, where the movie DB can be viewed as an entity-centric KB consisting of entity-attribute-value triples. 

\begin{figure}[t] 
\centering 
\includegraphics[width=0.66\linewidth]{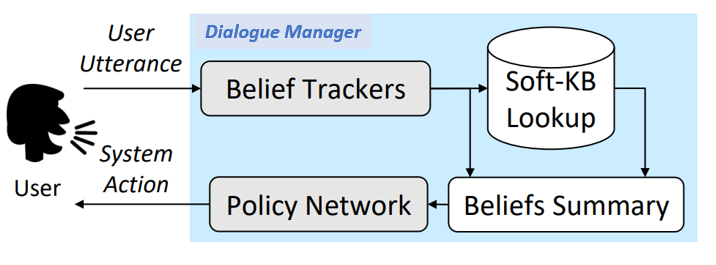}
\vspace{-2mm}
\caption{An overview of a conversational KB-QA agent. Figure credit:  \citet{dhingra17towards}.} 
\label{fig:multi-turn-kb-qa-agent-architecture} 
\vspace{0mm}
\end{figure}

In addition to the core KB-QA engine which typically consists of a semantic parser and a KBR engine, a conversational KB-QA agent is also equipped with a Dialogue Manager (DM) which tracks the dialogue state
and decides what question to ask to effectively help users navigate the KB in search of an entity (movie). The high-level architecture of the conversational agent for movie-on-demand is illustrated in \figref{fig:multi-turn-kb-qa-agent-architecture}.  At each turn, the agent receives a natural language utterance $u_t$ as input, and selects an action $a_t \in \mathcal{A}$ as output. The action space $\mathcal{A}$ consists of a set of questions, each for requesting the value of an attribute, and an action of informing the user with an ordered list of retrieved entities.  The agent is a typical task-oriented dialogue system of \figref{fig:two-dialogue-system} (Top), consisting of (1) a \emph{belief tracker} module for resolving coreferences and ellipsis in user utterances using conversation context,  identifying user intents, extracting associated attributes, and tracking the dialogue state; (2) an interface with the KB to query for relevant results (\ie, the \emph{Soft-KB Lookup} component, which can be implemented using the KB-QA models described in the previous sections, except that we need to form the query based on the dialogue history captured by the belief tracker, not just the current user utterance, as described in \citet{suhr2018learning}); (3) a \emph{beliefs summary} module to summarize the state into a vector; and (4) a \emph{dialogue policy} which selects the next action based on the dialogue state. The policy can be either programmed \citep{wu2015probabilistic} or trained on dialogues \citep{wen17network,dhingra17towards}. 

\citet{wu2015probabilistic} presented an Entropy Minimization Dialogue Management (EMDM) strategy. The agent always asks for the value of the attribute with maximum entropy over the remaining entries in the KB. EMDM is proved optimal in the absence of language understanding errors. However, it does not take into account the fact that some questions are easy for users to answer, whereas others are not. For example, in the movie-on-demand task, the agent could ask users to provide the movie release ID which is unique to each movie but is often unknown to regular users.

\citet{dhingra17towards} proposed KB-InfoBot -- a fully neural end-to-end multi-turn dialogue agent for the movie-on-demand task. The agent is trained entirely from user feedback. It does not suffer from the problem of EMDM, and always asks users easy-to-answer questions to help search in the KB.  Like all KB-QA agents, KB-InfoBot needs to interact with an external KB to retrieve real-world knowledge. This is traditionally achieved by issuing a symbolic query to the KB to retrieve entries based on their attributes. However, such symbolic operations break the differentiability of the system and prevent end-to-end training of the dialogue agent. KB-InfoBot addresses this limitation by replacing symbolic queries with an induced posterior distribution over the KB that indicates which entries the user is interested in. The induction can be achieved using the neural KB-QA methods described in the previous sections. Experiments show that integrating the induction process with RL leads to higher task success rate and reward in both simulations and against real users \footnote{It remains to be verified whether the method can deal with large-scale KBs with millions of entities.}.

\begin{figure}[t] 
\centering 
\includegraphics[width=1.0\linewidth]{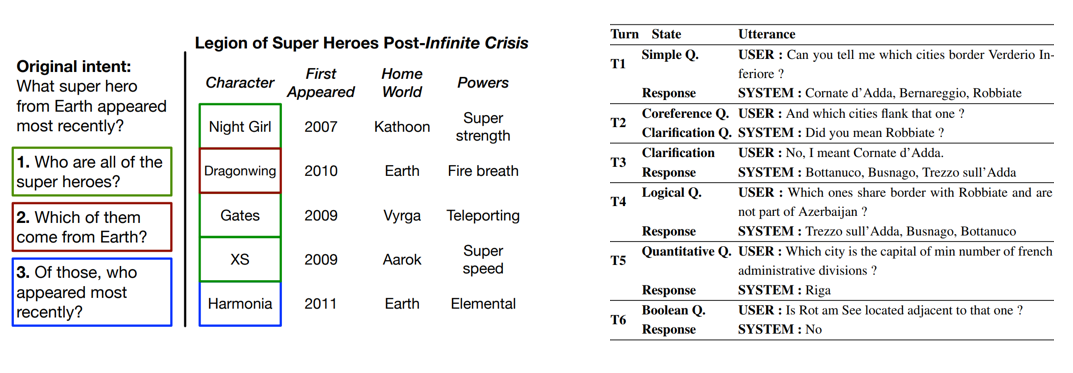}
\vspace{-2mm}
\caption{The examples from two conversational KB-QA datasets. (Left) An example question sequence created from a compositional question intent in the SQA dataset. Figure credit: \citet{iyyer2017search}. (Right)  An example dialogue from the CSQA dataset. Figure credit: \citet{saha2018complex}.} 
\label{fig:two-conversational-kb-qa-datasets}
\vspace{0mm}
\end{figure}

Recently, several datasets have been developed for building conversational KB-QA agents. \citet{iyyer2017search} collected a Sequential Question Answering (SQA) dataset via crowd sourcing by leveraging WikiTableQuestions (WTQ \citep{pasupat2015compositional}), which contains highly compositional questions associated with HTML tables from Wikipedia. As shown in the example in \figref{fig:two-conversational-kb-qa-datasets} (Left), each crowd sourcing task contains a long, complex question originally from WTQ as the question \emph{intent}. The workers are asked to compose a sequence of simpler but inter-related questions that lead to the final intent. The answers to the simple questions are subsets of the cells in the table. 

\citet{saha2018complex} presented a dataset consisting of 200K QA dialogues for the task of Complex Sequence Question Answering (CSQA). CSQA combines two sub-tasks: (1) answering factoid questions through complex reasoning over a large-scale KB, and (2) learning to converse through a sequence of coherent QA pairs. As the example in \figref{fig:two-conversational-kb-qa-datasets} (Right) shows, CSQA calls for a conversational KB-QA agent that combines many technologies described in this chapter, including 
(1) parsing complex natural language queries (\secref{sec:semantic-parsing-kb-qa}), 
(2) using conversation context to resolve coreferences and ellipsis in user utterances 
like the belief tracker in \figref{fig:multi-turn-kb-qa-agent-architecture}, (3) asking for clarification questions for ambiguous queries, like the dialogue manager in \figref{fig:multi-turn-kb-qa-agent-architecture}, and 
(4) retrieving relevant paths in the KB to answer questions (\secref{sec:multi-step-reasoning-kb}).
 

\section{Machine Reading for Text-QA}

Machine Reading Comprehension (MRC) is a challenging task: the goal is to have machines read a (set of) text passage(s) and then answer any question about the passage(s). The MRC model is the core component of text-QA agents. 

The recent big progress on MRC is largely due to the availability of a multitude of large-scale datasets that the research community has created over various text sources such as Wikipedia (WikiReading \citep{hewlett2016wikireading}, SQuAD \citep{rajpurkar2016squad}, WikiHop \citep{welbl2017constructing}, DRCD \citep{shao2018drcd}), news and other articles (CNN/Daily Mail \citep{hermann2015teaching}, NewsQA \citep{trischler2016newsqa}, RACE \citep{lai2017race}, ReCoRD \citep{zhang2018record}), fictional stories (MCTest \citep{richardson2013mctest}, CBT \citep{hill2015goldilocks}, NarrativeQA \citep{kovcisky2017narrativeqa}), 
science questions (ARC \citep{clark2018think}),
and general Web documents (MS MARCO \citep{nguyen2016ms}, TriviaQA \citep{joshi2017triviaqa}, SearchQA \citep{dunn2017searchqa}, DuReader \citep{he2017dureader}).

\begin{figure}[t] 
\centering 
\includegraphics[width=1.0\linewidth]{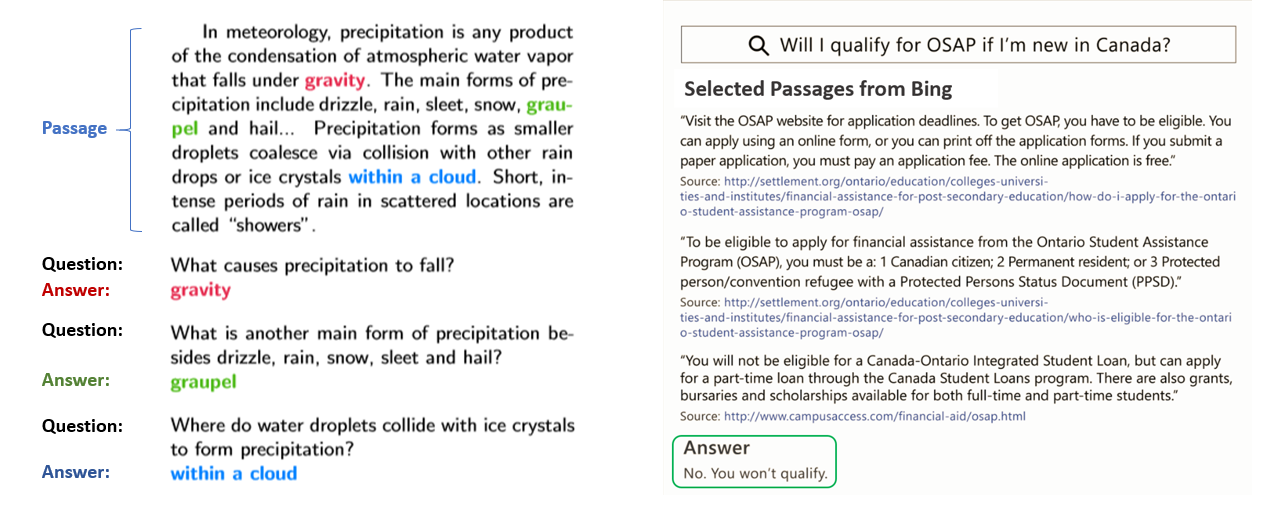}
\vspace{-2mm}
\caption{The examples from two MRC datasets. (Left) Question-answer pairs for a sample passage in the SQuAD dataset, adapted from \citet{rajpurkar2016squad}. Each of the answers is a text span in the passage. (Right) A question-answer pair for a set of passages in the MS MARCO dataset, adapted from \citet{nguyen2016ms}. The answer, if there is one, is human generated.} 
\label{fig:two-mrc-datasets}
\vspace{0mm}
\end{figure}

\paragraph{SQuAD.} This is the MRC dataset released by the Stanford NLP group. It consists of 100K questions posed by crowdworkers on a set of Wikipedia articles.
As shown in the example in \figref{fig:two-mrc-datasets} (Left), the MRC task defined on SQuAD involves a question and a passage, and aims to find an answer span in the passage. For example, in order to answer the question ``what causes precipitation to fall?'', one might first locate the relevant part of the passage ``precipitation ... falls under gravity'', then reason that ``under'' refers to a cause (not location), and thus determine the correct answer: ``gravity''. Although the questions with span-based answers are more constrained than the real-world questions users submit to Web search engines such as Google and Bing, SQuAD provides a rich diversity of question and answer types and became one of the most widely used MRC datasets in the research community. 

\paragraph{MS MARCO.} This is a large scale real-world MRC dataset, released by Microsoft, aiming to address the limitations of other academic datasets. For example, MS MARCO differs from SQuAD in that (1) SQuAD consists of the questions posed by crowdworkers while MS MARCO is sampled from the real user queries; (2) SQuAD uses a small set of high quality Wikipedia articles while MS MARCO is sampled from a large amount of Web documents, (3) MS MARCO includes some unanswerable queries{\footnote {SQuAD v2 \citep{rajpurkar2018know} also includes unanswerable queries.}} and (4) SQuAD requires identifying an answer span in a passage while MS MARCO requires generating an answer (if there is one) from multiple passages that may or may not be relevant to the given question. As a result, MS MARCO is far more challenging, and requires more sophisticated reading comprehension skills. As shown in the example in \figref{fig:two-mrc-datasets} (Right), given the question ``will I qualify for OSAP if I'm new in Canada'', one might first locate the relevant passage that includes: ``you must be a 1 Canadian citizen; 2 permanent resident; or 3 protected person...'' and reason that being new to the country is usually the opposite of being a citizen, permanent resident etc., thus determine the correct answer: ``no, you won't qualify''. 

In addition, TREC\footnote{\url{https://trec.nist.gov/data/qamain.html}} also provides a series of text-QA benchmarks:
\paragraph{The automated QA track.} This is one of the most popular tracks in TREC for many years, up to year 2007 \citep{dang2007overview,agichtein2015overview}. It has focused on the task of providing automatic answers for human questions. The track primarily dealt with factual questions, and the answers provided by participants were extracted from a corpus of News articles. While the task evolved to model increasingly realistic information needs, addressing question series, list questions, and even interactive feedback, a major limitation remained: the questions did not directly come from real users, in real time.

\paragraph{The LiveQA track.} This track started in 2015 \citep{agichtein2015overview}, focusing on answering user questions in real time. Real user questions, \ie, fresh questions submitted on the Yahoo Answers (YA) site that have not yet been answered, were sent to the participant systems, which provided an answer in real time. Returned answers were judged by TREC editors on a 4-level Likert scale. LiveQA revived this popular QA track which has been frozen for several years, attracting significant attention from the QA research community.

\section{Neural MRC Models}
\label{sec:neural-mrc-models}

\begin{figure}[t] 
\centering 
\includegraphics[width=1.00\linewidth]{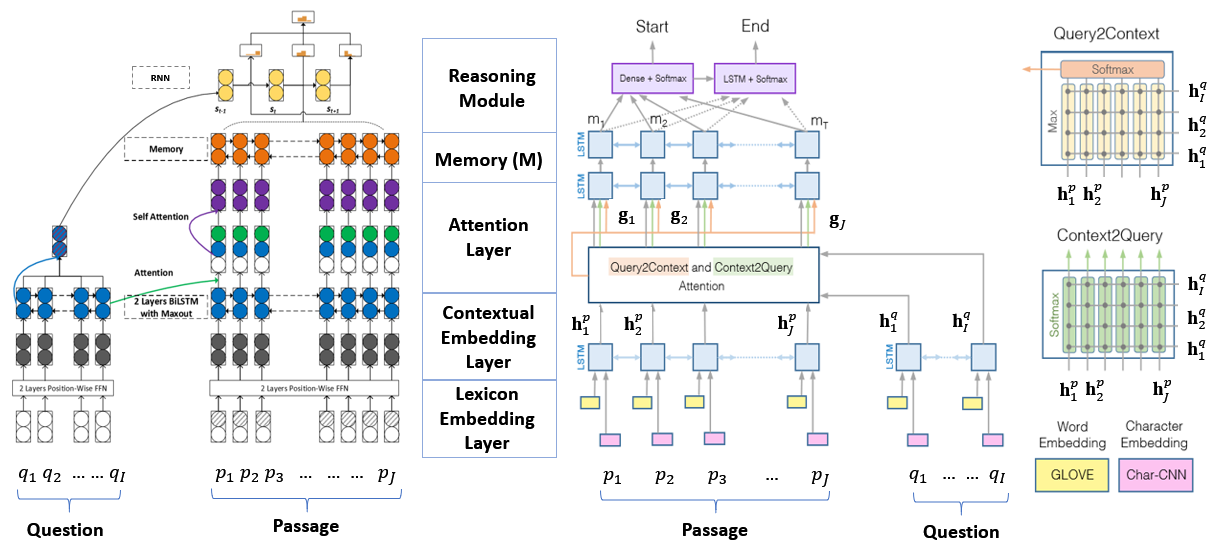}
\vspace{-2mm}
\caption{Two examples of state of the art neural MRC models. (Left) The Stochastic Answer Net (SAN) model. Figure credit: \citet{liu2018stochastic}. (Right) The BiDirectional Attention Flow (BiDAF) model. Figure credit: \citet{seo2016bidirectional}. } 
\label{fig:two-mrc-models}
\vspace{0mm}
\end{figure}

The description in this section is based on the state of the art models developed on SQuAD, where given a question $Q=(q_1,...,q_I)$ and a passage $P=(p_1,...,p_J)$, we need to locate an answer span $A=( a_{start},a_{end} )$ in $P$.

In spite of the variety of model structures and attention types \citep{chen2016thorough,xiong2016dynamic,seo2016bidirectional,shen2017empirical,wang2017gated}, a typical neural MRC model performs reading comprehension in three steps, as outlined in \figref{fig:symbolic-to-neural-shift}: (1) \emph{encoding} the symbolic representation of the questions and passages into a set of vectors in a neural space; (2) \emph{reasoning} in the neural space to identify the answer vector (\eg, in SQuAD, this is equivalent to ranking and re-ranking the embedded vectors of all possible text spans in $P$); and (3) \emph{decoding} the answer vector into a natural language output in the symbolic space (\eg, this is equivalent to mapping the answer vector to its text span in $P$). Since the decoding module is straightforward for SQuAD models, we will focus on encoding and reasoning below.

\figref{fig:two-mrc-models} illustrate two examples of neural MRC models.  BiDAF \citep{seo2016bidirectional} is among the most widely used state of the art MRC baseline models in the research community, and SAN \citep{liu2018stochastic} is the best documented MRC model on the SQuAD1.1 leaderboard\footnote{\url{https://rajpurkar.github.io/SQuAD-explorer/}} as of Dec. 19, 2017.


\subsection{Encoding}

Most MRC models encode questions and passages through three layers: a lexicon embedding layer, a contextual embedding layer, and an attention layer, as reviewed below.

\paragraph{Lexicon Embedding Layer.} 

This extracts information from $Q$ and $P$ at the word level and normalizes for lexical variants. It typically maps each word to a vector space using a pre-trained word embedding model, such as word2vec \citep{mikolov2013distributed} or GloVe \citep{pennington2014glove}, such that semantically similar words are mapped to the vectors that are close to each other in the neural space (also see \secref{sec:deep-learning-foundations}). Word embedding can be enhanced by concatenating each word embedding vector with other linguistic embeddings such as those derived from characters, Part-Of-Speech (POS) tags, and named entities etc. Given $Q$ and $P$, the word embeddings for the tokens in $Q$ is a matrix $\mathbf{E}^q \in \Rset^{d \times I}$ and tokens in $P$ is $ \mathbf{E}^p \in \Rset^{d \times J}$ , where $d$ is the dimension of word embeddings.

\paragraph{Contextual Embedding Layer.}

This utilizes contextual cues from surrounding words to refine the embedding of the words. As a result, the same word might map to different vectors in a neural space depending on its context, such as ``\emph{bank} of a river'' vs. `` \emph{bank} of America''. This is typically achieved by using a Bi-directional Long Short-Term Memory (BiLSTM) network,\footnote{Long Short-Term Memory (LSTM) networks are an extension for recurrent neural networks (RNNs). The units of an LSTM are used as building units for the layers of a RNN. LSTMs enable RNNs to remember their inputs over a long period of time because LSTMs contain their information in a gated cell, where gated means that the cell decides whether to store or delete information based on the importance it assigns to the information. The use of BiLSTM for contextual embedding is suggested by \citet{melamud2016context2vec,mccann2017learned}.} an extension of RNN of \figref{fig:rnn-example}. As shown in \figref{fig:two-mrc-models}, we place two LSTMs in both directions, respectively, and concatenate the outputs of the two LSTMs. Hence, we obtain a matrix
$\mathbf{H}^q \in \Rset^{2d \times I}$ as a contextually aware representation of $Q$ and a matrix
$\mathbf{H}^p \in \Rset^{2d \times J}$ as a contextually aware representation of $P$.

ELMo \citep{peters2018deep} is one of the state of the art contextual embedding models. It is based on deep BiLSTM. Instead of using only the output layer representations of BiLSTM, ELMo combines the intermediate layer representations in the BiLSTM, where the combination weights are optimized on task-specific training data. 

BERT \citep{devlin2018bert} differs from ELMo and BiLSTM in that it is designed to pre-train deep bidirection representations by \emph{jointly} conditioning on both left and right context in all layers. The pre-trained BERT representations can be fine-tuned with just one additional output layer to create state of the art models for a wide range of NLP tasks, including MRC. 

Since an RNN/LSTM is hard to train efficiently using parallel computing, \citet{yu2018qanet} presents a new contextual embedding model which does not require an RNN: Its encoder consists exclusively of convolution and self-attention, where convolution models local interactions and self-attention models global interactions. Such a model can be trained an order of magnitude faster than an RNN-based model on GPU clusters.

\paragraph{Attention Layer.}

This couples the question and passage vectors and produces a set of query-aware feature vectors for each word in the passage, and generates the working memory $\mathbf{M}$ over which reasoning is performed. This is achieved by summarizing information from both $\mathbf{H}^q$ and $\mathbf{H}^p$ via the \emph{attention} process\footnote{Interested readers may refer to Table 1 in \citet{huang2017fusionnet} for a summarized view on the attention process used in several state of the art MRC models.} 
that consists of the following steps:
%
\begin{enumerate}
\item Compute an attention score, which signifies which query words are most relevant to each passage word: $s_{ij} = \text{sim}_{\theta_s} (\mathbf{h}^q_i, \mathbf{h}^p_j) \in \Rset$ for each $\mathbf{h}^q_i$ in $\mathbf{H}^q$, where $\text{sim}_{\theta_s}$ is the similarity function \eg, a bilinear model, parameterized by $\theta_s$.
\item Compute the normalized attention weights through softmax: $ \alpha_{ij} = \exp(s_{ij}) / \sum_k \exp(s_{kj}) $.
\item Summarize information for each passage word via 
$\hat{\mathbf{h}}^p_j = \sum_i \alpha_{ij} \mathbf{h}^q_i$. Thus, we obtain a matrix $\hat{\mathbf{H}}^p \in \Rset^{2d \times J}$ as the question-aware representation of $P$. 
\end{enumerate}

Next, we form the working memory $\mathbf{M}$ in the neural space as 
$\mathbf{M} = f_{\theta}(\hat{\mathbf{H}}^p, \mathbf{H}^p)$, where $f_{\theta}$ is a  function of fusing its input matrices, parameterized by $\theta$.  $f_{\theta}$ can be an arbitrary trainable neural network. For example, the fusion function in SAN includes a concatenation layer, a self-attention layer and a BiLSTM layer. BiDAF computes attentions in two directions: from passage to question $\hat{\mathbf{H}}^q$ as well as from question to passage $\hat{\mathbf{H}}^p$. The fusion function in BiDAF includes a  layer that concatenates three matrices $\mathbf{H}^p$, $\hat{\mathbf{H}}^p$ and $\hat{\mathbf{H}}^q$, and a two-layer BiLSTM to encode for each word its contextual information with respect to the entire passage and the query. 


\subsection{Reasoning}

MRC models can be grouped into different categories based on how they perform reasoning to generate the answer.  Here, we distinguish single-step models from multi-step models.

\paragraph{Single-Step Reasoning.}

A single-step reasoning model matches the question and document only once and produces the final answers. 
We use the single-step version of SAN\footnote{This is a special version of SAN where the maximum number of reasoning steps $T=1$. SAN in \figref{fig:two-mrc-models} (Left) uses $T=3$.} in \figref{fig:two-mrc-models} (Left) as an example to describe the reasoning process. We need to find the answer span (\ie, the start and end points) over the working memory $\mathbf{M}$. First, a summarized question vector is formed as 

\begin{equation} \label{eqn:san-query}
\mathbf{h}^q = \sum_i \beta_i \mathbf{h}^q_i,  
\end{equation}
where $\beta_i = \exp(\mathbf{w}^\top \mathbf{h}^q_i) / \sum_{k} \exp(\mathbf{w}^\top \mathbf{h}^q_{k})$, and $\mathbf{w}$ is a trainable vector. Then, a bilinear function is used to obtain the probability distribution of the start index over the entire passage by 

\begin{equation} \label{eqn:san-start-index}
\mathbf{p}^{(start)} = \text{softmax} ({\mathbf{h}^q}^\top \mathbf{W}^{(start)} \mathbf{M}),  
\end{equation}
where $\mathbf{W}^{(start)}$ is a weight matrix. Another bilinear function is used to obtain the probability distribution of the end index, incorporating the information of the span start obtained by \eqnref{eqn:san-start-index}, as 

\begin{equation} \label{eqn:san-end-index}
\mathbf{p}^{(end)} = \text{softmax} ([{\mathbf{h}^q} ; \sum_j p^{(start)}_j \mathbf{m}_j]^\top \mathbf{W}^{(end)} \mathbf{M}),
\end{equation}
where the semicolon mark $;$ indicates the vector or matrix concatenation operator, $p^{(start)}_j$ is the probability of the $j$-th word in the passage being the start of the answer span, $\mathbf{W}^{(end)}$ is a weight matrix, and $\mathbf{m}_j$ is the $j$-th vector of $\mathbf{M}$.

\begin{figure}[t] 
\centering 
\includegraphics[width=1.0\linewidth]{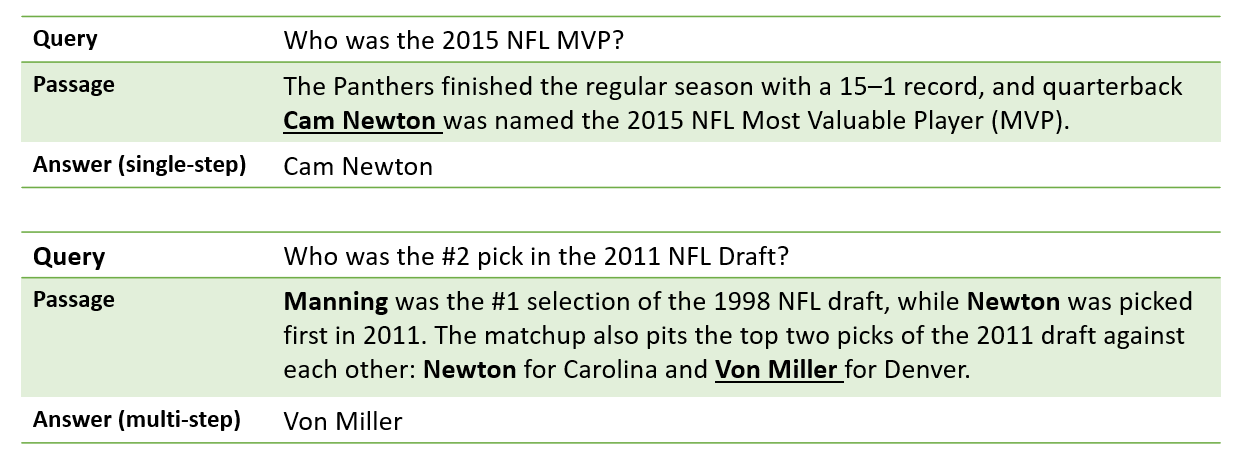}
\vspace{-2mm}
\caption{(Top) A human reader can easily answer the question by reading the passage only once. (Bottom) A human reader may have to read the passage multiple times to answer the question.} 
\label{fig:single-multi-step-reasoning-examples}
\vspace{0mm}
\end{figure}

Single-step reasoning is simple yet efficient and the model parameters can be trained using the classical back-propagation algorithm, thus it is adopted by most of the systems \citep{chen2016end,seo2016bidirectional,wang2017gated,liu2017phase,chen2017reading,weissenborn2017fastqa,hu2017mnemonic}. 
However, since humans often solve question answering tasks by re-reading and re-digesting the document multiple times before reaching the final answer (this may be based on the complexity of the questions and documents, as illustrated by the examples in \figref{fig:single-multi-step-reasoning-examples}), it is natural to devise an iterative way to find answers as multi-step reasoning.

\paragraph{Multi-Step Reasoning.}

Multi-step reasoning models are pioneered by \citet{hill2015goldilocks,dhingra2016gated,sordoni2016iterative,kumar2016ask}, who used a pre-determined fixed number of reasoning steps. \citet{shen2017reasonet,shen2017empirical} showed that multi-step reasoning outperforms single-step ones and dynamic multi-step reasoning further outperforms the fixed multi-step ones on two distinct MRC datasets (SQuAD and MS MARCO). But the dynamic multi-step reasoning models have to be trained using RL methods, \eg, policy gradient, which are tricky to implement due to the instability issue. SAN combines the strengths of both types of multi-step reasoning models. As shown in \figref{fig:two-mrc-models} (Left), SAN \citep{liu2018stochastic} uses a fixed number of reasoning steps, and generates a prediction at each step. During decoding, the answer is based on the average of predictions in all steps. During training, however, SAN drops predictions via \emph{stochastic dropout}, and generates the final result based on the average of the remaining predictions. Albeit simple, this technique significantly improves the robustness and overall accuracy of the model. Furthermore, SAN can be trained using back-propagation which is simple and efficient.

Taking SAN as an example, the multi-step reasoning module computes over $T$ memory steps and outputs the answer span. It is based on an RNN, similar to IRN in \figref{fig:multi-turn-kb-qa-agent-architecture}. It maintains a state vector, which is updated on each step. At the beginning, the initial state $\mathbf{s}_0$ is the summarized question vector computed by \eqnref{eqn:san-query}. At time step $t \in \{1,2,\ldots,T\}$, the state is defined by 
$\mathbf{s}_t = \text{RNN} (\mathbf{s}_{t-1}, \mathbf{x}_t)$, where
$\mathbf{x}_t$ contains retrieved information from memory using the previous state vector as a query via the attention process:
$\mathbf{M}$: $\mathbf{x}_t = \sum_j \gamma_j \mathbf{m}_j$ and
$\mathbf{\gamma} = \text{softmax}({\mathbf{s}_{t-1}}^\top \mathbf{W}^{(att)} \mathbf{M})$, where $\mathbf{W}^{(att)}$ is a trainable weight matrix. Finally, a bilinear function is used to find the start and end points of answer spans at each reasoning step $t$, similar to \eqnref{eqn:san-start-index} and \ref{eqn:san-end-index}:
\begin{align} 
  \label{eqn:san-start-index-t}
    &\mathbf{p}^{(start)}_t = \text{softmax} ({\mathbf{s}_{t}}^\top       \mathbf{W}^{(start)} \mathbf{M}), \\
  \label{eqn:san-end-index-t}
    &\mathbf{p}^{(end)}_t = \text{softmax} ([{\mathbf{s}_{t}} ; \sum_j p^{(start)}_{t,j} \mathbf{m}_j]^\top \mathbf{W}^{(end)} \mathbf{M}),
\end{align} 
where $p^{(start)}_{t,j}$ is the $j$-th value of the vector $\mathbf{p}^{(start)}_t$, indicating the probability of the $j$-th passage word being the start of the answer span at reasoning step $t$.

\subsection{Training}

A neural MRC model can be viewed as a deep neural network that includes all component modules (\eg, the embedding layers and reasoning engines) which by themselves are also neural networks. Thus, it can be optimized on training data in an end-to-end fashion via back-propagation and SGD, as outlined in \figref{fig:symbolic-to-neural-shift}. For SQuAD models, we optimize model parameters $\theta$ by minimizing the loss function defined as the sum of the negative log probabilities of the ground truth answer span start and end points by the predicted distributions, averaged over all training samples:

\begin{equation} \label{eqn:mrc-loss}
L(\theta) = - \frac{1}{|\mathcal{D}|} \sum_i^{|\mathcal{D}|} \left( \log \left( p^{(start)}_{y^{(start)}_i} \right) + \log \left( p^{(end)}_{y^{(end)}_i} \right) \right),
\end{equation}
where $\mathcal{D}$ is the training set, $y^{(start)}_i$ and $y^{(end)}_i$ are the true start and end of the answer span of the $i$-th training sample, respectively, and $p_k$ the $k$-th value of the vector $\mathbf{p}$.  

\section{Conversational Text-QA Agents}
\label{sec:conversational-text-qa}

While all the neural MRC models described in \secref{sec:neural-mrc-models} assume a single-turn QA setting, in reality, humans often ask questions in a conversational context \citep{ren2018conversational}. For example, a user might ask the question ``when was California founded?'', and then depending on the received answer, follow up by ``who is its governor?'' and ``what is the population?'', where both refer to ``California'' mentioned in the first question. This incremental aspect, although making human conversations succinct, presents new challenges that most state-of-the-art single-turn MRC models do not address directly, such as referring back to conversational history using coreference and pragmatic reasoning\footnote{Pragmatic reasoning is defined as ``the process of finding the intended meaning(s) of the
given, and it is suggested that this amounts to the process of inferring the appropriate context(s) in which to interpret the given'' 
\citep{bell1999pragmatic}. The analysis by \citet{jia2017adversarial,chen2016thorough} revealed that state of the art neural MRC models, \eg, developed on SQuAD, mostly excel at matching questions to local context via lexical matching and paragraphing, but struggle with questions that require reasoning.}
\citep{reddy2018coqa}. 

\begin{figure}[t] 
\centering 
\includegraphics[width=1.0\linewidth]{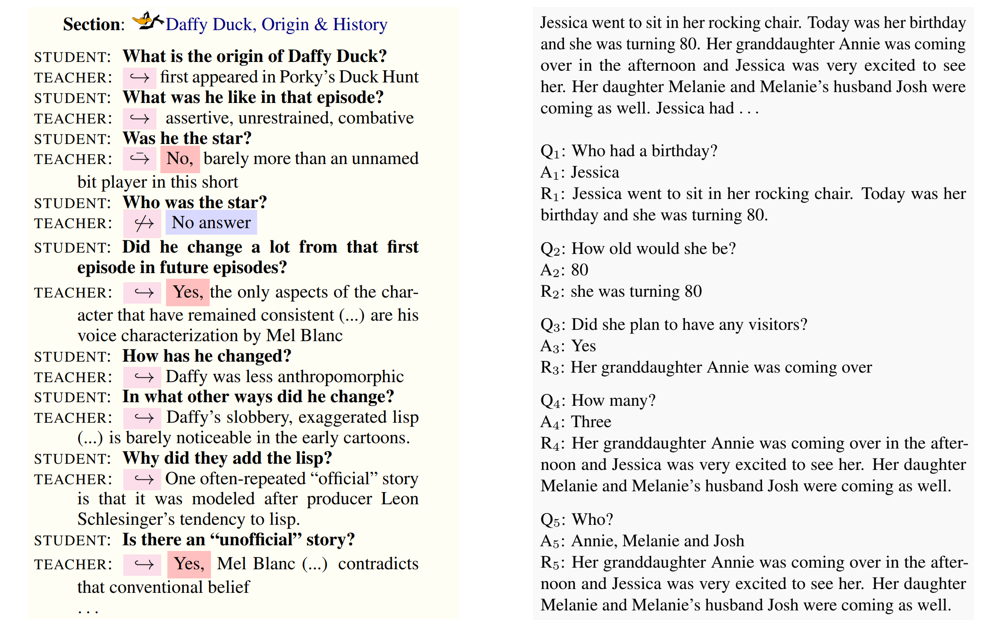}
\vspace{-2mm}
\caption{The examples from two conversational QA datasets. (Left) A QA dialogue example in the QuAC dataset.  The student, who does not see the passage (section text), asks questions. The teacher provides answers in the form of text spans and dialogue acts. These acts include (1) whether the student should $\hookrightarrow$, could 
$ \bar{\hookrightarrow} $, 
or should not 
$ \not\hookrightarrow $ 
ask a follow-up; (2) affirmation (Yes / No), and, when appropriate, (3) No answer. Figure credit: \citet{choi2018quac}. (Right) A QA dialogue example in the CoQA dataset.  Each dialogue turn contains a question ($\text{Q}_i$), an answer ($\text{A}_i$) and a rationale ($\text{R}_i$) that supports the answer. Figure credit: \citet{reddy2018coqa}.} 
\label{fig:two-conversational-mrc-datasets}
\vspace{0mm}
\end{figure}

A conversational text-QA agent uses a similar architecture to \figref{fig:multi-turn-kb-qa-agent-architecture}, except that the \emph{Soft-KB Lookup} module is replaced by a text-QA module which consists of a search engine (\eg, Google or Bing) that retrieves relevant passages for a given question, and an MRC model that generates the answer from the retrieved passages. The MRC model needs to be extended to address the aforementioned challenges in the conversation setting, henceforth referred to as a \emph{conversational MRC model}.

Recently, several datasets have been developed for building conversational MRC models.  Among them are CoQA (Conversational Question Answering \citep{reddy2018coqa}) and QuAC (Question Answering in Context \citep{choi2018quac}), as shown in \figref{fig:two-conversational-mrc-datasets}.  The task of conversational MRC is defined as follows. Given a passage $P$, the conversation history in the form of question-answer pairs $\{Q_1, A_1, Q_2, A_2,...,Q_{i-1}, A_{i-1} \}$ and a question $Q_i$, the MRC model needs to predict the answer $A_i$.  

A conversational MRC model extends the models described in \secref{sec:neural-mrc-models} in two aspects. First, the encoding module is extended to encode not only $P$ and $A_i$ but also the conversation history. Second, the reasoning module is extended to be able to generate an answer (via pragmatic reasoning) that might not overlap $P$. For example, \citet{reddy2018coqa} proposed a reasoning module that combines the text-span MRC model of DrQA \citep{chen2017reading} and the generative model of PGNet \citep{see2017get}. To generate a free-form answer, DrQA first points to the answer evidence in text (\eg, R5 in \figref{fig:two-conversational-mrc-datasets} (Right)), and PGNet generates the an answer  (\eg, A5) based on the evidence.




\chapter{Task-oriented Dialogue Systems}
\label{sec:dialogue}



This chapter focuses on task-oriented dialogue systems that assist users in solving a task.  Different from applications where the user seeks an answer or certain information (previous chapter), dialogues covered here are often for completing a task, such as making a hotel reservation or booking movie tickets.  Furthermore, compared to chatbots (next chapter), these dialogues often have a specific goal to achieve, and are typically domain dependent.

While task-oriented dialogue systems have been studied for decades, they have quickly gaining increasing interest in recent years, both in the research community and in industry.  This chapter focuses on the foundation and algorithmic aspects, while industrial applications are discussed in \chref{sec:commercial}.  Furthermore, we restrict ourselves to dialogues where user input is in the form of raw text, not spoken language, but many of the techniques and discussions in this chapter can be adapted to spoken dialogues systems.

The chapter is organized as follows.  It starts with an overview of basic concepts, terminology, and a typical architecture for task-oriented dialogue systems.  Second, it reviews several representative approaches to dialogue system evaluation.  This part is critical in the development cycle of dialogue systems, but is largely orthogonal to the concrete techniques used to build them.
The next three sections focus on each of three main components in a typical dialogue system, with an emphasis on recent, neural approaches.  Finally, we review several recent works on end-to-end dialogue systems, which are enabled by recent advances at the intersection of deep learning and reinforcement learning, with further discussions and pointers provided in the last section.

\section{Overview}
\label{sec:dialogue:overview}

We start with a brief overview of task-oriented dialogue systems, focusing on aspects that facilitate later discussions.  For more information and historical developments, readers are referred to the textbook of \citet[Chapters~25~\&~26]{jurafsky18speech}.

\subsection*{Slot-filling Dialogues}

Throughout the chapter, we focus on a relatively simple yet important class of dialogues that involve filling in values for a predefined set of \emph{slots} before a task can be successfully completed.
These dialogues, known as \emph{slot-filling} or \emph{form-filling}, find a number of uses in practice. 
\tabref{tab:movie-example} gives an example conversation between a user and a dialogue system.  To successfully complete a transaction to book tickets, the system must collect necessary information by conversing with the user.
\begin{table}[ht]
\caption{An example movie-booking dialogue, adapted from \citet{li16user}.} \label{tab:movie-example}
\small
\begin{tabular}{p{0.1\textwidth} | p{0.82\textwidth}}
\hline
System & Hello!  How may I assist you? \\
User & Can I get tickets for Zoolander 2 tomorrow? \\
System & Sure!  Which city? \\
User & I want to watch it in Seattle. \\
System & How many tickets do you need? \\
User & Two, please.  And I'd like to watch at Regal Meridian 16. \\
System & 9:25 pm is available at Regal Meridian 16.  Does it work for you?  \\
User & Well, yes. \\
System & Okay, I was able to book 3 tickets for Zoolander 2 tomorrow at Regal Meridian 16 theater in Seattle at 9:25 pm. \\
User & Thank you. \\
System & Thank you.  Good bye! \\
\hline
\end{tabular}
\end{table}

Slot-filling dialogue may be applied to many other \emph{domains}, such as movie (as shown in the example above), restaurant, air ticket booking, etc.  For each domain, a set of \emph{slots} are defined by domain experts and are application specific.  For example, in the movie domain, slots like \texttt{movie\_name}, \texttt{theater\_name}, \texttt{time}, \texttt{date}, \texttt{ticket\_price}, \texttt{num\_tickets}, etc. are necessary.

Finally, a slot is called \emph{informable} if the value for this slot can be used to constrain the conversation, such as \texttt{phone\_number}; a slot is called \emph{requestable} if the speaker can ask for its value, such as \texttt{ticket\_price}.  Note that a slot can be both informable and requestable, an example being \texttt{movie\_name}.

\subsection*{Dialogue Acts}

The interaction between a dialogue agent and a user, as shown in the previous example, mirrors the interaction between an RL agent and the environment (\figref{fig:agent-env}), where a user utterance is the observation, and the system utterance is the action selected by the dialogue agent.  The dialogue acts theory gives a formal foundation for this intuition~\citep{core97coding,traum99speech}.
%

In this framework, the utterances of a user or agent are considered  actions that can change the (mental) state of both the user and the system, thus the state of the conversation.  These actions can be used to suggest, inform, request certain information, among others.  A simple example dialogue act is \texttt{greet}, which corresponds to natural language sentences like ``Hello! How may I assist you?''.  It allows the system to greet the user and start a conversation. Some dialogue acts may have slots or slot-value pairs as arguments.  For example, the following question in the movie-booking example above:
\[
\text{``How many tickets do you need?''}
\]
is to request information about a certain slot:
\[
\texttt{request(num\_tickets)},
\]
while the following sentence
\[
\text{``I want to watch it in Seattle.''}
\]
is to inform the city name:
\[
\texttt{inform(city=``seattle'')}.
\]
In general, dialogue acts are domain specific.  Therefore, the set of dialogue acts in a movie domain, for instance, will be different from that in the restaurant domain~\citep{schatzmann09hidden}.


\subsection*{Dialogue as Optimal Decision Making}

Equipped with dialogue acts, we are ready to model multi-turn conversations between a dialogue agent and a user as an RL 
problem.  Here, the dialogue system is the RL agent, and the user is the environment.  At every turn of the dialogue,
\begin{itemize}
\item{the agent keeps track of the dialogue state, based on information revealed so far in the conversation, and then takes an action; the action may be a response to the user in the form of dialogue acts, or an internal operation such as a database lookup or an API call;}
\item{the user responds with the next utterance, which will be used by the agent to update its internal dialogue state in the next turn;}
\item{an immediate reward is computed to measure the quality and/or cost for this turn of conversation.}
\end{itemize}

This process is precisely the agent-environment interaction discussed in \secref{sec:basics:rl}.  We now discuss how a reward function is determined.

An appropriate reward function should capture desired features of a dialogue system.  In task-oriented dialogues, we would like the system to succeed in helping the user in as few turns as possible.  Therefore, it is natural to give a high reward (say $+20$) at the end of the conversation if the task is successfully solved, or a low reward (say $-20$) otherwise. Furthermore, we may give a small penalty (say, $-1$ reward) to every intermediate turn of the conversation, so that the agent is encouraged to make the dialogue as short as possible.  The above is of course just a simplistic illustration of how to set a reward function for task-oriented dialogues, but in practice more sophisticated reward functions may be used, such as those that measure diversity and coherence of the conversation.  Further discussion of the reward function can be found in Sections~\ref{sec:dialogue:reward}, \ref{sec:dialogue:evaluation:metrics} and  \ref{sec:beyond-supervised-learning}.

\begin{figure}
\centering
\includegraphics[width=0.95\textwidth]{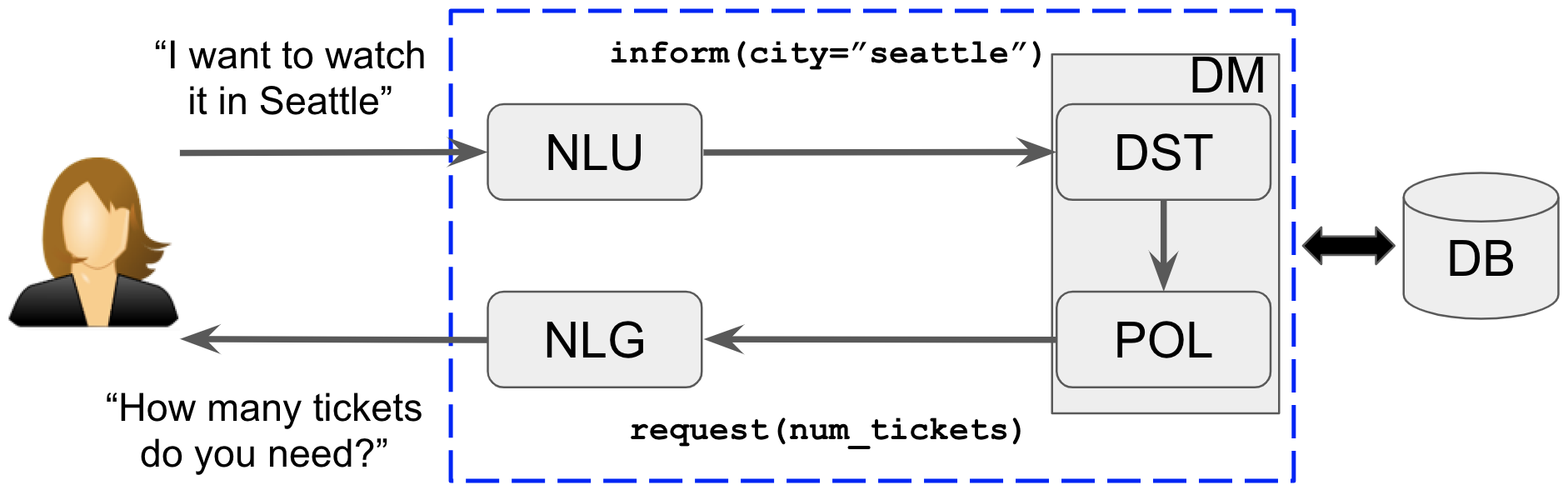}
\caption{An architecture for multi-turn task-oriented dialogues.  It consists of the following modules: NLU (Natural Language Understanding), DM (Dialogue Manager), and NLG (Natural Language Generation).  DM contains two sub-modules, DST (Dialogue State Tracker) and POL (Dialogue Policy).  The dialogue system, indicated by the dashed rectangle, may have access to an external database (DB).} \label{fig:dialogue-arch}
\end{figure}

To build a system, the pipeline architecture depicted in \figref{fig:dialogue-arch} is often used in practice.  It consists of the following modules.
\begin{itemize}
\item{Natural Language Understanding (NLU): This module takes the user's raw utterance as input and converts it to the semantic form of dialogue acts.}
\item{Dialogue Manager (DM): This module is the central controller of the dialogue system.  It often has a Dialogue State Tracking (DST) sub-module that is responsible for keeping track of the current dialogue state.  The other sub-module, the policy, relies on the internal state provided by DST to select an action.  Note that an action can be a response to the user, or some operation on backend databases (e.g., looking up certain information).}
\item{Natural Language Generation (NLG): If the policy chooses to respond to the user, this module will convert this action, often a dialogue act, into a natural language form.}
\end{itemize}

\subsection*{Dialogue Manager}


There is a huge literature on building (spoken) dialogue managers.  A comprehensive survey is out of the scope of the this chapter.  Interested readers are referred to some of the earlier examples~\citep{cole99tools,
larsson00information,
rich01collagen,allen01toward,
bos03dipper,
bohus09ravenclaw}, as well as excellent surveys like \citet{mctear02spoken}, \citet{paek08automating}, and \citet{young2013pomdp} for more information.  Here, we review a small subset of traditional approaches from the decision-theoretic view we take in this paper.

\citet{levin00stochastic} viewed conversation as a decision making problem.  \citet{walker00application} and \citet{singh02optimizing} are two early applications of reinforcement learning to manage dialogue systems.  While promising, these approaches assumed that the dialogue state can only take finitely many possible values, and is fully observable (that is, the DST is perfect).  Both assumptions are often violated in real-world applications, given ambiguity in user utterance and unavoidable errors in NLU.

To handle uncertainty inherent in dialogue systems, \citet{roy00spoken} and \citet{williams07partially} proposed to use Partially Observable Markov Decision Process (POMDP) as a principled mathematical framework for modeling and optimizing dialogue systems.  The idea is to take user utterances as observations to maintain a posterior distribution of the unobserved dialogue state; the distribution is sometimes referred to as the ``belief state.'' 
Since exact optimization in POMDPs is computationally intractable, authors have studied approximation techniques~\citep{roy00spoken,williams07partially,
young10hidden,li09reinforcement,gasic14gaussian} and alternative representations such as the information states framework~\citep{larsson00information,daubigney12comprehensive}.  Still, compared to the neural approaches covered in later sections, these methods often require more domain knowledge to engineer features and design states. 

Another important limitation of traditional approaches is that each module in \figref{fig:dialogue-arch} is often optimized separately.  Consequently, when the system does not perform well, it can be challenging to solve the ``credit assignment'' problem, namely, to identify which component in the system causes undesired system response and needs to be improved.  Indeed, as argued by \citet{mctear02spoken}, ``[t]he key to a successful dialogue system is the integration of these components into a working system.''  The recent marriage of differentiable neural models and reinforcement learning allows a dialogue system to be optimized in an end-to-end fashion, potentially leading to higher conversation quality; see \secref{sec:dialogue:e2e} for further discussions and recent works on this topic.

\section{Evaluation and User Simulation}
\label{sec:dialogue:evaluation}

Evaluation has been an important research topic for dialogue systems.  Different approaches have been used, including corpus-based approaches, user simulation, lab user study, actual user study, etc.  We will discuss pros and cons of these various methods, and in practice trade-offs are made to find the best option or a combination of them.

\subsection{Evaluation Metrics}
\label{sec:dialogue:evaluation:metrics}

While individual components in a dialogue system can often be optimized against more well-defined metrics such as accuracy, precision, recall, F1 and BLEU scores, evaluating a whole dialogue system requires a more holistic view and is more challenging~\citep{walker97paradise,walker98evaluating,walker00towards,paek01empirical,hartikainen04subjective}.   In the reinforcement-learning framework, it implies that the reward function has to take multiple aspects of dialogue quality into consideration.  In practice, the reward function is often a weighted linear combination of a subset of the following metrics.

The first class of metrics measures \emph{task completion success}.  The most common choice is perhaps \emph{task success rate}---the fraction of dialogues that successfully solve the user's problem (buying the right movie tickets, finding proper restaurants, etc.).  Effectively, the reward corresponding to this metric is $0$ for every turn, except for the last turn where it is $+1$ for a successful dialogue and $-1$ otherwise.  Many examples are found in the literature~\citep{walker97paradise,williams06pPartially,peng17composite}.   
Other variants have also been used, such as those to measure partial success~\citep{singh02optimizing,young16evaluation}.

The second class measures cost incurred in a dialogue, such as time elapsed.  A simple yet useful example is the number of turns, which reflects the intuition that a more succinct dialogue is preferred with everything else being equal.  The reward is simply $-1$ per turn, although more complicated choices exist~\citep{walker97paradise}.

In addition, other aspects of dialogue quality may also be encoded into the reward function, although this is a relatively under-investigated direction.  In the context of chatbots (\chref{sec:chitchat}), coherence, diversity and personal styles have been used to result in more human-like dialogues~\citep{li2015diversity,li2016persona}.  They can be useful for task-oriented dialogues as well.  In \secref{sec:dialogue:reward}, we will review a few recent works that aim to learn reward functions automatically from data.


\subsection{Simulation-Based Evaluation}
\label{sec:dialogue:evaluation:usersim}


Typically, an RL algorithm needs to interact with a user to learn (\secref{sec:basics:rl}). But running RL on either recruited users or actual users can be expensive.  A natural way to get around this challenge is to build a \emph{simulated} user, with which an RL algorithm can interact at virtually no cost.  Essentially, a simulated user tries to mimic what a real user does in a conversation: it keeps track of the dialogue state, and converses with an RL dialogue system.

Substantial research has gone into building realistic user simulators~\citep{schatzmann05quantitative,georgila06user,pietquin06probabilistic,pietquin13survey}.  There are many different dimensions to categorize a user simulator, such as deterministic vs. stochastic, content-based vs. collaboration-based, static vs. non-static user goals during  the conversations, among others.  Here, we highlight two dimensions, and refer interested users to \citet{schatzmann06survey} for further details on creating and evaluating user simulators~:

\begin{itemize}
\item{Along the \emph{granularity} dimension, the user simulator can operate either at the dialogue-act level (also known as intention level), or at the utterance level~\citep{jung09data}.}
\item{Along the \emph{methodology} dimension, the user simulator can be implemented using a rule-based approach, or a model-based approach with the model learned from a real conversational corpus.}
\end{itemize}

\paragraph{Agenda-Based Simulation.}
As an example, we describe a popular hidden agenda-based user simulator developed by \citet{schatzmann09hidden}, as instantiated in \citet{li16user} and \citet{ultes17pydial}.  Each dialogue simulation starts with a randomly generated user goal that is unknown to the dialogue manager.  In general the user goal consists of two parts: the \texttt{inform}-slots 
contain a number of slot-value pairs that serve as constraints the user wants to impose on the dialogue; the \texttt{request}-slots  
are slots whose values are initially unknown to the user and will be filled out during the conversation.  \figref{fig:user-goal-example} shows an example user goal in a movie domain, in which the user is trying to buy $3$ tickets for tomorrow for the movie \texttt{batman vs. superman}.

\begin{figure}[t]
\begin{center}
\small
\begin{verbatim}
                {
                  request_slots:
                  {
                    ticket: UNK
                    theater: UNK
                    start_time: UNK
                  },
                  inform_slots:
                  {
                    number_of_people: 3
                    date: tomorrow
                    movie_name: batman vs. superman
                  }
                }
\end{verbatim}
\end{center}
\caption{An example user goal in the movie-ticket-booking domain}
\label{fig:user-goal-example}
\end{figure}

Furthermore, to make the user goal more realistic, domain-specific constraints are added, so that certain slots are required to appear in the user goal.  For instance, it makes sense to require a user to know the number of tickets she wants in the movie domain.

During the course of a dialogue, the simulated user maintains a stack data structure known as \emph{user agenda}.  Each entry in the agenda corresponds to a pending intention the user aims to achieve, and their priorities are implicitly determined by the first-in-last-out operations of the agenda stack.  In other words, the agenda provides a convenient way of encoding the history of conversation and the ``state-of-mind'' of the user.  Simulation of a user boils down to how to maintain the agenda after each turn of the dialogue, when more information is revealed.  Machine learning or expert-defined rules can be used to set parameters in the stack-update process. 

\paragraph{Model-based Simulation.}

Another approach to building user simulators is entirely based on data~\citep{eckert97user,levin00stochastic,chandramohan11user}.  Here, we describe a recent example due to \citet{asri16sequence}.  Similar to the agenda-based approach, the simulator also starts an episode with a randomly generated user goal and constraints.  These are fixed during a conversation.

In each turn, the user model takes as input a sequence of contexts collected so far in the conversation, and outputs the next action.  Specifically, the context at a turn of conversation consists of:
\begin{itemize}
\item{the most recent machine action, }
\item{inconsistency between machine information and user goal, }
\item{constraint status, and}
\item{request status.}
\end{itemize}
With these contexts, an LSTM or other sequence-to-sequence models are used to output the next user utterance.  The model can be learned from human-human dialogue corpora.  In practice, it often works well by combining both rule-based and model-based techniques to create user simulators.

\paragraph{Further Remarks on User Simulation.}
While there has been much work on user simulation, building a human-like simulator remains challenging.  In fact, even user simulator evaluation itself continues to be an ongoing research topic~\citep{williams08evaluating,ai08assessing,pietquin13survey}.  In practice, it is often observed that dialogue policies that are overfitted to a particular user simulator may not work well when serving another user simulator or real humans~\citep{schatzmann05effects,dhingra17towards}.  The gap between a user simulator and humans is the major limitation of user simulation-based dialogue policy optimization.

Some user simulators are publicly available for research purposes.  Other than the aforementioned agenda-based simulators by \citet{li16user,ultes17pydial}, a large corpus with an evaluation environment, called AirDialogue (in the flight booking domain), was recently made available~\citep{wei18airdialogue}. At the IEEE workshop on Spoken Language Technology in 2018, Microsoft organized a dialogue challenge\footnote{\url{https://github.com/xiul-msr/e2e_dialog_challenge}} of building end-to-end task-oriented dialogue systems by providing an experiment platform with built-in user simulators in several domains \citep{li2018microsoft}.  

\subsection{Human-based Evaluation}

Due to the discrepancy between simulated users and human users, it is often necessary to test a dialogue system on human users to reliably evaluate its quality.  There are roughly two types of human users.

The first is human subjects recruited in a lab study, possibly through crowd-sourcing platforms.  Typically, the participants are asked to test-use a dialogue system to solve a given task (depending on the domain of the dialogues), so that a collection of dialogues are obtained.  Metrics of interest such as task-completion rate and average turns per dialogue can be measured, as done with a simulated user.  In other cases, a fraction of these subjects are asked to test-use a baseline dialogue system, so that the two can be compared against various metrics.


Many published studies involving human subjects are of the first type~\citep{walker00application,singh02optimizing,ai07comparing,rieser11learning,gasic13online,wen2015semantically,young16evaluation,peng17composite,lipton18bbq}.  While this approach has benefits over simulation-based evaluation, it is rather expensive and time-consuming to get a large number of subjects that can participate for a long time.  Consequently, it has the following limitations:
\begin{itemize}
\item{The small number of subjects prevents detection of statistically significant yet numerically small differences in metrics, often leading to inconclusive results.}
\item{Only a very small number of dialogue systems may be compared.}
\item{It is often impractical to run an RL agent that learns by interacting with these users, except in relatively simple dialogue applications.}
\end{itemize}

The other type of humans for dialogue system evaluation is \emph{actual} users (\eg, \citet{black11spoken}).  They are similar to the first type of users, except that they come with their actual tasks to be solved by conversing with the system.  Consequently, metrics evaluated on them are even more reliable than those computed on recruited human subjects with artificially generated tasks.  Furthermore, the number of actual users can be much larger, thus resulting in greater flexibility in evaluation.  In this process, many online and offline evaluation techniques such as A/B-testing and counterfactual estimation can be used~\citep{hofmann16online}.  The major downside of experimenting with actual users is the risk of negative user experience and disruption of normal services.

\subsection{Other Evaluation Techniques}

Recently, researchers have started to investigate a different approach to evaluation that is inspired by the self-play technique in RL~\citep{tesauro95temporal,mnih15human}.   This technique is typically used in a two-player game (such as the game of Go), where both players are controlled by the same RL agent, possibly initialized differently.  By playing the agent against itself, a large amount of trajectories can be generated at relatively low cost, from which the RL agent can learn a good policy.

Self-play must be adapted to be used for dialogue management, as the two parties involved in a conversation often play asymmetric roles (unlike in games such as Go).  \citet{shah18bootstrapping} described such a {dialogue self-play} procedure, which can generate conversations between a simulated user and the system agent.  Promising results have been observed in negotiation dialogues~\citep{lewis17deal} and task-oriented dialogues~\citep{liu17iterative,shah18bootstrapping,wei18airdialogue}.  It provides an interesting solution to avoid the evaluation cost of involving human users as well as overfitting to untruthful simulated users.

In practice,  it is reasonable to have a hybrid approach to evaluation.  One possibility is to start with simulated users, then validate or fine-tune the dialogue policy on human users (\cf, \citet{shah18bootstrapping}).  Furthermore, there are more systematic approaches to using both sources of users for policy learning (see \secref{sec:dialogue:deep-dyna}).

\section{Natural Language Understanding and Dialogue State Tracking}
\label{sec:dialogue:nlu-dst}

NLU and DST are two closely related components essential to a dialogue system.  They can have a significant impact on the overall system's performance (see, e.g., \citet{li17investigation}).  This section reviews some of the classic and state-of-the-art approaches.

\subsection{Natural Language Understanding}

\begin{figure}
\centering
\includegraphics[width=0.8\columnwidth]{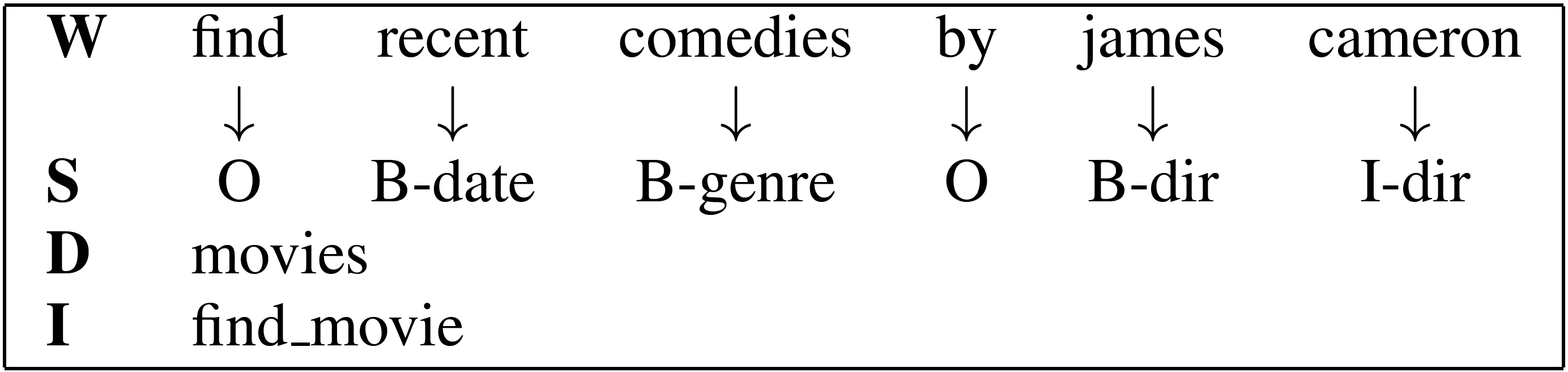}
\caption{An example output of NLU, where the utterance (W) is used to predict domain (I), intent (I), and the slot tagging (S).  The IOB representation is used.  Figure credit: \citet{hakkanitur16multi}.} \label{fig:iob-example}
\end{figure}

The NLU module takes user utterance as input, and performs three tasks: domain detection, intent determination, and slot tagging.  An example output for the three tasks is given in \figref{fig:iob-example}.  Typically, a pipeline approach is taken, so that the three tasks are solved one after another.  Accuracy and F1 score are two of the most common metrics used to evaluate a model's prediction quality.  NLU is a pre-processing step for later modules in the dialogue system, whose quality has a significant impact on the system's overall quality~\citep{li17end}.

Among them, the first two tasks are often framed as a classification problem, which infers the domain or intent (from a predefined set of candidates) based on the current user utterance~\citep{schapire00boostexter,yaman08integrative,sarikaya14application}.   Neural approaches to multi-class classification have been used in the recent literature and outperformed traditional statistical methods.  \citeauthor{ravuri15recurrent}~(\citeyear{ravuri15recurrent}; \citeyear{ravuri16comparative}) studied the use of standard recurrent neural networks, and found them to be more effective. 
For short sentences where information has to be inferred from the context, \citet{lee16sequential} proposed to use recurrent and convolutional neural networks that also consider texts prior to the current utterance.  Better results were shown on several benchmarks.

The more challenging task of slot tagging is often treated as sequence classification, where the classifier predicts semantic class labels for subsequences of the input utterance~\citep{wang05spoken,mesnil13investigation}.  \figref{fig:iob-example} shows an ATIS (Airline Travel Information System) utterance example in the Inside-Outside-Beginning (IOB) format~\citep{ramshaw95text}, where for each word the model predicts a semantic tag.

\citet{yao13recurrent} and \citet{mesnil15using} applied recurrent neural networks to slot tagging, where inputs are one-hot encoding of the words in the utterance, and obtained higher accuracy than statistical baselines such as conditional random fields and support vector machines.  Moreover, it is also shown that a-prior word information can be effectively incorporated into basic recurrent models to yield further accuracy gains.

\begin{figure}
    \centering
\includegraphics[width=0.35\linewidth]{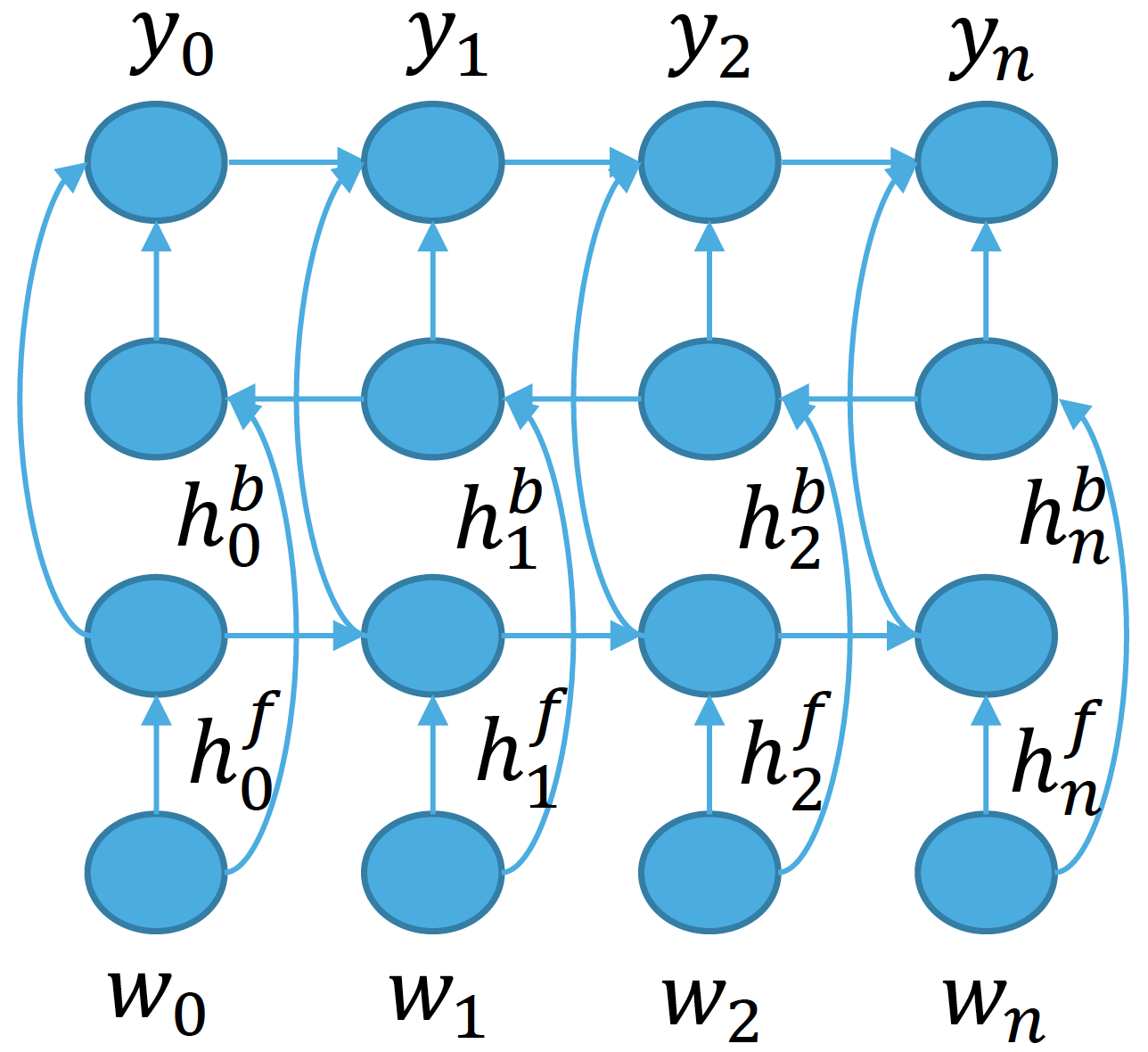}
    \caption{A bLSTM model for joint optimization in NLU.  Picture credit: \citet{hakkanitur16multi}.}
    \label{fig:blstm-nlu}
\end{figure}

As an example, this section describes the use of bidirectional LSTM~\citep{graves05framewise}, or bLSTM in short, in NLU tasks, following \citet{hakkanitur16multi} who also discussed other models for the same tasks.  The model, as shown in \figref{fig:blstm-nlu}, uses two sets of LSTM cells applied to the input sequence (the forward) and the \emph{reversed} input sequence (the backward).  The concatenated hidden layers of the forward and backward LSTMs are used as input to another neural network to compute the output sequence.  Mathematically, upon the $t^{\mathrm{th}}$ input token, $\mathbf{w}_t$, operations of the forward part of bLSTM are defined by the following set of equations:
\begin{align*}
\mathbf{i}_t &= g(\mathbf{W}_{wi} \mathbf{w}_t + \mathbf{W}_{hi} \mathbf{h}_{t-1}) \\
\mathbf{f}_t &= g(\mathbf{W}_{wf} \mathbf{w}_t + \mathbf{W}_{hf} \mathbf{h}_{t-1}) \\
\mathbf{o}_t &= g(\mathbf{W}_{wo} \mathbf{w}_t + \mathbf{W}_{ho} \mathbf{h}_{t-1}) \\
\hat{\mathbf{c}}_t &= \operatorname{tanh}(\mathbf{W}_{wc} \mathbf{w}_t + \mathbf{W}_{hc} \mathbf{h}_{t-1}) \\
\mathbf{c}_t &= \mathbf{f}_t \odot \mathbf{c}_{t-1} + \mathbf{i}_t \odot \hat{\mathbf{c}}_t \\
\mathbf{h}_t &= \mathbf{o}_t \odot \operatorname{tanh}(\mathbf{c}_t) \,,
\end{align*}
where $\mathbf{h}_{t-1}$ is the hidden layer, $\mathbf{W}_{\star}$ the trainable parameters, and $g(\cdot)$ the sigmoid function.  As in standard LSTMs, $\mathbf{i}_t$, $\mathbf{f}_t$ and $\mathbf{o}_t$ are the input, forget, and output gates, respectively.  The backward part is similar, with the input reversed.

To predict the slot tags as shown in \figref{fig:iob-example}, the input $\mathbf{w}_t$ is often a one-hot vector of a word embedding vector.  The output upon input $\mathbf{w}_t$ is predicted according to the following distribution $p_t$:
\[
p_t = \operatorname{softmax}(\mathbf{W}_{hy}^{(f)} \mathbf{h}_t^{(f)} + \mathbf{W}_{hy}^{(b)} \mathbf{h}_t^{(b)})\,,
\]
where the superscripts, $(f)$ and $(b)$, denote forward and backward parts of the bLSTM, respectively.  For tasks like domain and intent classification, the output is predicted at the end of the input sequence, and simpler architectures may be used~\citep{ravuri15recurrent,ravuri16comparative}.

In many situations, the present utterance alone can be ambiguous or lack all necessary information.  Contexts that include information from previous utterances are expected to help improve model accuracy.  \citet{hori15context} treated conversation history as a long sequence of words, with alternating roles (words from user, vs. words from system), and proposed a variant to LSTM with role-dependent layers.   \citet{chen2016end} built on memory networks that learn which part of contextual information should be attended to, when making slot-tagging predictions.  Both models achieved higher accuracy than context-free models.

Although the three NLU tasks are often studied separately, there are benefits to jointly solving them (similar to multi-task learning), and over multiple domains, so that it may require fewer labeled data when creating NLU models for a new domain~\citep{hakkanitur16multi,liu16attention}.  Another line of interesting work that can lead to substantial reduction of labeling cost in new domains is \emph{zero-shot learning}, where slots from different domains are represented in a shared latent semantic space through embedding of the slots' (text) descriptions~\citep{bapna17towards,lee18zero}.  Interested readers are referred to recent tutorials, such as \citet{chen17open} and \citet{chen17deep}, for more details.


\subsection{Dialogue State Tracking}


In slot-filling problems, a dialogue state contains all information about what the user is looking for at the current turn of the conversation.  This state is what the dialogue policy takes as input for deciding what action to take next (\figref{fig:dialogue-arch}).

For example, in the restaurant domain, where a user tries to make a reservation, the dialogue state may consists of the following components~\citep{henderson15machine}:
\begin{itemize}
\item{The goal constraint for every informable slot, in the form of a value assignment to that slot.  The value can be ``\texttt{don't care}'' (if the user has no preference) or ``\texttt{none}'' (if the user has not yet specified the value).}
\item{The subset of requested slots that the user has asked the system to inform.}
\item{The current dialogue search method, taking values \texttt{by constraint}, \texttt{by alternative} and \texttt{finished}.  It encodes how the user is trying to interact with the dialogue system.}
\end{itemize}
Many alternatives have also been used in the literature, such as a compact, binary representation recently proposed by \citet{kotti18case}, and the StateNet tracker of 
\citet{ren18towards} that is more scalable with the domain size (number of slots and number of slot values).

In the past, DST can either be created by experts, or obtained from data by statistical learning algorithms like conditional random fields~\citep{henderson15machine}.  More recently, neural approaches have started to gain popularity, with applications of deep neural networks~\citep{henderson13deep} and recurrent networks~\citep{mrksic15multi} as some of the early examples.

\begin{figure}
\centering
\includegraphics[width=\textwidth]{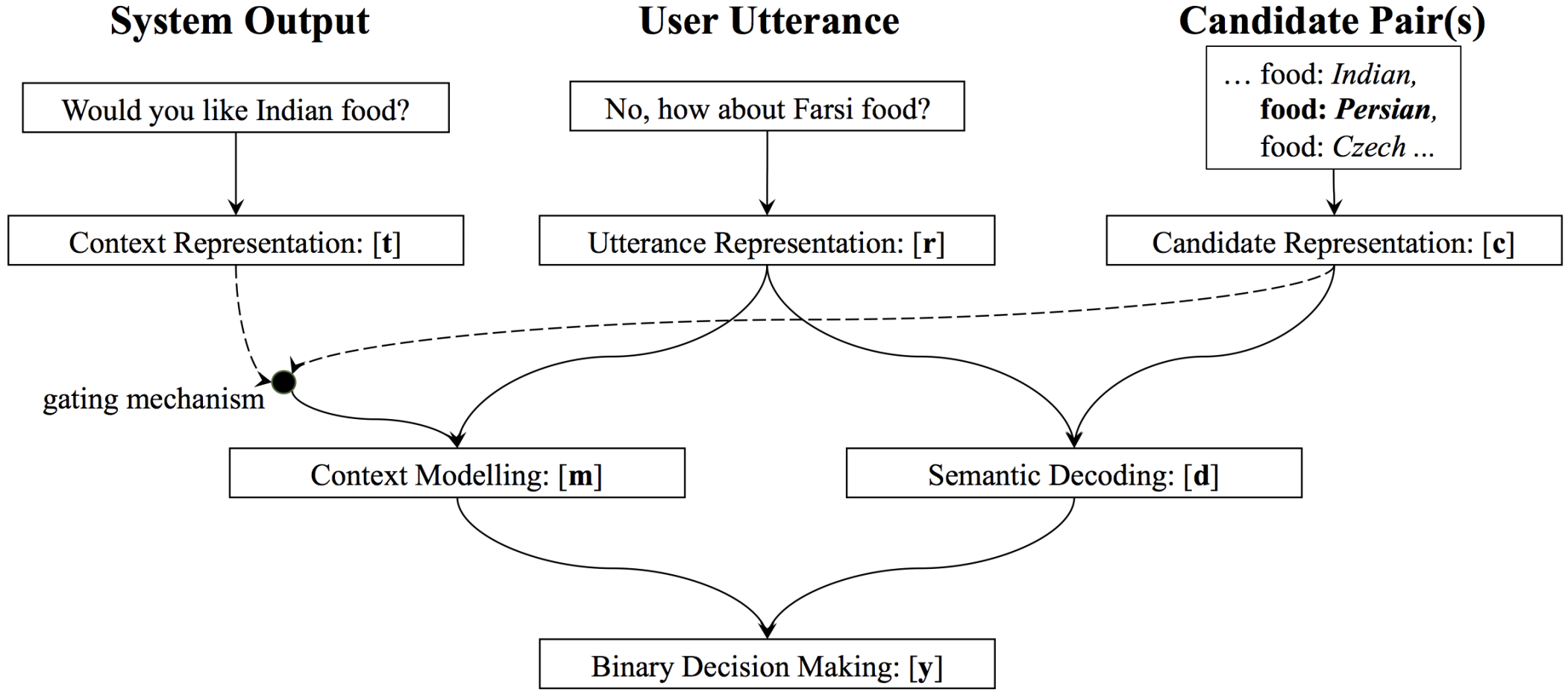}
\caption{Neural Belief Tracker. Figure credit: \citet{mrksic17neural}.} \label{fig:neural-belief-tracker}
\end{figure}

A more recent DST model is the Neural Belief Tracker proposed by \citet{mrksic17neural}, shown in \figref{fig:neural-belief-tracker}.  The model takes three items as input.  The first two are the last system and user utterances, each of which is first mapped to an internal, vector representation.  The authors studied two models for representation learning, based on multi-layer perceptrons and convolutional neural networks, both of which take advantage of pre-trained collections of word vectors and output an embedding for the input utterance.  The third input is any slot-value pair that is being tracked by DST.  Then, the three embeddings may interact among themselves for context modeling, to provide further contextual information from the flow of conversation, and semantic decoding, to decide if the user explicitly expressed an intent matching the input slot-value pair.  Finally, the context modeling and semantic decoding vectors go through a softmax layer to produce a final prediction.  The same process is repeated for all possible candidate slot-value pairs.

A different representation of dialogue states, called \emph{belief spans}, is explored by \citet{lei18sequicity} in the \emph{Sequicity} framework.  A belief span consists of two fields: one for informable slots and the other for requestable slots.  Each field collects values that have been found for respective slots in the conversation so far.  One of the main benefits of belief spans and Sequicity is that it facilitates the use of neural sequence-to-sequence models to learn dialogue systems, which take the belief spans as input and output system responses.  This greatly simplifies system design and optimization, compared to more traditional, pipeline approaches (c.f., \secref{sec:dialogue:e2e}).

\paragraph{Dialogue State Tracking Challenge (DSTC)} is a series of challenges that provide common testbeds and evaluation measures for dialogue state tracking.  Starting from \citet{williams13dialog}, it has successfully attracted many research teams to focus on a wide range of technical problems in  DST~\citep{williams14dialog,henderson14second,henderson14third,
kim16fourth,kim16fifth,hori17sixth}.  Corpora used by DSTC over the years have covered  human-computer and human-human conversations, different domains such as restaurant and tourist, cross-language learning.  More information may be found in the DSTC website.\footnote{\url{https://www.microsoft.com/en-us/research/event/dialog-state-tracking-challenge}} 

\section{Dialogue Policy Learning}


In this section, we will focus on dialogue policy optimization based on reinforcement learning.

\subsection{Deep RL for Policy Optimization}


The dialogue policy may be optimized by many standard reinforcement learning algorithms.  There are two ways to use RL: online and batch.  The online approach requires the learner to interact with users to improve its policy; the batch approach assumes a fixed set of transitions, and optimizes the policy based on the data only, without interacting with users (see, e.g., \citet{li09reinforcement,pietquin11sample}).  In this chapter, we discuss the online setting which often has batch learning as an internal step.  Many covered topics can be useful in the batch setting.  Here, we use the DQN as an example, following \citet{lipton18bbq}, to illustrate the basic work flow.

\paragraph{Model: Architecture, Training and Inference.}  The DQN's input is an encoding of the current dialogue state.  One option is to encode it as a feature vector, consisting of: (1) one-hot representations of the dialogue act and slot corresponding to the last user action; 
(2) the same one-hot representations of the dialogue act and slot corresponding to the last system action; %
(3) a bag of slots corresponding to all previously filled slots in the conversation so far; 
(4) the current turn count; and %
(5) the number of results from the knowledge base that match the already filled-in constraints for informed slots. 
Denote this input vector by $\mathbf{s}$.

DQN outputs a real-valued vector, whose entries correspond to all possible (dialogue-act, slot) pairs that can be chosen by the dialogue system.  Available prior knowledge can be used to reduce the number of outputs, if some (dialogue-act, slot) pairs do not make sense for a system, such as \dafont{request}\slotfont{(price)}.  Denote this output vector by $\mathbf{q}$.

The model may have $L \ge 1$ hidden layers, parameterized by matrices 
$\{\mathbf{W}_1, \mathbf{W}_2, \ldots, \mathbf{W}_L\}$, so that
\begin{align*}
\mathbf{h}_0 &= \mathbf{s} \\
\mathbf{h}_l &= g(\mathbf{W}_l \mathbf{h}_{l-1})\,, \qquad l = 1, 2, \ldots, L - 1 \\
\mathbf{q} &= \mathbf{W}_L \mathbf{h}_{L-1}\,,
\end{align*}
where $g(\cdot)$ is an activation function such as ReLU or sigmoid.  Note that the last layer does not need an activation function, and the output $\mathbf{q}$ is to approximate $Q(\mathbf{s}, \cdot)$, the Q-values in state $\mathbf{s}$.

To learn parameters in the network, one can use an off-the-shelf reinforcement-learning algorithm (e.g., \eqnref{eqn:qlearn} or \ref{eqn:qlearn2} with experience replay); see \secref{sec:basics:rl} for the exact update rules and improved algorithms.  Once these parameters are learned, the network induces a \emph{greedy} action-selection policy as follows: for a current dialogue state $\mathbf{s}$, use a forward pass on the network to compute $\mathbf{q}$, the Q-values for all actions.  One can pick the action, which is a (dialogue act, slot) pair, that corresponds to the entry in $\mathbf{q}$ with the largest value.   Due to the need for exploration, the above greedy action selection may not be desired; see \secref{sec:dialogue:exploration} for a discussion on this subject.


\paragraph{Warm-start Policy.}
Learning a good policy from scratch often requires many data, but the process can be significantly sped up by restricting the policy search using expert-generated dialogues~\citep{henderson08hybrid} or teacher advice~\citep{chen17agent}, or by initializing the policy to be a reasonable one before switching to online interaction with (simulated) users.

One approach is to use imitation learning (also known as behavioral cloning) to mimic an expert-provided policy.  A popular option is to use supervised learning to directly learn the expert's action in a state; see \citet{su16continuously,dhingra17towards,williams2017hybrid,liu17iterative} for a few recent examples.  \cite{li14temporal} turned imitation learning into an induced reinforcement learning problem, and then applied an off-the-shelf RL algorithm to learn the expert's policy.

Finally, \citet{lipton18bbq} proposed a simple yet effective alternative known as Replay Buffer Spiking (RBS) that is particularly suited to DQN.  The idea is to pre-fill the experience replay buffer of DQN with a small number of dialogues generated by running a na\"{i}ve yet occasionally successful, rule-based agent. This technique is shown to be essential for DQN in simulated studies.

\paragraph{Other Approaches.}
In the above example, a standard multi-layer perceptron is used in the DQN to approximate the Q-function.  It may be replaced by other models, such as a Bayesian version described in the next subsection for efficient exploration, and recurrent networks~\citep{zhao16towards,williams2017hybrid} that can more easily capture information from conversational histories than expert-designed dialogue states.  In another recent example, \citet{chen18structured} used graph neural networks to model the Q-function, with nodes in the graph corresponding to slots of the domain.  The nodes may share some of the parameters across multiple slots, therefore increasing learning speed.

Furthermore, one may replace the above value function-based methods by others like policy gradient (\secref{sec:basics:rl:algorithms}), as done by \citet{fatemi16policy,dhingra17towards,strub17end,williams2017hybrid,liu18dialogue}.

\subsection{Efficient Exploration and Domain Extension}
\label{sec:dialogue:exploration}

Without a teacher, an RL agent learns from data collected by interacting with an initially unknown environment.  In general, the agent has to try new actions in novel states, in order to discover potentially better policies.  Hence, it has to strike a good trade-off between exploitation (choosing good actions to maximize reward, based on information collected thus far) and exploration (choosing novel actions to discover potentially better alternatives), leading to the need for efficient exploration~\citep{sutton18reinforcement}.  In the context of dialogue policy learning, the implication is that the policy learner actively tries new ways to converse with a user, in the hope of discovering a better policy in the long run~\citep[e.g.,][]{daubigney11uncertainty}.

While exploration in finite-state RL is relatively well-understood~\citep{strehl09reinforcement,jaksch10near,osband17why,dann17unifying}, exploration when parametric models like neural networks are used is an active research topic~\citep{bellemare16unifying,osband16deep,houthooft16vime,jiang17contextual}.  Here, a general-purpose exploration strategy is described, which is particularly suited for dialogue systems that may evolve over time.  

After a task-oriented dialogue system is deployed to serve users, there may be a need over time to add more intents and/or slots to make the system more versatile.  This problem, referred to as \emph{domain extension}~\citep{gasic14incremental}, makes exploration even more challenging: the agent needs to explicitly quantify the uncertainty in its parameters for intents/slots, so as to explore new ones more aggressively while avoiding exploring those that have already been learned.  \citet{lipton18bbq} approached the problem using a Bayesian-by-Backprop variant of DQN.

Their model, called BBQ, is identical to DQN, except that it maintains a posterior \emph{distribution} $q$ over the network weights $\vecb{w}=(w_1,w_2,\ldots,w_d)$.  For computational convenience, $q$ is a multivariate Gaussian distribution with diagonal covariance, parameterized by $\theta=\{(\mu_i,\rho_i)\}_{i=1}^d$, where weight $w_i$ has a Gaussian posterior distribution, $\mathcal{N}(\mu_i, \sigma_i^2)$ and $\sigma_i = \log(1+\exp(\rho_i))$.  The posterior information leads to a natural exploration strategy, inspired by Thompson Sampling~\citep{thompson33likelihood,chapelle12empirical,russo18tutorial}.  When selecting actions, the agent simply draws a random weight $\tilde{\vecb{w}}\sim q$, and then selects the action with the highest value output by the network.  Experiments show that BBQ explores more efficiently than state-of-the-art baselines for dialogue domain extension.

The BBQ model is updated as follows.  Given observed transitions $\mathcal{T} = \{(s, a, r, s')\}$, one uses the target network (see \secref{sec:basics:rl}) to compute the target values for each $(s,a)$ in $\mathcal{T}$, resulting in the set $\mathcal{D} = \{(x,y)\}$, where $x=(s,a)$ and $y$ may be computed as in DQN.  Then, parameter $\theta$ is updated to represent the posterior distribution of weights.  Since the exact posterior is not Gaussian any more, and thus not representable by BBQ, it is approximated as follows: $\theta$ is chosen by minimizing the \emph{variational free energy}~\citep{hinton1993keeping}, the KL-divergence between the variational approximation $q(\mathbf{w}|\theta)$
and the posterior $p(\mathbf{w}|\mathcal{D})$:
\begin{eqnarray*}
\theta^* &=& \argmin_{\theta} \operatorname{KL} 
[q(\mathbf{w}|\theta) || p(\mathbf{w}| \mathcal{D})] \\
&=& \operatorname{argmin}_{\theta} \Big\{ \operatorname{KL}[q(\mathbf{w}|\theta) || p(\mathbf{w})]
- \mathbf{E}_{q(\mathbf{w}|\theta)}
[\log p(\mathcal{D}|\mathbf{w})] \Big\} \,.
\label{eqn:var-free-energy}
\end{eqnarray*}
In other words, the new parameter $\theta$ is chosen so that the new Gaussian distribution is closest to the posterior measured by KL-divergence.

\subsection{Composite-task Dialogues}

In many real-world problems, a task may consist of a set of subtasks that need to be solved collectively.  Similarly, dialogues can often be decomposed into a sequence of related sub-dialogues, each of which focuses on a subtopic~\citep{litman87plan}.  Consider for example a travel planning dialogue system, which needs to book flights, hotels and car rental in a collective way so as to satisfy certain cross-subtask constraints known as \emph{slot constraints}~\citep{peng17composite}. Slot constraints are application specific.  In a travel planning problem, one natural constraint is that the outbound flight's arrival time should be earlier than the hotel check-in time. 

Complex tasks with slot constraints are referred to as \emph{composite tasks} by \citet{peng17composite}.   Optimizing the dialogue policy for a composite task is challenging for two reasons.  First, the policy has to handle many slots, as each subtask often corresponds to a domain with its own set of slots, and the set of slots of a composite-task consists of slots from all subtasks.  Furthermore, thanks to slot constraints, these subtasks cannot be solved independently.  Therefore, the state space considered by a composite-task is much larger.  Second, a composite-task dialogue often requires many more turns to complete.  Typical reward functions give a success-or-not reward only at the end of the whole dialogue.  As a result, this reward signal is very sparse and considerably delayed, making policy optimization much harder.

\citet{cuayahuitl10evaluation} proposed to use hierarchical reinforcement learning to optimize a composite task's dialogue policy, with tabular versions of the MAXQ~\citep{dietterich00hierarchical} and Hierarchical Abstract Machine~\citep{parr98reinforcement} approaches.  While promising, their solutions assume finite states, so do not apply directly to larger-scale conversational problems.

\begin{figure}
\centering
\includegraphics[width=0.75\textwidth]{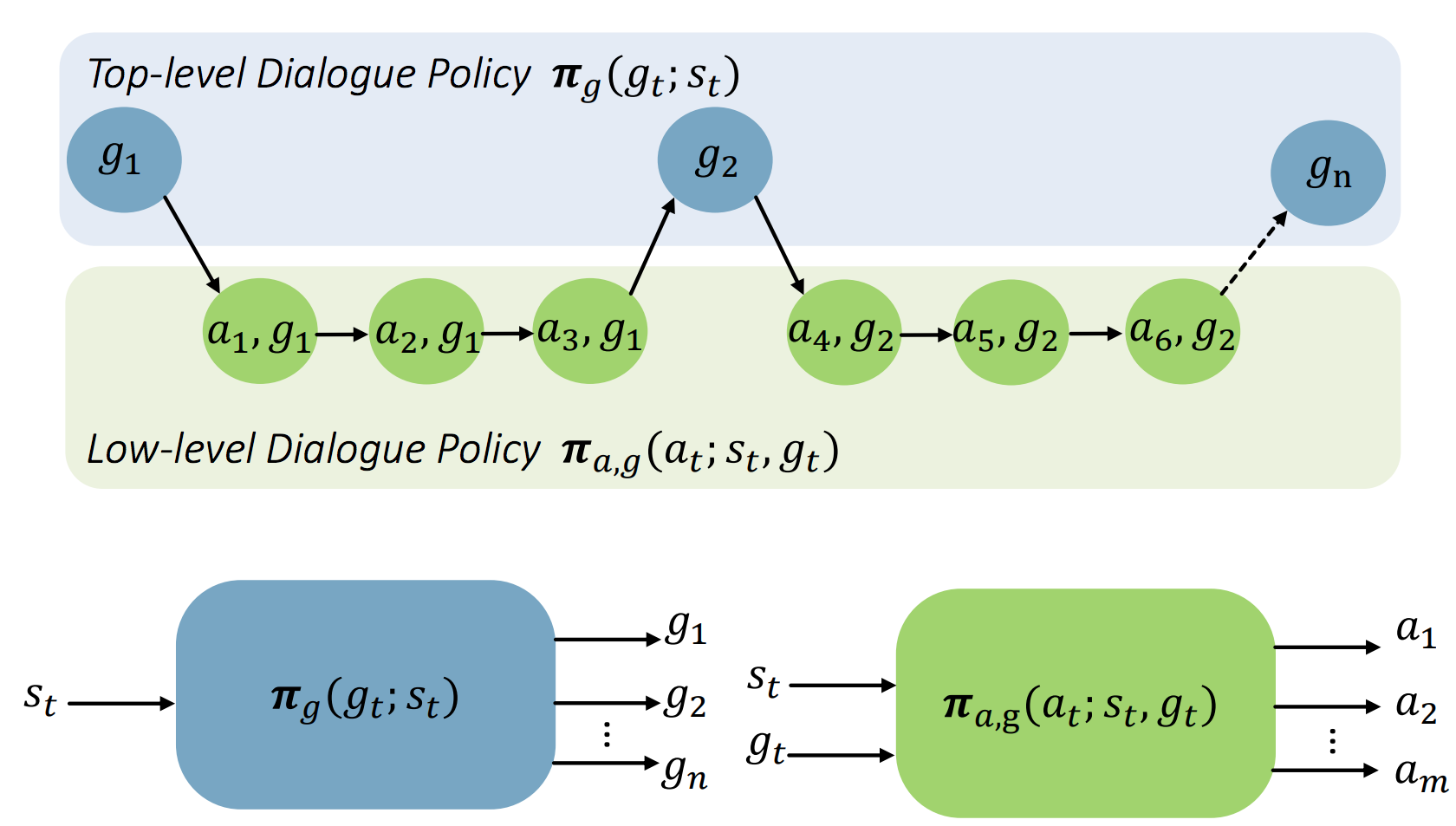}
\caption{A two-level hierarchical dialogue policy. Figure credit: \citet{peng17composite}.} \label{fig:hdqn_crop}
\end{figure}

More recently, \citet{peng17composite} tackled the composite-task dialogue policy learning problem under the more general \emph{options} framework~\citep{sutton99between}, where the task hierarchy has two levels.  As illustrated in \figref{fig:hdqn_crop}, a top-level policy $\pi_g$ selects which subtask $g$ to solve, and a low-level policy $\pi_{a,g}$  solves the subtask specified by $\pi_g$.	  Assuming predefined subtasks, they extend the DQN model that results in substantially faster learning speed and superior policies.  A similar approach is taken by \citet{budzianowski17subdomain}, who used Gaussian process RL instead of deep RL for policy learning.

A major assumption in options/subgoal-based hierarchical reinforcement learning is the need for reasonable options and subgoals.  \cite{tang18subgoal} considered the problem of discovering subgoals from dialogue demonstrations.  Inspired by a sequence segmentation approach that is successfully applied to machine translation~\citep{wang17sequence}, the authors developed the Subgoal Discovery Network (SDN), which learns to identify ``bottleneck'' states in successful dialogues.   It is shown that the hierarchical DQN optimized with subgoals discovered by SDN is competitive to expert-designed subgoals.

Finally, another interesting attempt is made by \citet{casanueva18feudal} based on Feudal Reinforcement Learning (FRL)~\citep{dayan93feudal}.  In contrast to the above methods that decompose a task into \emph{temporally} separated subtasks, FRL decomposes a complex decision \emph{spatially}.  In each turn of a dialogue, the feudal policy first decides between information-gathering actions and information-providing actions, then chooses a primitive action that falls in the corresponding high-level category.

\subsection{Multi-domain Dialogues}

A multi-domain dialogue can converse with a user to have a conversation that may involve more than one domain~\citep{komatani06multi,hakkanitur12discriminative,wang14policy}. 
\tabref{tab:multi-domain} shows an example, where the dialogue covers both the \dnfont{hotel} and \dnfont{restaurant} domains, in addition to a special \dnfont{meta} domain for sub-dialogues that contain domain-independent system and user responses.

\begin{table}
\caption{An example of multi-domain dialogue, adapted from \cite{cuayahuitl2016deep}.  The first column specifies which domain is triggered in the system, based on user utterances received so far.} \label{tab:multi-domain}
\centering
\footnotesize
\begin{tabular}{c|ll}
Domain & Agent & Utterance \\
\hline
\dnfont{meta} & system & ``Hi!  How can I help you?'' \\
           & user & ``I'm looking for a hotel in Seattle on January 2nd\\
           &      & ~for 2 nights.'' \\
\hline
\dnfont{hotel} & system & ``A hotel for 2 nights in Seattle on January 2nd?'' \\
           & user & ``Yes.'' \\
           & system & ``I found Hilton Seattle.'' \\
\hline 
\dnfont{meta} & system & ``Anything else I can help with?'' \\
           & user &  ``I'm looking for cheap Japanese food in the downtown.'' \\
\hline
\dnfont{restaurant} & system & ``Did you say cheap Japanese food?'' \\
           & user & ``Yes.'' \\
           & system & ``I found the following results.'' \\
           & ... & 
\end{tabular}
\end{table}

Different from composite tasks, sub-dialogues corresponding to different domains in a conversation are separate tasks, without cross-task slot constraints.  Similar to composite-task systems, a multi-domain dialogue system needs to keep track of a much larger dialogue state space that has slots from all domains, so directly applying RL can be inefficient.  It thus raises the need to learn re-usable policies whose parameters can be shared across multiple domains as long as they are related.

\cite{gasic15policy} proposed to use a Bayesian Committee Machine (BCM) for efficient multi-domain policy learning.  During training time, a number of policies are trained on different, potentially small, datasets.  The authors used Gaussian processes RL algorithms to optimize those policies, although they can be replaced by deep learning alternatives.  During test time, in each turn of a dialogue, these policies recommend an action, and all recommendations are aggregated into a final action to be taken by the BCM policy.

\citet{cuayahuitl2016deep} developed another related technique known as NDQN---Network of DQNs, where each DQN is trained for a specialized skill to converse
in a particular sub-dialogue.  A meta-policy controls how to switch between these DQNs, and can also be optimized using (deep) reinforcement learning.

More recently, \citet{papangelis18towards} studied another approach in which policies optimized for difference domains can be shared, through a set of features that describe a domain.  It is shown to be able to handle unseen domains, and thus reduce the need for domain knowledge to design the ontology.

\subsection{Integration of Planning and Learning}
\label{sec:dialogue:deep-dyna}

\begin{figure}
\centering
\includegraphics[width=\textwidth]{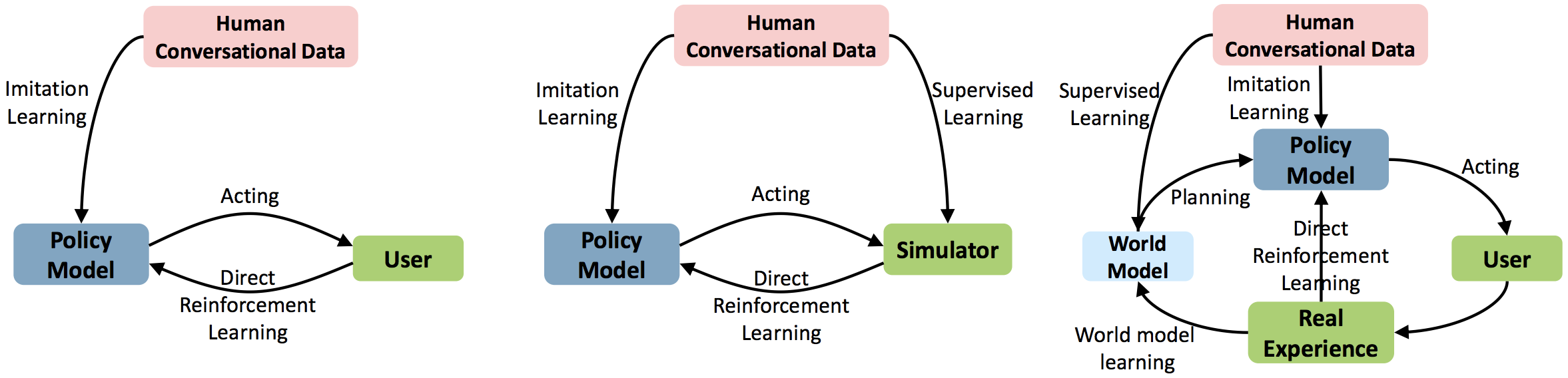}
\caption{Three strategies for optimizing dialogue policies based on reinforcement learning. Figure credit: \citet{peng18integrating}.} \label{fig:deep-dyna}
\end{figure}

As mentioned in \secref{sec:dialogue:evaluation}, optimizing the policy of a task-oriented dialogue against humans is costly, since it requires many interactions between the dialogue system and humans (left panel of \figref{fig:deep-dyna}).  Simulated users provide an inexpensive alternative to RL-based policy optimization (middle panel of \figref{fig:deep-dyna}), but may not be a sufficiently truthful approximation of human users. 


Here, we are concerned with the use of a user model to generate more data to improve sample complexity in optimizing a dialogue system.  Inspired by the Dyna-Q framework~\citep{sutton1990integrated}, \citet{peng18integrating} proposed Deep Dyna-Q (DDQ) to handle large-scale problems with deep learning models, as shown by the right panel of \figref{fig:deep-dyna}.  Intuitively, DDQ allows interactions with both human users and simulated users.  Training of DDQ consists of three parts:
\begin{itemize}
\item{\emph{direct reinforcement
learning}: the dialogue system interacts with a real user, collects real dialogues and improves the policy by either imitation learning or reinforcement learning;}
\item{\emph{world model learning}: the world model (user simulator) is refined using real dialogues collected by direct reinforcement learning;}
\item{\emph{planning}: the dialogue policy is improved against simulated users by reinforcement learning.}
%
%
%
\end{itemize}
Human-in-the-loop experiments show that DDQ is able to efficiently improve the dialogue policy by interacting with real users, which is important for deploying dialogue systems in practice.

One challenge with DDQ is to balance samples from real users (direct reinforcement learning) and simulated users (planning).  \citet{peng18integrating} used a heuristics that reduces planning steps in later stage of DDQ when more real user interactions are available.  In contrast, \citet{su18discriminative} proposed the Discriminative
Deep Dyna-Q (D3Q) that is
inspired by generative adversarial networks.  Specifically, it incorporates a discriminator which is trained to differentiate experiences of simulated users from those of real users.  During the planning step, a simulated experience is used for policy training only when it appears to be a real-user experience according to the discriminator.

\subsection{Reward Function Learning}
\label{sec:dialogue:reward}

The dialogue policy is often optimized to maximize long-term reward when interacting with users.  The reward function is therefore critical to creating high-quality dialogue systems.  One possibility is to have users provide feedback during or at the end of a conversation to rate the quality, but feedback like this is intrusive and costly.  Often, easier-to-measure quantities such as time-elapsed are used to compute a reward function. Unfortunately, in practice, designing an appropriate reward function is not always obvious, and substantial domain knowledge is needed (\secref{sec:dialogue:overview}).  This inspires the use of machine learning to find a good reward function from data~\citep{walker00towards,rieser08learning,rieser10optimising,asri12reward} which can better correlate with user satisfaction~\citep{rieser11learning}, or is more consistent with expert demonstrations~\citep{li14temporal}.

\citet{su15learning} proposed to rate dialogue success with two neural network models, a recurrent and a convolutional network.  Their approach is found to result in competitive dialogue policies, when compared to a baseline that uses prior knowledge of user goals.  However, these models assume the availability of labeled data in the form of (dialogue, success-or-not) pairs, in which the success-or-not feedback provided by users can be expensive to obtain.  To reduce the labeling cost, \citeauthor{su16online}~(\citeyear{su16online}; \citeyear{su18reward}) investigated an active learning approach based on Gaussian processes, which aims to learn the reward function and policy at the same time while interacting with human users.

\citet{ultes17domain} argued that dialogue success only measures one aspect of the dialogue policy's quality. 
Focusing on information-seeking tasks, the authors proposed a new reward estimator based on \emph{interaction quality} that balances multiple aspects of the dialogue policy.  Later on, \citet{ultes17reward} used multi-objective RL to automatically learn how to linearly combine multiple metrics of interest in the definition of reward function.

Finally, inspired by adversarial training in deep learning, \citet{liu18adversarial} proposed to view the reward function as a discriminator that distinguishes dialogues generated by humans from those by the dialogue policy.  Therefore, there are two learning processes in their approach: the reward function as a discriminator, and the dialogue policy optimized to maximize the reward function.  The authors showed that such an adversarially learned reward function can lead to better dialogue policies than with hand-designed reward functions.

\section{Natural Language Generation}
\label{sec:dialogue:nlg}


Natural Language Generation (NLG) is responsible for converting a communication goal, selected by the dialogue manager, into a natural language form.  It is an important component that affects naturalness of a dialogue system, and thus the user experience.

There exist many approaches to language generation.  The most common in practice is perhaps template- or rule-based ones, where domain experts design a set of templates or rules, and hand-craft heuristics to select a proper candidate to generate sentences.  Even though machine learning can be used to train certain parts of these systems~\citep{langkilde98generation,stent04trainable,walker07individual}, the cost to write and maintain templates and rules leads to challenges in adapting to new domains or different user populations.  Furthermore, the quality of these NLG systems is limited by the quality of hand-crafted templates and rules.

These challenges motivate the study of more data-driven approaches, known as \emph{corpus-based} methods, that aim to optimize a generation module from corpora~\citep{oh02stochastic,angeli10simple,kondadadi13statistical,mairesse14stochastic}.  Most such methods are based on supervised learning, while \citet{rieser10natural} takes a decision-theoretic view and uses reinforcement learning to make a trade-off between sentence length and information revealed.\footnote{Some authors~\citep{stone03microplanning,koller07sentence} have taken a similar, decision-theoretic point of view for NLG.  Their formulate NLG as a planning problem, as opposed to data-driven or corpus-based methods being discussed here.}

\begin{figure}[t]
    \centering
    \includegraphics[width=0.6\textwidth]{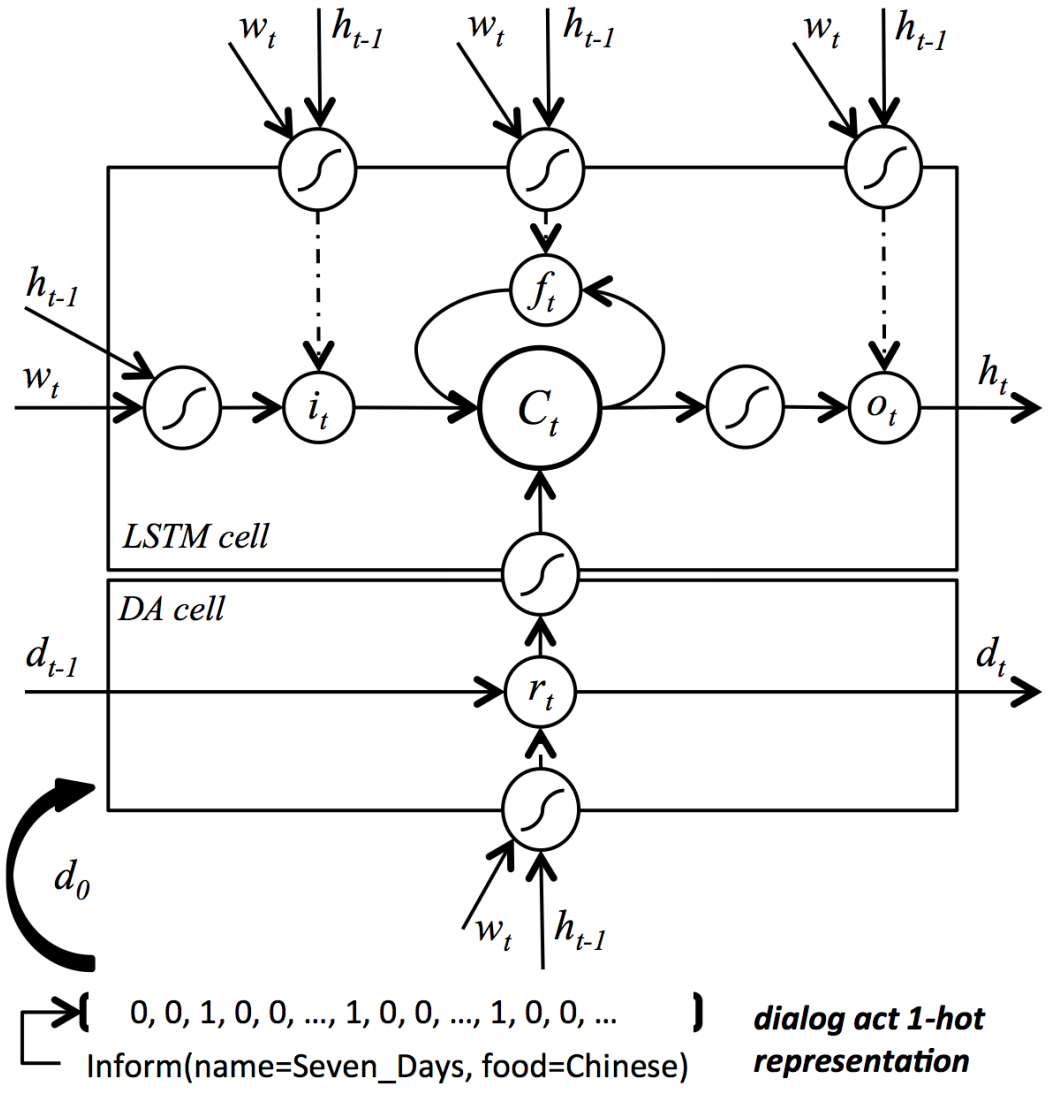}
    \caption{A Semantic Controlled LSTM (SC-LSTM) Cell.  Picture credit: \citet{wen2015semantically}.}
    \label{fig:sc-lstm}
\end{figure}

In recent years, there is a growing interest in neural approaches to language generation.  An elegant model, known as Semantically Controlled LSTM (SC-LSTM)~\citep{wen2015semantically}, is a variant of LSTM~\citep{hochreiter1997long}, with an extra component that gives a semantic control on the language generation results.  As shown in \figref{fig:sc-lstm}, a basic SC-LSTM cell has two parts: a typical LSTM cell (upper part in the figure) and a sentence planning cell (lower part) for semantic control.

More precisely, the operations upon receiving the $t^{\mathrm{th}}$ input token, denoted $\mathbf{w}_t$, are defined by the following set of equations:
\begin{align*}
\mathbf{i}_t &= g(\mathbf{W}_{wi} \mathbf{w}_t + \mathbf{W}_{hi} \mathbf{h}_{t-1}) \\
\mathbf{f}_t &= g(\mathbf{W}_{wf} \mathbf{w}_t + \mathbf{W}_{hf} \mathbf{h}_{t-1}) \\
\mathbf{o}_t &= g(\mathbf{W}_{wo} \mathbf{w}_t + \mathbf{W}_{ho} \mathbf{h}_{t-1}) \\
\mathbf{r}_t &= g(\mathbf{W}_{wr} \mathbf{w}_t + \alpha \mathbf{W}_{hr} \mathbf{h}_{t-1}) \\
\mathbf{d}_t &= \mathbf{r}_t \odot \mathbf{d}_{t-1} \\
\hat{\mathbf{c}}_t &= \operatorname{tanh}(\mathbf{W}_{wc} \mathbf{w}_t + \mathbf{W}_{hc} \mathbf{h}_{t-1}) \\
\mathbf{c}_t &= \mathbf{f}_t \odot \mathbf{c}_{t-1} + \mathbf{i}_t \odot \hat{\mathbf{c}}_t + \operatorname{tanh}(\mathbf{W}_{dc} \mathbf{d}_t) \\
\mathbf{h}_t &= \mathbf{o}_t \odot \operatorname{tanh}(\mathbf{c}_t) \,,
\end{align*}
where $\mathbf{h}_{t-1}$ is the hidden layer, 
$\mathbf{W}_{\star}$ 
the trainable parameters, and $g(\cdot)$ the sigmoid function.  As in a standard LSTM cell, $\mathbf{i}_t$, $\mathbf{f}_t$ and $\mathbf{o}_t$ are the input, forget, and output gates, respectively.  The extra component introduced to SC-LSTM is the \emph{reading gate} $\mathbf{r}_t$, which is used to compute a sequence of dialogue acts $\{\mathbf{d}_t\}$ starting from the original dialogue act $\mathbf{d}_0$.  This sequence is to ensure that the generated utterance represents the intended meaning, and the reading gate is to control what information to be retained for future steps.  It is in this sense that the gate $\mathbf{r}_t$ plays the role of \emph{sentence planning}~\citep{wen2015semantically}.  Finally, given the hidden layer $\mathbf{h}_t$, the output distribution is given by a softmax function:
\[
w_{t+1} \sim \operatorname{softmax}(\mathbf{W}_{ho} \mathbf{h}_t)\,.
\]

\citet{wen2015semantically} proposed several improvements to the basic SC-LSTM architecture.  One was to make the model deeper by stacking multiple LSTM cells on top of the structure in \figref{fig:sc-lstm}.  Another was utterance reranking: they trained another instance of SC-LSTM on the reversed input sequence, similar to bidirectional recurrent networks, and then combined both instances to finalize reranking.

The basic approach outlined above may be extended in several ways.  For example, \citet{wen16multi} investigated the use of multi-domain learning to reduce the amount of data to train a neural language generator, and \citet{su18natural} proposed a hierarchical approach that leverages linguistic patterns to further improve generation results.  Language generation remains an active research area.  The next chapter will cover more recent works for chitchat conversations, in which many techniques can also be useful in task-oriented dialogue systems.

\section{End-to-end Learning}
\label{sec:dialogue:e2e}

Traditionally, components in most dialogue systems are optimized separately.  This modularized approach provides the flexibility that allows each module to be created in a relatively independent way.  However, it often leads to a more complex system design, and improvements in individual modules do not necessarily translate into improvement of the whole dialogue system.  \citet{lemon11learning} argued for, and empirically demonstrated, the benefit of jointly optimizing dialogue management and natural language generation, within a reinforcement-learning framework.  More recently, with the increasing popularity of neural models, there have been growing interests in jointly optimizing multiple components, or even end-to-end learning of a dialogue system.  

One benefit of neural models is that they are often differentiable and can be optimized by gradient-based methods like back-propagation~\citep{goodfellow2016deep}.   In addition to language understanding, state tracking and policy learning that have been covered in previous sections, speech recognition \& synthesis (for spoken dialogue systems) may be learned by neural models and back-propagation to achieve state-of-the-art performance~\citep{hinton12deep,vandenoord16wavenet,wen2015semantically}.  In the extreme, if all components in a task-oriented dialogue system (\figref{fig:dialogue-arch}) are differentiable, the whole system becomes a larger differentiable system that can be optimized by back-propagation against metrics that quantify overall quality of the whole system.  This is an advantage compared to traditional approaches that optimize individual components separately.  There are two general classes of approaches to building an end-to-end dialogue system:

\paragraph{Supervised Learning.}
The first is based on supervised learning, where desired system responses are first collected and then used to train multiple components of a dialogue system in order to maximize prediction accuracy~\citep{bordes17learning,wen17network,yang17end,eric17key,madotto18mem2seq,wu18end}.

\citet{wen17network} introduced a modular neural dialogue system, where most modules are represented by a neural network.  However, their approach relies on non-differentiable knowledge-base lookup operators, so training of the components is done separately in a supervised manner.   This challenge is addressed by \citet{dhingra17towards} who proposed ``soft'' knowledge-base lookups; see \secref{sec:conversational-kbqa} for more details.

\citet{bordes17learning} treated dialogue system learning as the problem of learning a mapping from dialogue histories to system responses.  They show memory networks and supervised embedding models outperform standard baselines on a number of simulated dialogue tasks.  A similar approach was taken by \citet{madotto18mem2seq} in their Mem2Seq model.  This model uses mechanisms from pointer networks~\citep{vinyals15pointer} so as to incorporate external information from knowledge bases.

Finally, \citet{eric17key} proposed an end-to-end trainable Key-Value Retrieval
Network, which is equipped with an attention-based key-value retrieval mechanism over entries
of a KB, and can learn to extract relevant information from the KB.

\paragraph{Reinforcement Learning.} 
While supervised learning can produce promising results, they require training data that may be expensive to obtain.  Furthermore, this approach does not allow a dialogue system to explore different policies that can potentially be better than expert policies that produce responses for supervised training.  This inspire another line of work that uses reinforcement learning to optimize end-to-end dialogue systems~\citep{zhao16towards,williams2016end,dhingra17towards,li17end,braunschweiler18comparison,strub17end,liu17iterative,liu18dialogue}.

\citet{zhao16towards} proposed a model that takes user utterance as input and outputs a semantic system action.  Their model is a recurrent variant of DQN based on LSTM, which learns to compress a user utterance sequence to infer an internal state of the dialogue.  Compared to classic approaches, this method is able to jointly optimize the policy as well as language understanding and state tracking beyond standard supervised learning.

Another approach, taken by \citet{williams2017hybrid}, is to use LSTM to avoid the tedious step of state tracking engineering, and jointly optimize state tracker and the policy.  Their model, called Hybrid Code Networks (HCN), also makes it easy for engineers to incorporate business rules and other prior knowledge via software and action templates.  They show that HCN can be trained end-to-end, demonstrating much faster learning than several end-to-end techniques.

\citet{strub17end} applied policy gradient to optimize a visually grounded task-oriented dialogue in the GuessWhat?! game in an end-to-end fashion.  In the game, both the user and the dialogue system have access to an image.  The user chooses an object in the image without revealing it, and the dialogue system is to locate this object by asking the user a sequence of yes-no questions.

Finally, it is possible to combine supervised and reinforcement learning in an end-to-end trainable system.  \citet{liu18dialogue} proposed such a hybrid approach.  First, they used supervised learning on human-human dialogues to pre-train the policy.  Second, they used an imitation learning algorithm, known as DAgger~\citep{ross11reduction}, to fine tune the policy with human teachers who can suggest correct dialogue actions.  In the last step, reinforcement learning was used to continue policy learning with online user feedback.

%
%

\section{Further Remarks}

%
%
%
%
%
%
%
%
%


In this chapter, we have surveyed recent neural approaches to task-oriented dialogue systems, focusing on slot-filling problems.  This is a new area with many exciting research opportunities.  While it is out of the scope of the paper to give a full coverage of more general dialogue problems and all research directions, we briefly describe a small sample of them to conclude this chapter.

\paragraph{Beyond Slot-filling Dialogues.} 
Task-oriented dialogues in practice can be much more diverse and complex than slot-filling ones.  Information-seeking or navigation dialogues are another popular example that has been mentioned in different contexts (e.g., \citet{dhingra17towards}, \citet{papangelis18spoken}, and \secref{sec:conversational-kbqa}).  Another direction is to enrich the dialogue context.  Rather than text-only or speech-only ones, our daily dialogues are often \emph{multimodal}, and involve both verbal and nonverbal inputs like vision~\citep{bohus14directions,devault14simsensei,devries17guesswhat,zhang18multimodal}.  Challenges such as how to combine information from multiple modalities to make decisions arise naturally.

So far, we have looked at dialogues that involve two parties---the user and the dialogue agent, and the latter is to assist the former.  In general, the task can be more complex such as 
mixed-initiative dialogues~\citep{horvitz99principles} and
negotiations~\citep{barlier15human,lewis17deal}.  More generally, there may be multiple parties involved in a conversation, where turn taking becomes more challenging~\citep{bohus09models,bohus11multiparty}.  In such scenarios, it is helpful to take a game-theoretic view, more general than the MDP view as in single-agent decision making. 



\paragraph{Weaker Learning Signals.}
In the literature, a dialogue system can be optimized by supervised, imitation, or reinforcement learning.  Some require expert labels/demonstrations, while some require a reward signal from a (simulated) user.  There are other \emph{weaker} form of learning signals that facilitate dialogue management at scale.  A promising direction is to consider preferential input: instead of having an absolute judgment (either in the form of label or reward) of the policy quality, one only requires a preferential input that indicates which one of two dialogues is better.  Such comparable feedback is often easier and cheaper to obtain, and can be more reliable than absolute feedback.

\paragraph{Related Areas.}

Evaluation remains a major research challenge.  Although user simulation can be useful (\secref{sec:dialogue:evaluation:usersim}), 
a more appealing and robust solution is to use real human-human conversation corpora directly for evaluation.  Unfortunately, this problem, known as off-policy evaluation in the RL literature, is challenging with numerous current research efforts~\citep{precup00eligibility,jiang16doubly,thomas16data,liu18breaking}.  Such off-policy techniques can find important use in evaluating and optimizing dialogue systems.

Another related line of research is deep reinforcement learning applied to text games~\citep{narasimhan15language,cote18textworld}, which are in many ways similar to a conversation, except that the scenarios are predefined by the game designer.  Recent advances for solving text games, such as handling natural-language actions~\citep{narasimhan15language,he16deep,cote18textworld} and interpretable policies~\citep{chen17qlda} may be useful for task-oriented dialogues as well.

\chapter{Fully Data-Driven Conversation Models and Social Bots}
\label{sec:chitchat}

Researchers have recently begun to explore fully data-driven and end-to-end (E2E) approaches to conversational response generation, e.g., within the sequence-to-sequence (seq2seq) framework~\citep{hochreiter1997long,sutskever2014sequence}.
These models are trained entirely from data without resorting to any expert knowledge, 
which means they do not rely on the four traditional components of dialogue systems noted in \chref{sec:dialogue}.
Such end-to-end models have been particularly successful with social bot (chitchat) scenarios, as social bots rarely require interaction with the user's environment,
and the lack of external dependencies such as API calls 
simplifies end-to-end training.
By contrast, task-completion scenarios typically require such APIs in the form of, \eg, knowledge base access. 
The other reason this framework has been successful with chitchat is that it easily scales to large free-form and open-domain datasets, which means the user can typically chat on any topic of her liking.
While social bots are of significant importance in facilitating smooth interaction between humans and their devices, more recent work also focuses on scenarios going beyond chitchat, \eg, recommendation.


\section{End-to-End Conversation Models}
\label{sec:e2econvo}

Most of the earliest end-to-end (E2E) conversation models are inspired by statistical machine translation (SMT)~\citep{koehn03statistical,och04alignment}, including neural machine translation~\citep{kalchbrenner13recurrent,cho14properties,bahdanau2014neural}. 
The casting of the conversational response generation task (\ie, predict a response $T_i$ based on the previous dialogue turn $T_{i-1}$) as an SMT problem is a relatively natural one, 
as one can treat turn $T_{i-1}$ as the ``foreign sentence'' and turn $T_i$ as its ``translation''.
This means one can apply any off-the-shelf SMT algorithm to a conversational dataset to build a response generation system.
This was the idea originally proposed in one of the first works on fully data-driven conversational AI~\citep{ritter2011data}, which applied a phrase-based translation approach~\citep{koehn03statistical} to dialogue datasets extracted from Twitter~\citep{serban2015survey}. A different E2E approach was proposed in \citep{jafarpour10filter}, but it relied on IR-based methods rather than machine translation.

While these two papers constituted a paradigm shift, 
they had several limitations. 
The most significant one is their representation of the data as (query, response) pairs, which hinders their ability to generate responses that are contextually appropriate. 
This is a serious limitation as dialogue turns in chitchat are often short (\eg, a few word utterance such as ``really?''), in which case conversational models critically need longer contexts to produce plausible responses. 
This limitation motivated the work of~\citet{sordoni2015neural}, which proposed an RNN-based approach to conversational response generation (similar to \figref{fig:rnn-example}) to exploit longer context.
Together with the contemporaneous works~\citep{shang2015neural,vinyals2015neural},
these papers presented the first neural approaches to fully E2E conversation modeling. 
While these three papers have some distinct properties, they are all based on RNN architectures, which nowadays are often modeled with a Long Short-Term Memory (LSTM) model~\citep{hochreiter1997long,sutskever2014sequence}.

\subsection{The LSTM Model}
\label{sec:lstm}

We give an overview of LSTM-based response generation. LSTM is arguably the most popular seq2seq model, although alternative models like GRU~\citep{cho14learning} are often as effective. 
LSTM is an extension of the RNN model in \figref{fig:rnn-example}, and is often more effective at exploiting long-term context.

An LSTM-based response generation system is usually modeled as follows~\citep{vinyals2015neural,li2015diversity}:
Given a dialogue history represented as a sequence of words $S=\{s_1,s_2,...,s_{N_s}\}$ ($S$ here stands for source), the LSTM associates each time step $k$ with input, memory, and output gates, denoted respectively as $\mathbf{i}_k$, $\mathbf{f}_k$ and $\mathbf{o}_k$.
$N_s$ is the number of words in the source $S$.\footnote{The notation distinguishes $\mathbf{e}$ and $\mathbf{h}$ where $\mathbf{e}_k$ is the embedding vector for an individual word at time step~$k$, and $\mathbf{h}_k$ is the vector computed by the LSTM model at time $k$ by combining $\mathbf{e}_k$ and $\mathbf{h}_{k-1}$. $\mathbf{c}_k$ is the cell state vector at time $k$, and $\sigma$ represents the sigmoid function.}
Then, the hidden state $\mathbf{h}_k$ of the LSTM for each time step $k$ is computed as follows:
\begin{eqnarray}
\mathbf{i}_k=\sigma (\mathbf{W}_i [\mathbf{h}_{k-1};\mathbf{e}_k])\\
\mathbf{f}_k=\sigma (\mathbf{W}_f [\mathbf{h}_{k-1};\mathbf{e}_k])\\
\mathbf{o}_k=\sigma (\mathbf{W}_o [\mathbf{h}_{k-1};\mathbf{e}_k])\\
\mathbf{l}_k=\text{tanh}(\mathbf{W}_l [\mathbf{h}_{k-1};\mathbf{e}_k])\\
\mathbf{c}_k=\mathbf{f}_k\circ \mathbf{c}_{k-1}+\mathbf{i}_k\circ \mathbf{l}_k\\
\mathbf{h}_{k}^s=\mathbf{o}_k \circ \text{tanh}(\mathbf{c}_k)
\end{eqnarray}
where matrices $\mathbf{W}_i$, $\mathbf{W}_f$, $\mathbf{W}_o$, $\mathbf{W}_l$ belong to $\mathbb{R}^{d\times 2d}$, $\circ$ denotes the element-wise product.
As it is a response generation task, each conversational context $S$ is paired with a sequence of output words to predict: $T=\{t_1,t_2,...,t_{N_t}\}$.  Here, $N_t$ is the length of the response and $t$ represents a word token that is associated with a $d$-dimensional word embedding $e_t$ (distinct from the source).

The LSTM model defines the probability of the next token to predict using the softmax function.  Specifically, let $f(\mathbf{h}_{k-1}, \mathbf{e}_{y_k})$ be the softmax activation function of $\mathbf{h}_{k-1}$ and $\mathbf{e}_{y_k}$, where $\mathbf{h}_{k-1}$ is the hidden vector at time \mbox{$k-1$}. 
Then, the probability of outputing token $T$ is given by
\begin{equation*}
\begin{aligned}
p(T|S)
&=\prod_{k=1}^{N_t}p(t_k|s_1,s_2,...,s_t,t_1,t_2,...,t_{k-1})\\
&=\prod_{k=1}^{N_t}\frac{\exp(f(h_{k-1},e_{y_k}))}{\sum_{y'}\exp(f(h_{k-1},e_{y'}))}\,.
\end{aligned}
\label{equ-lstm}
\end{equation*}

\subsection{The HRED Model}

While the LSTM model has been shown to be effective in encoding textual contexts up to 500 words~\citep{khandelwal18sharp}, dialogue histories can often be long and there is sometimes a need to exploit longer-term context. 
Hierarchical models were designed to address this limitation by capturing longer context \citep{yao2015attention,serban2016building,serban2017hierarchical,xing2018hierarchical}.
One popular approach is the Hierarchical Recurrent Encoder-Decoder (HRED) model, originally proposed in~\citep{sordoni15hier} for query suggestion and applied to response generation in~\citep{serban2016building}.

\begin{figure}[t]
\centering 
\includegraphics[width=0.6\linewidth]{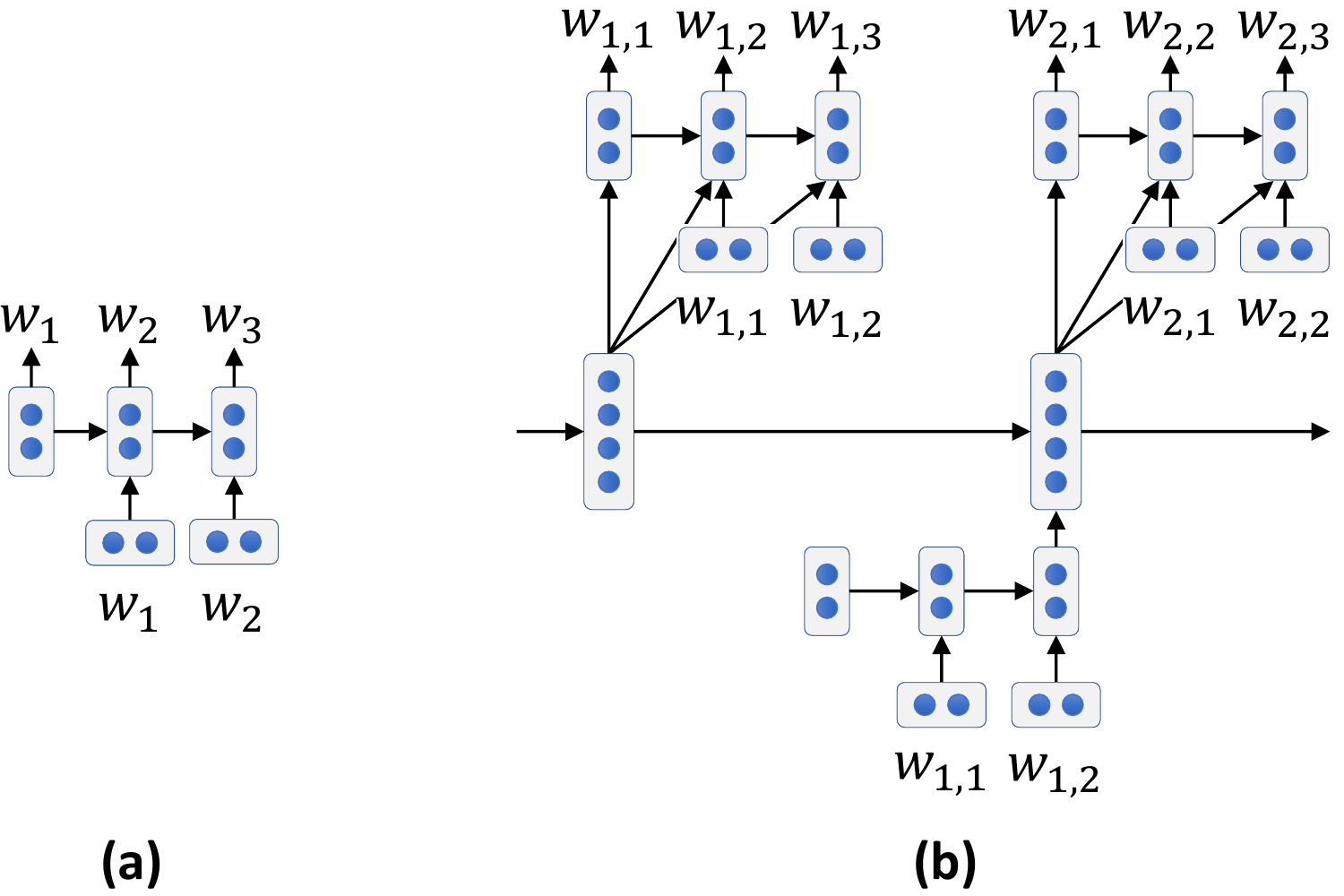}
\caption{(a) Recurrent architecture used by models such as RNN, GRU, LSTM, etc. (2) Two-level hierarchy representative of HRED.
Note: To simplify the notation, the figure represents utterances of length 3.}
\label{fig:hred} 
\end{figure}

The HRED architecture is depicted in \figref{fig:hred}, where it is compared to the standard RNN architecture. 
HRED models dialogue using a two-level hierarchy that combines two RNNs: one at a word level and one at the dialogue turn level. 
This architecture models the fact that dialogue history consists of a sequence of turns, each consisting of a sequence of tokens.
This model introduces a temporal structure that makes the hidden state of the current dialogue turn directly dependent on the hidden state of the previous dialogue turn, effectively allowing information to flow over longer time spans, and helping reduce the vanishing gradient problem \citep{hochreiter91}, a problem that limits RNN's (including LSTM's) ability to model very long word sequences.
Note that, in this particular work, RNN hidden states are implemented using GRU \citep{cho14learning} instead of LSTM.

\subsection{Attention Models}

The seq2seq framework has been tremendously successful in text generation tasks such as machine translation, but its encoding of the entire source sequence into a fixed-size vector has certain limitations, especially when dealing with long source sequences. 
Attention-based models~\citep{bahdanau2014neural,vaswani:17} alleviate this limitation by allowing the model to search and condition on parts of a source sentence that are relevant to predicting the next target word, thus moving away from a framework that represents the entire source sequence merely as a single fixed-size vector. 
While attention models and variants \citep[etc.]{bahdanau2014neural,luong15effective} have contributed to significant progress in the state-of-the-art in translation \citep{wu16google} and are very commonly used in neural machine translation nowadays, attention models have been somewhat less effective in E2E dialogue modeling. 
This can probably be explained by the fact that attention models effectively attempt to ``jointly translate and align'' \citep{bahdanau2014neural}, which is a desirable goal in machine translation as each information piece in the source sequence (foreign sentence) typically needs to be conveyed in the target (translation) exactly once, but this is less true in dialogue data. Indeed, in dialogue entire spans of the source may not map to anything in the target and vice-versa.\footnote{\citet{ritter2011data} also found that alignment produced by an off-the-shelf word aligner \citep{och03asc} produced alignments of poor quality, and an extension of their work with attention models (Ritter 2018, pc) yield attention scores that did not correspond to meaningful alignments.} 
Some specific attention models for dialogue have been shown to be useful \citep{yao2015attention,mei2016coherent,shao2017generating}, \eg, to avoid word repetitions (which are discussed further in \secref{sec:challenges}).

\subsection{Pointer-Network Models}

Multiple model extensions~\citep{gu2016copynet,he2017generating} of the seq2seq framework improve the model's ability to ``copy and paste'' words between the conversational context and the response.
Compared to other tasks such as translation, this ability is particularly important in dialogue, as the response often repeats spans of the input (\eg, ``good morning'' in response to ``good morning'') or uses rare words such as proper nouns, which the model would have difficulty generating with a standard RNN.
Originally inspired by the Pointer Network model~\citep{vinyals15pointer}---which produces an output sequence consisting of elements from the input sequence---these models hypothesize target words that are either drawn from a fixed-size vocabulary (akin to a seq2seq model) or selected from the source sequence (akin to a pointer network) using an attention mechanism. 
An instance of this model is CopyNet \citep{gu2016copynet}, which was shown to significantly improve over RNNs thanks to its ability to repeat proper nouns and other words of the input.

\section{Challenges and Remedies}
\label{sec:challenges}

The response generation task faces challenges that are rather specific to conversation modeling. Much of the recent research is aimed at addressing the following issues.

\subsection{Response Blandness}

Utterances generated by neural response generation systems are often bland and deflective. 
While this problem has been noted in other tasks such as image captioning~\citep{mao2014deep}, the problem is particularly acute in E2E response generation, as commonly used models such as seq2seq tend to generate uninformative responses such as ``I don't know'' or ``I'm OK''. 
\citet{li2015diversity} suggested that this is due to their training objective, which optimizes the likelihood of the training data according to $p(T|S)$, where $S$ is the source (dialogue history) and $T$ is the target response.
The objective $p(T|S)$ is asymmetrical in $T$ and $S$, which causes the trained systems to prefer responses $T$ that unconditionally enjoy high probability, \ie, irrespectively of the context $S$. 
For example, such systems often respond ``I don't know'' if $S$ is a question, as the response ``I don't know'' is plausible for almost all questions.
\citet{li2015diversity} suggested replacing the conditional probability $p(T|S)$ with mutual information $\frac{p(T,S)}{p(T)p(S)}$ as an objective, since the latter formulation is symmetrical in $S$ and $T$, thus giving no incentive for the learner to bias responses $T$ to be particularly bland and deflective, unless such a bias emerges from the training data itself. 
While this argument may be true in general, optimizing the mutual information objective (also known as Maximum Mutual Information or MMI \citep{huang2001spoken}) can be challenging, so \citet{li2015diversity} used that objective at inference time. More specifically, 
given a conversation history $S$, the goal at inference time is to find the best $T$ according to:\footnote{Recall that $\log\frac{p(S,T)}{p(S)p(T)} = \log\frac{p(T|S)}{p(T)} = \log p(T|S) - \log p(T)$}
\begin{equation}
\begin{split}
\hat{T} &= \argmax_{T} \big\{\log \frac{p(S,T)}{p(S)p(T)} \big\} \\
        &= \argmax_{T} \big\{\log p(T|S) - \log p(T)\big\}
\end{split}
\end{equation}
A hyperparameter $\lambda$ was introduced to control how much to penalize generic responses, with either formulations:\footnote{The second formulation is derived from:\\ $\log p(T)=\log p(T|S)+\log p(S)-\log p(S|T)$.}
\begin{equation}
\begin{split}
\hat{T} &=\argmax_{T} \big\{\log p(T|S) - \lambda\log p(T)\big\}\\
&=\argmax_{T} \big\{(1-\lambda)\log p(T|S)\\
&~~~~~~~~~~~~~~~~+\lambda\log p(S|T)-\lambda\log p(S) \big\} \\[0.2cm]
&=\argmax_{T} \big\{(1-\lambda)\log p(T|S)+\lambda\log p(S|T) \big\}\,.
\label{eqbayesexpanded}
\end{split}
\end{equation}
Thus, this weighted MMI objective function can be viewed as representing a tradeoff between sources given targets (\ie, $p(S|T)$) and targets given sources (\ie, $p(T|S)$), which is also a tradeoff between response appropriateness and lack of blandness. Note, however, that despite this tradeoff, \citet{li2015diversity} have not entirely solved the blandness problem, as this objective is only used at inference and not training time. This approach first generates $N$-best lists according to $p(T|S)$ and rescores them with MMI. Since such $N$-best lists tend to be overall relatively bland due to the $p(T|S)$ inference criterion (beam search), MMI rescoring often mitigates rather than completely eliminates the blandness problem.

More recently, researchers \citep{li2017adversarial,xu2017neural,zhang18towards} have used adversarial training and Generative Adversarial Networks (GAN) \citep{goodfellow2014generative}, which often have the effect of reducing blandness. 
Intuitively, the effect of GAN on blandness can be understood as follows: adversarial training puts a Generator and Discriminator against each other (hence the term ``adversarial'') using a minimax objective, and the objective for each of them is to make their counterpart the least effective. 
The Generator is the response generation system to be deployed, while the goal of the Discriminator is to be able to identify whether a given response is generated by a human (\ie, from the training data) or is the output of the Generator. 
Then, if the Generator always responds ``I don't know'' or with other deflective responses, the Discriminator would have little problem distinguishing them from human responses in most of the cases, as most humans do not respond with ``I don't know'' all the time. 
Therefore, in order to fool the Discriminator, the Generator progressively steers away from such predictable responses. 
More formally, the optimality of GAN is achieved when the hypothesis distribution matches the oracle distribution, thus encouraging the generated responses to spread out to reflect the true diversity of real responses.
To promote more diversity, \citet{zhang18towards} explicitly optimize a variational lower bound on pairwise mutual information between query and response to encourage generating more informative responses during training time. 

\citet{serban2017hierarchical} presented a latent Variable Hierarchical Recurrent Encoder-Decoder (VHRED) model that also aims to generate less bland and more specific responses. It extends the HRED model described previously in this chapter, by adding a high-dimensional stochastic latent variable to the target. 
This additional latent variable is meant to address the challenge associated with the {\it shallow generation} process.
As noted in \citep{serban2017hierarchical}, this process is problematic from an inference standpoint because the generation model is forced to produce a high-level structure---\ie, an entire response---on a word-by-word basis. 
This generation process is made easier in the VHRED model, as the model exploits a high-dimensional latent variable that determines high-level aspects of the response (topic, names, verb, etc.), so that the other parts of the model can focus on lower-level aspects of generation, \eg, ensuring fluency. The VHRED model incidentally helps reducing blandness as suggested by sample outputs of \citep{serban2017hierarchical}. Indeed, as the content of the response is conditioned on the latent variable, the generated response is only bland and devoid of semantic content if the latent variable determines that the response should be as such.
More recently, \citet{zhang18control} presented a model that also introduces an additional variable (modeled using a Gaussian kernel layer), which is added to control the level of specificity of the response, going from bland to very specific.

While most response generation systems surveyed earlier in this chapter are generation-based (\ie, generating new sentences word-by-word), a more conservative solution to mitigating blandness is to replace generation-based models with retrieval-based models for response generation \citep{jafarpour10filter,lu14deep,inaba2016,alrfou2016conversational,yan16learning}, in which the pool of possible responses is constructed in advance (\eg, pre-existing human responses). 
These approaches come at the cost of reduced flexibility: In generation, the set of possible responses grows exponentially in the number of words, but the set of responses of a retrieval system is fixed, and as such retrieval systems often do not have any appropriate responses for many conversational inputs. 
Despite this limitation, retrieval systems have been widely used in popular commercial systems, and we survey them in \chref{sec:commercial}.

\subsection{Speaker Consistency}

It has been shown that the popular seq2seq approach often produces conversations that are incoherent \citep{li2016persona}, where the system may for instance contradict what it had just said in the previous turn (or sometimes even in the same turn).
While some of this effect can be attributed to the limitation of the learning algorithms, \citet{li2016persona} 
suggested that the main cause of this inconsistency is probably due to the training data itself.
Indeed, conversational datasets (see \secref{sec:datasets}) feature multiple speakers, which often have different or conflicting personas and backgrounds. For example, to the question ``how old are you?'', a seq2seq model may give valid responses such as ``23'', ``27'', or ``40'', all of which are represented in the training data.

This sets apart the response generation task from more traditional NLP tasks:
While models for other tasks such as machine translation are trained on data that is mostly one-to-one semantically,
conversational data is often one-to-many or many-to-many as the above example implies.\footnote{Conversational data is also many-to-one, for example with multiple semantically-unrelated inputs that map to ``I don't know.''} 
As one-to-many training instances are akin to noise to any learning algorithm, one needs more expressive models that exploits a richer input to better account for such diverse responses.

To do this, \citet{li2016persona} proposed a persona-based response generation system, which is an extension of the LSTM model of \secref{sec:lstm} that uses speaker embeddings in addition to word embeddings. 
Intuitively, these two types of embeddings work similarly: 
while word embeddings form a latent space in which spacial proximity (\ie, low Euclidean distance) means two words are semantically or functionally close, speaker embeddings also constitute a latent space in which two nearby speakers tend to converse in the same way, \eg, having similar speaking styles (\eg, British English) or often talking about the same topic (\eg, sports).

\begin{figure}[t]
\centering 
\includegraphics[width=0.86\linewidth]{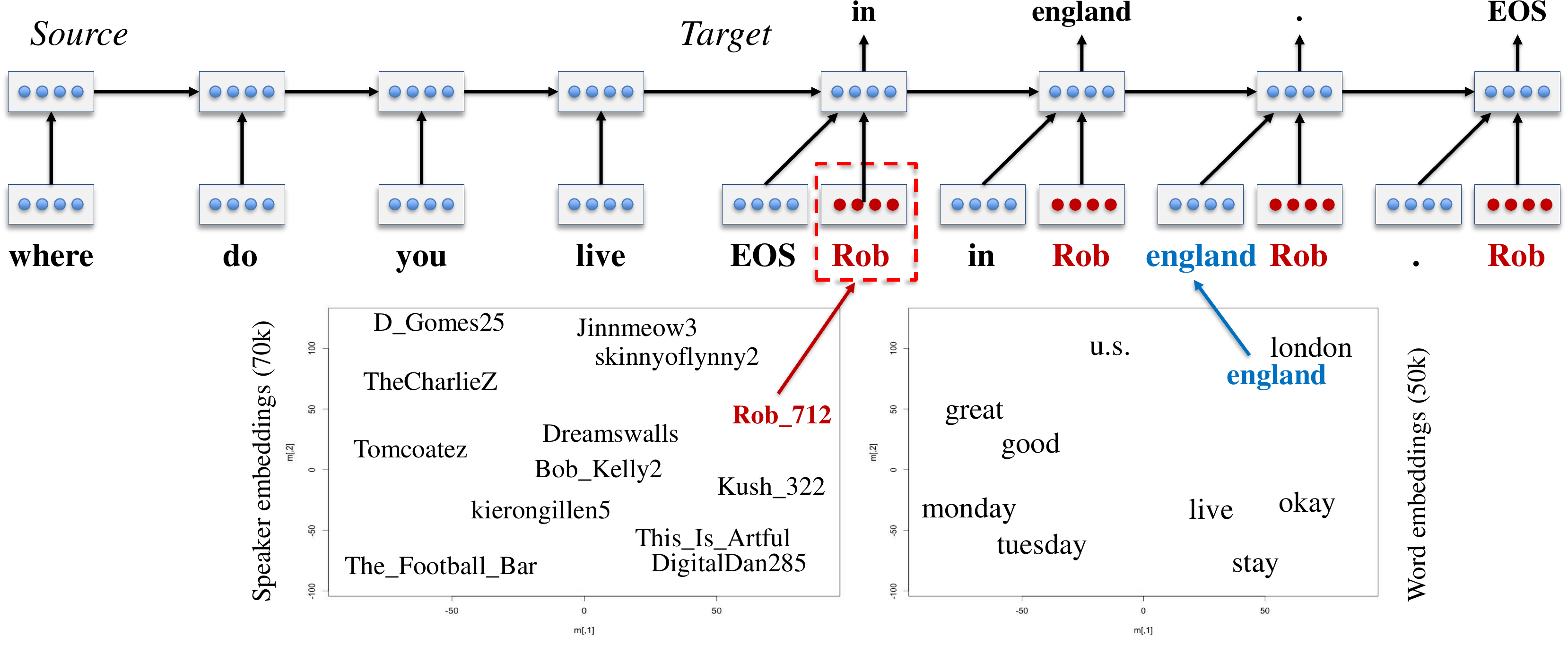}
\caption{Persona-based response generation system. Figure credit: \citet{li2016persona}}
\label{fig:persona} 
\end{figure}

Like word embeddings, speaker embedding parameters are learned jointly with all other parameters of the model from their one-hot representations. 
At inference time, one just needs to specify the one-hot encoding of the desired speaker to produce a response that reflects her speaking style. 
The global architecture of the model is displayed in \figref{fig:persona}, which shows that each target hidden state is conditioned not only on the previous hidden state and the current word embedding (\eg, ``England''), but also on the speaker embedding (\eg, of ``Rob''). This model not only helps generate more personalized responses, but also alleviates the one-to-many modeling problem mentioned earlier.

Other approaches also utilized personalized information. 
For example, \citet{alrfou2016conversational} presented a persona-based response generation model, but geared for retrieval using an extremely large dataset consisting of 2.1 billion responses. 
Their retrieval model is implemented as a binary classifier (\ie, good response or not) using a deep neural network.
The distinctive feature of their model is a multi-loss objective, which augments a single-loss model $p(R|I,A,C)$ of the response $R$, input $I$, speaker (``author'') $A$, and context $C$, by adding auxiliary losses that, \eg, model the probability of the response given the author $p(R|A)$.
This multi-loss model was shown to be quite helpful \citep{alrfou2016conversational}, as the multiple losses help cope with the fact that certain traits of the author are often correlated with the context or input, which makes it difficult to learn good speaker embedding representation. By adding a loss for $p(R|A)$, the model is able to learn a more distinctive speaker embedding representation for the author.

More recently, \citet{luan2017multi} presented an extension of the speaker embedding model of~\citet{li2016persona}, which combines a seq2seq model trained on conversational datasets with an autoencoder trained on non-conversational data, where the seq2seq and autoencoder are combined in a multi-task learning setup \citep{caruana1998multitask}. 
The tying of the decoder parameters of both seq2seq and autoencoder enables \citet{luan2017multi} to train a response generation system for a given persona without actually requiring any conversational data available for that persona. This is an advantage of their approach, as conversational data for a given user or persona might not always be available. 
In \citep{bhatia17soc}, the idea of \citep{li2016persona} is extended to a social-graph embedding model.

While \citep{serban2017hierarchical} is not a persona-based response generation model {\it per se}, their work shares some similarities with speaker embedding models such as \citep{li2016persona}. 
Indeed, both \citet{li2016persona} and \citet{serban2017hierarchical} introduced a continuous high-dimensional variable in the target side of the model in order to bias the response towards information encoded in a vector. 
In the case of \citep{serban2017hierarchical}, that variable is latent, and is trained by maximizing a variational lower-bound on the log-likelihood. 
In the case of \citep{li2016persona}, the variable (\ie, the speaker embedding) is technically also latent, although it is a direct function of the one-hot representation of speaker. 
\citep{li2016persona} might be a good fit when utterance-level information (\eg, speaker ID or topic) is available. On the other hand, the strength of \citep{serban2017hierarchical} is that it learns a latent variable that best ``explains'' the data, and may learn a representation that is more optimal than the one based strictly on speaker or topic information.

\subsection{Word Repetitions}

Word or content repetition is a common problem with neural generation tasks other than machine translation, as has been noted with tasks such as response generation, image captioning, visual story generation, and general language modeling \citep{shao2017generating,huang18hier,holtzman18learning}. 
While machine translation is a relatively one-to-one task where each piece of information in the source (\eg, a name) is usually conveyed exactly once in the target, other tasks such as dialogue or story generation are much less constrained, and a given word or phrase in the source can map to zero or multiple words or phrases in the target. 
This effectively makes the response generation task much more challenging, as generating a given word or phrase doesn't completely preclude the need of generating the same word or phrase again.
While the attention model \citep{bahdanau2014neural} helps prevent repetition errors in machine translation as that task is relatively one-to-one,\footnote{\citet{ding17vis} indeed found that word repetition errors, usually few in machine translation, are often caused by incorrect attention.} the attention models originally designed for machine translation \citep{bahdanau2014neural,luong15effective} often do not help reduce word repetitions in dialogue. 

In light of the above limitations, \citet{shao2017generating} proposed a new model that adds self-attention to the decoder, aiming at improving the generation of longer and coherent responses while incidentally mitigating the word repetition problem. 
Target-side attention helps the model more easily keep track of what information has been generated in the output so far,\footnote{A seq2seq model can also keep track of what information has been generated so far. However, this becomes more difficult as contexts and responses become longer, as a seq2seq hidden state is a fixed-size vector.} so that the model can more easily discriminate against unwanted word or phrase repetitions. 

\subsection{Further Challenges}

The above issues are significant problems that have only been partially solved and that require further investigation. However, 
a much bigger challenge faced by these E2E systems is response appropriateness. 
As explained in \chref{sec:intro}, one of the most distinctive characteristics of earlier E2E systems, when compared to traditional dialogue systems, is their lack of {\it grounding}. 
When asked ``what is the weather forecast for tomorrow?'', E2E systems are likely to produce responses such as ``sunny'' and ``rainy'', without a principled basis for selecting one response or the other, as the context or input might not even specify a geographical location.
\citet{ghazvininejad2017knowledge} argued that seq2seq and similar models are usually quite good at producing responses that have plausible overall {\it structure}, but often struggle when it comes to generating names and facts that connect to the real world, due to the lack of grounding. 
In other words, responses are often pragmatically correct (\eg, a question would usually be followed by an answer, and an apology by a downplay), but the semantic content of the response is often inappropriate. 
Hence, recent research in E2E dialogue has increasingly focused on designing {\it grounded} neural conversation models, which we will survey next.

\section{Grounded Conversation Models}
\label{sec:groundedconvo}

\begin{figure}[t]
\centering 
\includegraphics[width=0.76\linewidth]{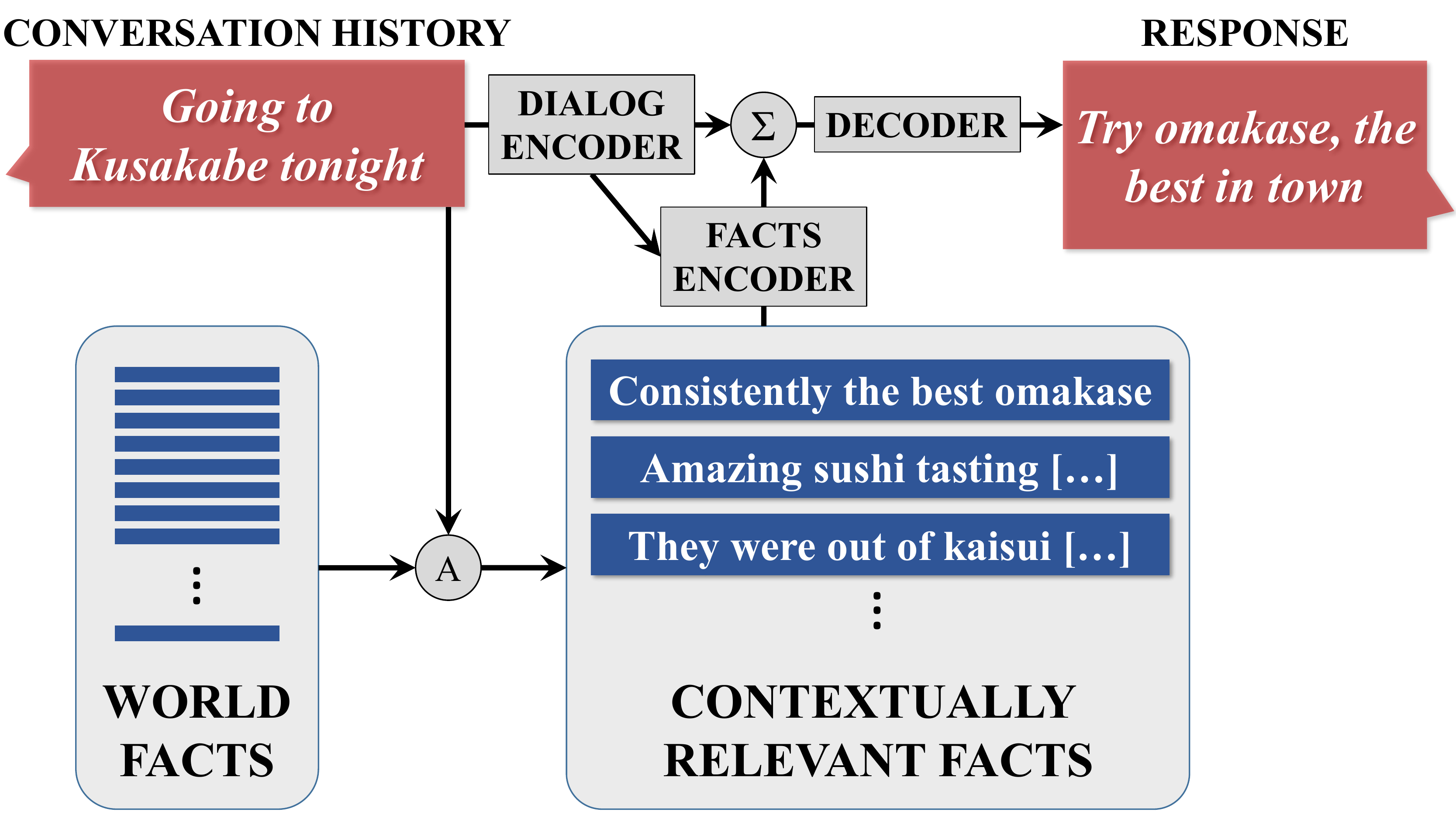}
\caption{A neural conversation model grounded in ``facts'' relevant to the current conversation. Figure credit: \citet{ghazvininejad2017knowledge}}
\label{fig:grounded} 
\end{figure}

Unlike task-oriented dialogue systems, most E2E conversation models are not grounded in the real world, which prevents these systems from effectively conversing about anything that relates to the user's environment. 
This limitation is also inherited from machine translation, which neither models nor needs are grounded. 
Recent approaches to neural response generation address this problem by grounding systems in the persona of the speaker or addressee \citep{li2016persona,alrfou2016conversational}, textual knowledge sources such as Foursquare \citep{ghazvininejad2017knowledge}, the user's or agent's visual environment \citep{das2017visual,mostafazadeh2017image}, and affect or emotion of the user \citep{huber2018emotional,winata2017nora,xu2018emo2vec}.  
At a high level, most of these works have in common the idea of augmenting their context encoder to not only represent the conversation history, but also some additional input drawn from the user's environment, such as an image \citep{das2017visual,mostafazadeh2017image} or textual information 
\citep{ghazvininejad2017knowledge}.

As an illustrative example of such grounded models, we give a brief overview of \citet{ghazvininejad2017knowledge}, whose underlying model is depicted in \figref{fig:grounded}.
The model mainly consists of two encoders and one decoder. The decoder and the dialogue encoder are similar to those of standard seq2seq models. The additional encoder is called the {\it facts encoder}, which infuses into the model factual information or so-called {\it facts} relevant to the conversation history, \eg, restaurant reviews (\eg, ``amazing sushi tasting'') that pertain to a restaurant that happened to be mentioned in the conversation history (\eg, ``Kusakabe''). 
While the model in this work was trained and evaluated with Foursquare reviews, this approach makes no specific assumption that the grounding consists of reviews, or that trigger words are restaurants (in fact, some of the trigger words are, \eg, hotels and museums).
To find facts that are relevant to the conversation, their system uses an IR system to retrieve text from a very large collection of facts or {\it world facts} (\eg, all Foursquare reviews of several large cities) using search words extracted from the conversation context. 
While the dialogue encoder of this model is a standard LSTM, the facts encoder is an instance of the Memory Network of \citet{chen2016end}, which uses an associative memory for modeling the facts relevant to a particular problem, which in this case is a restaurant mentioned in a conversation.

There are two main benefits to this approach and other similar work on grounded conversation modeling. 
First, the approach splits the input of the E2E system into two parts: the input from the user and the input from her environment. 
This separation is crucial because it addresses the limitation of earlier E2E (\eg, seq2seq) models which always respond deterministically to the same query (\eg to ``what's the weather forecast for tomorrow?''). 
By splitting input into two sources (user and environment), the system can effectively generate different responses to 
the same user input depending on what has changed in the real world, without having to retrain the entire system.
Second, this approach is much more sample efficient compared to a standard seq2seq approach. 
For an ungrounded system to produce a response like the one in \figref{fig:grounded}, the system would require that every entity any user might conceivably talk about  (\eg, ``Kusakabe'' restaurant) be seen in the {\it conversational} training data, which is an unrealistic and impractical assumption. While the amount of language modeling data (\ie, non-conversational data) is abundant and can be used to train grounded conversation systems (\eg, using Wikipedia, Foursquare), the amount of available conversational data is typically much more limited.
Grounded conversational models don't have that limitation, and, \eg, the system of \citet{ghazvininejad2017knowledge} can converse about venues that are not even mentioned in the conversational training data.

\section{Beyond Supervised Learning}
\label{sec:beyond-supervised-learning}

There is often a sharp disconnect between conversational training data (human-to-human) and envisioned online scenarios (human-computer).
This makes it difficult to optimize conversation models towards specific objectives, \eg, maximizing engagement by reducing blandness. 
Another limitation of the supervised learning setup 
is their tendency to optimize for an immediate reward (\ie, one response at a time) rather than a long-term reward. This also partially explains why their responses are often bland and thus fail to promote long-term user engagement.
To address these limitations, some researchers have explored reinforcement learning (RL) for E2E systems \citep{li2016deep} which could be augmented with human-in-the-loop architectures \citep{li2016dialogue,li2016learning}. 
Unlike RL for task-oriented dialogue, a main challenge that E2E systems are facing is the lack of well-defined metrics for success (\ie, reward functions), in part because they have to deal with informal genres such as chitchat, where the user goal is not explicitly specified.

\citet{li2016deep} constitutes the first attempt to use RL in a fully E2E approach to conversational response generation. 
Instead of training the system on human-to-human conversations as in the supervised setup of \citep{sordoni2015neural,vinyals2015neural}, the system of \citet{li2016deep} is trained by conversing with a user simulator which mimics human users' behaviors. 

\begin{figure}[t]
\centering 
\includegraphics[width=0.86\linewidth]{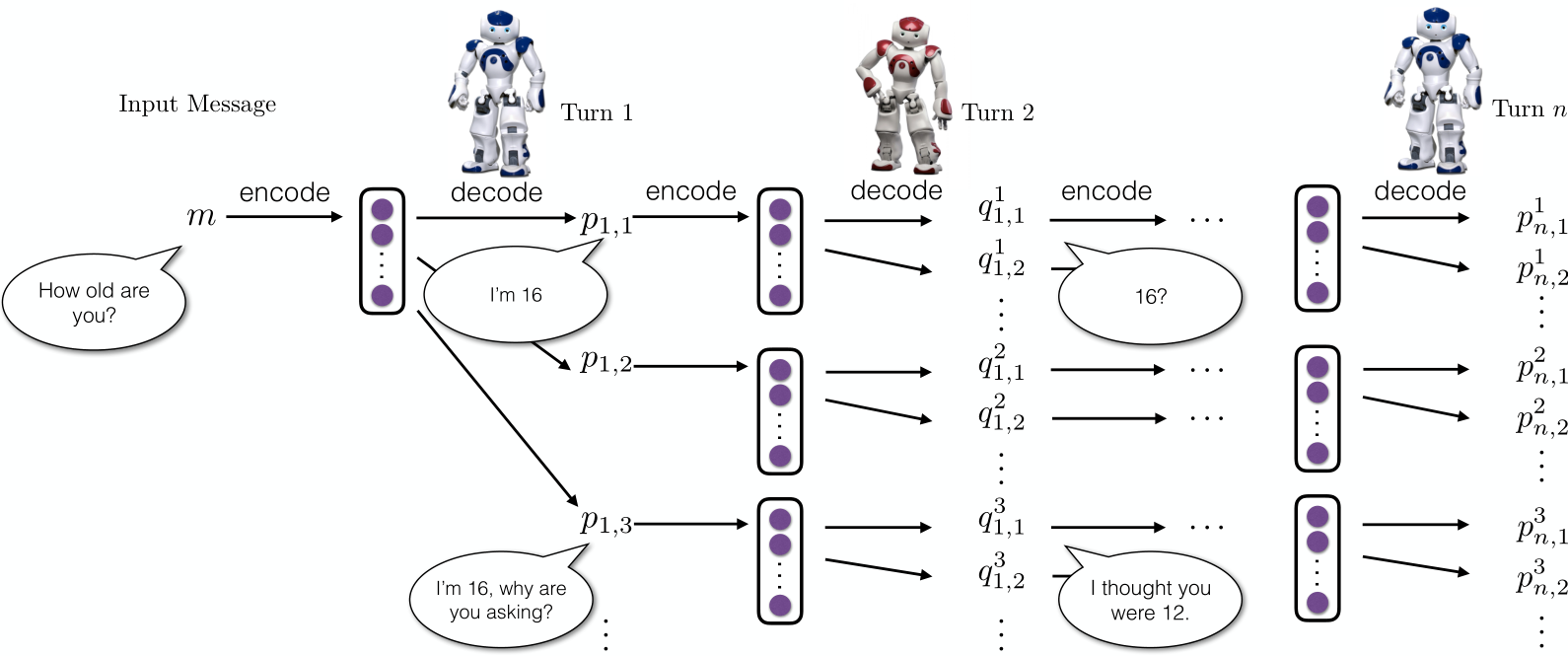}
\caption{Deep reinforcement learning for response generation, pitching the system to optimize against a user simulator (both systems are E2E generation systems.) Figure credit: \citet{li2016learning}}
\label{fig:rl} 
\end{figure}

As depicted in \figref{fig:rl}, human users have to be replaced with a user simulator because it is prohibitively expensive to train an RL system using thousands or tens of thousands of turns of real user dialogues. 
In this work, a standard seq2seq model is used as a user simulator.
The system is trained using policy gradient (\secref{sec:basics:rl}).
The objective is to maximize the expected total reward over the dialogues generated by the user simulator and the agent to be learned.
Formally, the objective is 
\begin{equation}
J(\theta) = \E[R(T_1,T_2,\ldots,T_N)]
\end{equation}
where $R(.)$ is the reward function, and $T_i$'s are dialogue turns.
The above objective can be optimized using gradient descent, by factoring the log probability of the conversation and the aggregated reward, which is independent of the model parameters:
\begin{equation}
\begin{split}
\nabla J(\theta) & = \nabla \log p(T_1,T_2,\ldots,T_N) R(T_1,T_2,...,T_N)\\
                 & \simeq \nabla \log \prod_i p(T_i|T_{i-1}) R(T_1,T_2,...,T_N)
\end{split}
\end{equation}
where $p(T_i|T_{i-1})$ is parameterized the same way as the standard seq2seq model of \secref{sec:lstm},
except that the model here is optimized using RL.
The above gradient is often approximated using sampling, and \citet{li2016deep} used a single sampled conversation for each parameter update.
While the above policy gradient setup is relatively common in RL, the main challenge in learning dialogue models is how to devise an effective reward function. 
\citet{li2016deep} used a combination of three reward functions that are designed to mitigate the problems of the supervised seq2seq model, which was used in their work 
as initialization parameters. 
The three reward functions are:

\begin{itemize}
\item $-p(\textnormal{Dull Response} | T_i)$: 
\citet{li2016deep} created a short list of dull responses such as ``I don't know'' selected from the training data. This reward function penalizes those turns $T_i$ that are likely to lead to any of these dull responses. 
This is called the {\it ease of answering} reward, as it promotes conversational turns that are not too difficult to respond to, so as to keep the user engaged in the conversation. For example, the reward function gives a very low reward to turns whose response is ``I don't know'', as this evasive response indicates that the previous turn was difficult to respond to, which may ultimately 
terminate the conversation.
\item $-\log \textnormal{Sigmoid} \cos(T_{i-1},T_{i})$: 
This {\it information flow} reward function ensures that consecutive turns $T_{i-1}$ and $T_{i}$ are not very similar to each other (\eg, ``how are you?'' followed by ``how are you?''), as \citet{li2016deep} assumed that conversations with little new information are often not engaging and therefore more likely to be terminated.
\item $\log p(T_{i-1} | T_{i}) + \log p(T_i | T_{i-1})$: 
This {\it meaningfulness} reward function was mostly introduced to counterbalance the aforementioned two rewards. For example, the two other reward functions prefer the type of conversations that constantly introduce new information 
and change topics so frequently that users find them hard to follow.
To avoid this, the meaningfulness reward encourages consecutive turns in a dialogue session to be related to each other.
\end{itemize}

\section{Data}
\label{sec:datasets}

\citet{serban2015survey} presented a comprehensive survey of existing datasets that are useful beyond the E2E and social bot research.
What distinguishes E2E conversation modeling from other NLP and dialogue tasks is that data is available in very large quantities, thanks in part to social media (\eg, Twitter and Reddit). 
On the other hand, most of this social media data is neither redistributable nor available through language resource organizations (such as the Linguistic Data Consortium), which means there are still no established public datasets (either with Twitter or Reddit) for training and testing response generation systems. 
Although these social media companies offer API access to enable researchers to download social media posts in relatively small quantities and then to re-construct conversations from them,
the strict legal terms of the service specified by these companies inevitably affect the reproducibility of the research. 
Most notably, Twitter makes certain tweets (\eg, retracted tweets or tweets from suspended user) unavailable through the API and requires that any such previously downloaded tweets be deleted. 
This makes it difficult to establish any standard training or test datasets, as these datasets deplete over time.\footnote{Anecdotally, the authors of \citet[pc]{li2015diversity} found that a Twitter dataset from 2013 had lost about 25\% of its tweets by 2015 due to retracted tweets and Twitter account suspensions.} 
Consequently, in most of the papers cited in this chapter, their authors have created their own (subsets of) conversational data for training and testing, and then evaluated their systems against baselines and competing systems on these fixed datasets. 
\citet{dodge2015evaluating} used an existing dataset to define standard training and test sets, but it is relatively small.
%
Some of the most notable E2E and chitchat datasets include:
\begin{itemize}
\item {\bf Twitter}: Used since the first data-driven response generation systems \citep{ritter2011data}, Twitter data offers a wealth of conversational data that is practically unbounded,
as Twitter produces new data each day that is more than most system developers can handle.\footnote{For example, the latest official statistics from Twitter, dating back from 2014, states that Twitter users post on average more than 500 million tweets per day: \url{https://blog.twitter.com/official/en_us/a/2014/the-2014-yearontwitter.html}} 
While the data itself is made accessible through the Twitter API as individual tweets, its metadata easily enables the construction of conversation histories, \eg, between two users.
This dataset forms the basis of the DSTC Task 2 competition in 2017 \citep{hori17end}.
\item {\bf Reddit}: Reddit is a social media source that is also practically unbounded, and represents about 3.2 billion dialogue turns as of July 2018. It was for example used in~\citet{alrfou2016conversational} to build a large-scale response retrieval system. Reddit data is organized by topics (\ie ``subreddits''), and its responses don't have a character limit as opposed to Twitter.
\item {\bf OpenSubtitles}: This dataset consists of subtitles made available on the opensubtitles.org website.  It offers captions of many commercial movies, and contains about 8 billion words as of 2011 in multiple languages~\citep{tiedemann12parallel}.
\item {\bf Ubuntu}: The Ubuntu dataset \citep{lowe2015ubuntu} has also been used extensively for E2E conversation modeling. It differs from other datasets such as Twitter in that it is less focused on chitchat but more goal-oriented, as it contains many dialogues that are specific to the Ubuntu operating system. 
\item {\bf Persona-Chat} dataset: This crowdsourced dataset \citep{zhang18perso} was developed to meet the need for conversational data where 
dialogues exhibit distinct user personas. 
In collecting Persona-Chat, every crowdworker was asked to impersonate a given character described using five facts. Then that worker took part in dialogues while trying to stay in character. The resulting dataset contains about 160k utterances.
\end{itemize}

\interfootnotelinepenalty=10000
\section{Evaluation}

Evaluation is a long-standing research topic for generation tasks such as machine translation and summarization. E2E dialogue is no different. While it is common to evaluate response generation systems using human raters \citep[etc.]{ritter2011data,sordoni2015neural,shang2015neural}, this type of evaluation is often expensive and researchers often have to resort to automatic metrics for quantifying day-to-day progress and for performing automatic system optimization. E2E dialogue research mostly borrowed those metrics from machine translation and summarization, using string and $n$-gram matching metrics like BLEU \citep{papineni2002bleu} and ROUGE \citep{lin2004rouge}. Proposed more recently, METEOR \citep{banerjee2005meteor} aims to improve BLEU by identifying synonyms and paraphrases between the system output and the human reference, and has also been used to evaluate dialogue. deltaBLEU \citep{galley2015deltableu} is an extension of BLEU that exploits numerical ratings associated with conversational responses.

There has been significant debate as to whether such automatic metrics are actually appropriate for evaluating conversational response generation systems. For example, \citet{liu2016how} argued that they are not appropriate by showing that most of these machine translation metrics correlate poorly with human judgments. However, their correlation analysis was performed at the sentence level,
but decent sentence-level correlation has long been known to be difficult to achieve even for machine translation 
\citep{ccb2009wmt,graham2015accurate},  the task for which the underlying metrics (\eg, BLEU and METEOR) were specifically intended.\footnote{For example, in the official report of the WMT shared task, \citet[Section 6.2]{ccb2009wmt} computed the percentage of times popular metrics are consistent with human ranking at the sentence level, but the results did not bode well for sentence-level studies: ``Many metrics failed to reach [a random] baseline (including most metrics in the out-of-English direction). This indicates that sentence-level evaluation of machine translation quality is very difficult.''}
In particular, BLEU \citep{papineni2002bleu} was designed from the outset to be used as a corpus-level rather than sentence-level metric, since assessments based on $n$-gram matches are brittle when computed on a single sentence. Indeed, the empirical study of \citet{koehn04stat} suggested that BLEU is not reliable on test sets consisting of fewer than 600 sentences. \citet{koehn04stat}'s study was on translation, a task that is arguably simpler than response generation, so the need to move beyond sentence-level correlation is probably even more critical in dialogue.
When measured at a corpus- or system-level, correlations are typically much higher than that at sentence-level \citep{metricsmatr2008}, e.g., with Spearman's $\rho$ above 0.95 for the best metrics on WMT translation tasks \citep{graham14testing}.\footnote{In one of the largest scale system-level correlation studies to date, \citet{graham14testing} found that BLEU is relatively competitive against most translation metrics proposed more recently, as they show there ``is currently insufficient evidence for a high proportion of metrics to conclude that they outperform BLEU''. Such a large scale study remains to be done for dialogue.} 
In the case of dialogue, \citet{galley2015deltableu} showed that the correlation of string-based metrics (BLEU and deltaBLEU) significantly increases with the units of measurement bigger than a sentence. Specifically, their Spearman's $\rho$ coefficient goes up from 0.1 (essentially no correlation) at sentence-level to nearly 0.5 when measuring correlation on corpora of 100 responses each.

Recently, \citet{lowe2017towards} proposed a machine-learned metric for E2E dialogue evaluation. They presented a variant of the VHRED model \citep{serban2017hierarchical} that takes context, user input, gold and system responses as input, and produces a qualitative score between 1 and~5. As VHRED is effective for modeling conversations, \citet{lowe2017towards} was able to achieve an impressive Spearman's $\rho$ correlation of 0.42 at the sentence level.
On the other hand, the fact that this metric is trainable leads to other potential problems such as overfitting and ``gaming of the metric''
\citep{albrecht07reex},\footnote{In discussing the potential pitfalls of machine-learned evaluation metrics, \citet{albrecht07reex} argued for example that it would be ``prudent to defend against the potential of a system gaming a subset of the features.'' In the case of deep learning, this gaming would be reminiscent of making non-random perturbations to an input to drastically change the network’s predictions, as it was done, e.g., with images in \citep{szegedy13intriging} to show how easily deep learning models can be fooled. However, preventing such a gaming is difficult if the machine-learned metric is to become a standard evaluation, and this would presumably require model parameters to be publicly available.}
which might explain why previously proposed machine-learned evaluation metrics \citep[etc.]{corston01ml,kulesza04learning,lita05blanc,albrecht07reex,gimenez08smor,pado09measuring,stanojevic14fitting} are not commonly used in official machine translation benchmarks.
The problem of ``gameable metrics'' is potentially serious, 
for example in the frequent cases where automatic evaluation metrics are used directly as training objectives \citep{och03mert,ranzato2015sequence} as unintended ``gaming'' may occur unbeknownst to the system developer. 
If a generation system is optimized directly on a trainable metric, then the system and the metric  become akin to an adversarial pair in GANs \citep{goodfellow2014generative}, where the only goal of the generation system (Generator) is to fool the metric (Discriminator). Arguably, such attempts become easier with trainable metrics as they typically incorporate thousands or millions of parameters, compared to a relatively parameterless metric like BLEU that is known to be fairly robust to such exploitation and was shown to be the best metric for direct optimization \citep{cer10best} among other established string-based metrics.
To prevent machine-learned metrics from being gamed,
one would need to iteratively train the Generator and Discriminator as in GANs, but most trainable metrics in the literature do not exploit this iterative process. Adversarial setups proposed for dialogue and related tasks \citep{kannan2016adversarial,li2017adversarial,holtzman18learning} offer solutions to this problem, but it is also well-known that such setups suffer from instability \citep{salimans06improved} due to the nature of GANs' minimax formulation.
This fragility is potentially troublesome as the outcome of an automatic evaluation should ideally be stable~\citep{cer10best} and reproducible over time, e.g., to track progress of E2E dialogue research over the years. All of this suggests that automatic evaluation for E2E dialogue is far from a solved problem.

\section{Open Benchmarks}

Open benchmarks have been the key to achieving progress in many AI tasks such as speech recognition, information retrieval, and machine translation. Although end-to-end conversational AI is a relatively nascent research problem, some open benchmarks have already been developed:

\begin{itemize}
\item {\bf Dialog System Technology Challenges (DSTC)}: In 2017, DSTC proposed for the first time an ``End-to-End Conversation Modeling'' track,\footnote{\url{http://workshop.colips.org/dstc6/call.html}} which requires systems to be fully data-driven using Twitter data.
Two of the tasks in the subsequent challenge (DSTC7) focus on grounded conversation scenarios. One is focused on audio-visual scene-aware dialogue and the other on response generation grounded in external knowledge (\eg, Foursquare and Wikipedia), 
with conversations extracted from Reddit.\footnote{\url{http://workshop.colips.org/dstc7/}}
\item {\bf ConvAI Competition}: This is a NIPS competition that has been featured so far at two conferences. It offers prizes in the form of Amazon Mechanical Turk funding. The competition aims at ``training and evaluating models for non-goal-oriented dialogue systems'', and in 2018 uses the Persona-Chat dataset~\citep{zhang18perso}, among other datasets.
\item {\bf NTCIR STC}: This benchmark focuses on conversation ``via short texts''. The first benchmark focused on retrieval-based methods, and in 2017 was expanded to evaluate generation-based approaches. 
\item {\bf Alexa Prize}: In 2017, Amazon organized an open competition on building ``social bots'' that can converse with humans on a range of current events and topics. 
The competition enables participants to test their systems with real users (Alexa users), and offers a form of indirect supervision as users are asked to rate each of their conversations with each of the Alexa Prize systems.
The inaugural prize featured 15 academic teams \citep{ram2018conversational}.\footnote{These 15 systems are described in the online proceeding: \url{https://developer.amazon.com/alexaprize/proceedings}}
\end{itemize}


\chapter{Conversational AI in Industry}
\label{sec:commercial}

This chapter pictures the landscape of conversational systems in industry, including task-oriented systems (\eg, personal assistants), QA systems, and chatbots.

\section{Question Answering Systems}

Search engine companies, including Google, Microsoft and Baidu, have incorporated multi-turn QA capabilities into their search engines to make user experience more conversational, which is particularly appealing for mobile devices. Since relatively little is publicly known about the internals of these systems (e.g., Google and Baidu), this section presents a few example commercial QA systems whose architectures have been at least partially described in public source, including Bing QA, Satori QA and customer support agents.

\begin{figure}[t] 
\centering 
\includegraphics[width=1.0\linewidth]{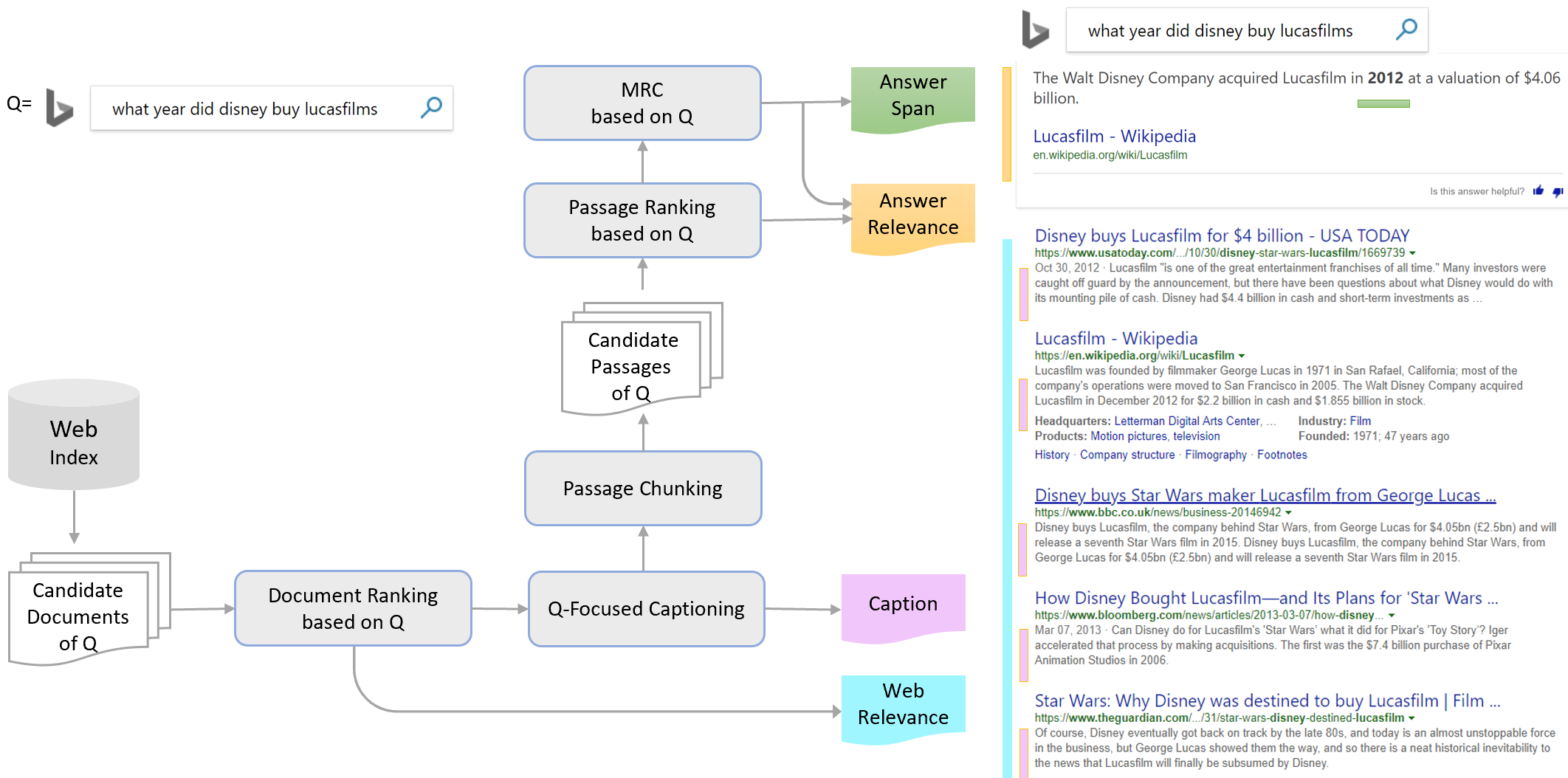}
\vspace{-2mm}
\caption{(Left) An overview of the Bing QA architecture. (Right) An example of a search engine result page of the question ``what year did disney buy lucasfilms?''. Example graciously provided by Rangan Majumder.} 
\label{fig:bing-qa}
\vspace{-2mm}
\end{figure}

\subsection{Bing QA}

Bing QA is an example of the Web-scale text-QA agents. It is an extension of the Microsoft Bing Web search engine. Instead of returning \emph{ten blue links}, Bing QA generates a direct answer to a user query by reading the passages retrieved by the Bing Web search engine using MRC models, as illustrated in \figref{fig:bing-qa}.

The Web QA task that Bing QA is dealing with is far more challenging than most of the academic MRC tasks described in Chapter \ref{sec:qa-bot}. For example, Web QA and SQuAD differs in:

\begin{itemize}
\item \textbf{Scale and quality of the text collection.} SQuAD assumes the answer is a text span in a passage which is a \emph{clean} text section from a Wikipedia page. Web QA needs to identify an answer from billions of Web documents which consist of trillions of \emph{noisy} passages that often contain contradictory, wrong, obsolete information due to the dynamic nature of Web content. 
\item \textbf{Runtime latency.} In an academic setting, an MRC model might take seconds to read and re-read documents to generate an answer, while in the Web QA setting the MRC part (\eg, in Bing QA) is required to add no more than 10 mini seconds to the entire serving stack.
\item \textbf{User experience.} While SQuAD MRC models provide a text span as an answer, Web QA needs to provide different user experiences depending on different devices where the answer is shown, \eg, a voice answer in a mobile device or a rich answer in a Search Engine Result Page (SERP).  \figref{fig:bing-qa} (Right) shows an example of the SERP for the question ``what year did Disney buy lucasfilms?'', where Bing QA presents not only the answer as a highlighted text span, but also various supporting evidence and related Web search results (i.e., captions of retrieved documents, passages, audios and videos) that are consistent with the answer.
\end{itemize}

As a result, a commercial Web QA agent such as Bing QA often incorporates a MRC module as a post-web component on top of its Web search engine stack. An overview of the Bing QA agent is illustrated in \figref{fig:bing-qa} (Left). Given the question ``what year did Disney buy lucasfilms?'', a set of candidate documents are retrieved from Web Index via a fast, primary ranker. Then in the Document Ranking module, a sophisticated document ranker based on boosted trees \citep{wu2010adapting} is used to assign relevance scores for these documents. The top-ranked relevant documents are presented in a SERP, with their captions generated from a Query-Focused Captioning module, as shown in \figref{fig:bing-qa} (Right).  The Passage Chunking module segments the top documents into a set of candidate passages, which are further ranked by the Passage Ranking module based on another passage-level boosted trees ranker \citep{wu2010adapting}. Finally, the MRC module identifies the answer span ``2012'' from the top-ranked passages.

\begin{figure}[t] 
\centering 
\includegraphics[width=0.68\linewidth]{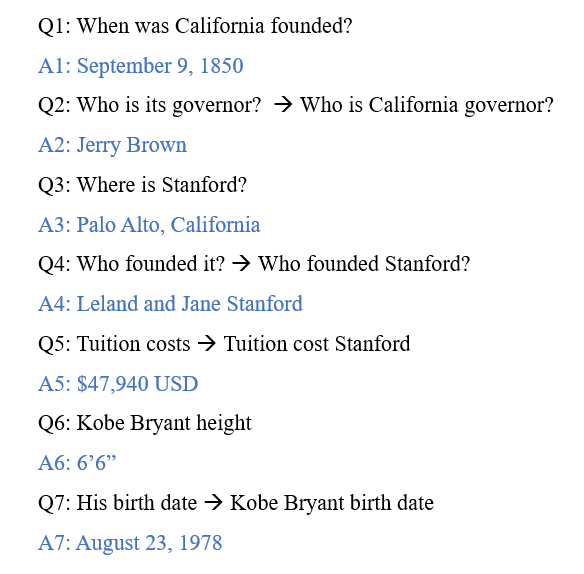}
\vspace{-2mm}
\caption{An example query session, where some queries are rewritten to include context information via the CQU module as indicated by the arrows. Examples adapted from \citet{ren2018conversational}.} 
\label{fig:cqu-example}
\vspace{-2mm}
\end{figure}

Although turning Bing QA into a conversational QA agent of \secref{sec:conversational-text-qa} requires the integration of additional components such as dialogue manager, which is a nontrivial ongoing engineering effort, Bing QA can already deal with conversational queries (\eg, follow up questions) using a Conversational Query Understanding (CQU) module \citep{ren2018conversational}. As the example in \figref{fig:cqu-example}, CQU reformulates a conversational query into a search engine friendly query in two steps: (1) determine whether a query depends upon the context in the same search session (\ie, previous queries and answers), and (2) if so, rewrite that query to include the necessary context \eg, replace ``its'' with ``California'' in Q2 and add ``Stanford'' in Q5 in \figref{fig:cqu-example}. 

\subsection{Satori QA}

Satori QA is an example of the KB-QA agents, as described in \secref{sec:knowledge-base}--\ref{sec:conversational-kbqa}. Satori is Microsoft's knowledge graph, which is seeded by Freebase, and now is several orders of magnitude larger than Freebase. Satori QA is a hybrid system that uses both neural approaches and symbolic approaches. It generates answers to factual questions. 

\begin{figure}[t] 
\centering 
\includegraphics[width=0.98\linewidth]{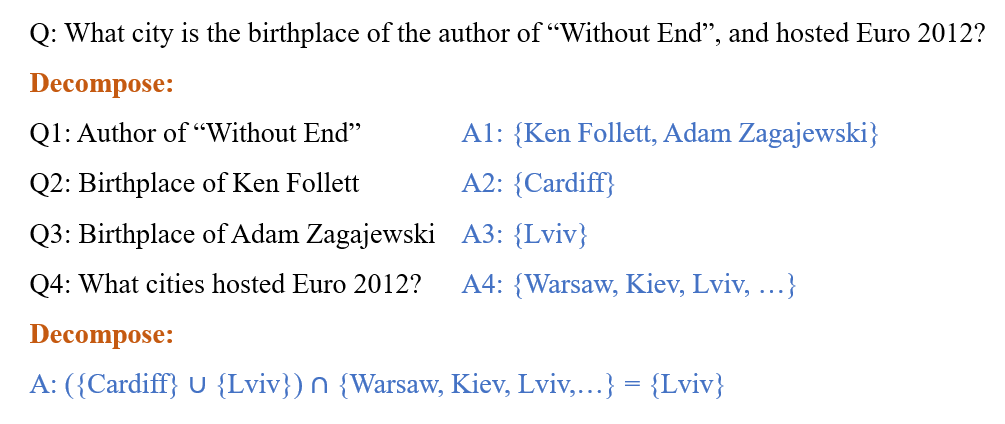}
\vspace{-2mm}
\caption{Given a complex question $Q$, we decompose it to a sequence of simple questions $Q_1, Q_2,...$, use a Web-scale KB-QA agent to generate for each $Q_i$ an answer $A_i$, from which we compute the final answer $A$. Figure credit: \citet{talmor2018web}.} 
\label{fig:satori-example}
\vspace{-5mm}
\end{figure}

Similar to Web QA, Satori QA has to deal with the issues regarding scalability, noisy content, speed, etc. One commonly used design strategy of improving system's robustness and runtime efficiency is to decompose a complex question into a sequence of simpler questions, which can be answered more easily by a Web-scale KB-QA system, and compute the final answer by recomposing the sequence of answers, as exemplified in \figref{fig:satori-example} \citep{talmor2018web}.

\subsection{Customer Support Agents}

Several IT companies, including Microsoft and Salesforce, have developed a variety of customer support agents. These agents are multi-turn conversational KB-QA agents, as described in 
\ref{sec:conversational-kbqa}. 
Given a user's description of a problem \eg, ``cannot update the personal information of my account'', the agent needs to recommend a pre-compiled solution or ask a human agent to help. The dialogue often consists of multiple turns as the agent asks the user to clarify the problem while navigating the knowledge base to find the solution. These agents often take both text and voice as input.


\section{Task-oriented Dialogue Systems (Virtual Assistants)}

Commercial task-oriented dialogue systems nowadays often reside in smart phones, smart speakers and personal computers.  They can perform a range of tasks or services for a user, and are sometimes referred to as virtual assistants or intelligent personal assistants.  Some of the example services are providing weather information, setting alarms, and calling center support.  In the US, the most widely used systems include Apple's Siri, Google Assistant, Amazon Alexa, and Microsoft Cortana, among others.  Users can interact with them naturally through voice, text or images.  To activate a virtual assistant using voice, a wake word might be used, such as ``OK Google.''

\begin{figure}[ht]
\centering
\includegraphics[width=\textwidth]{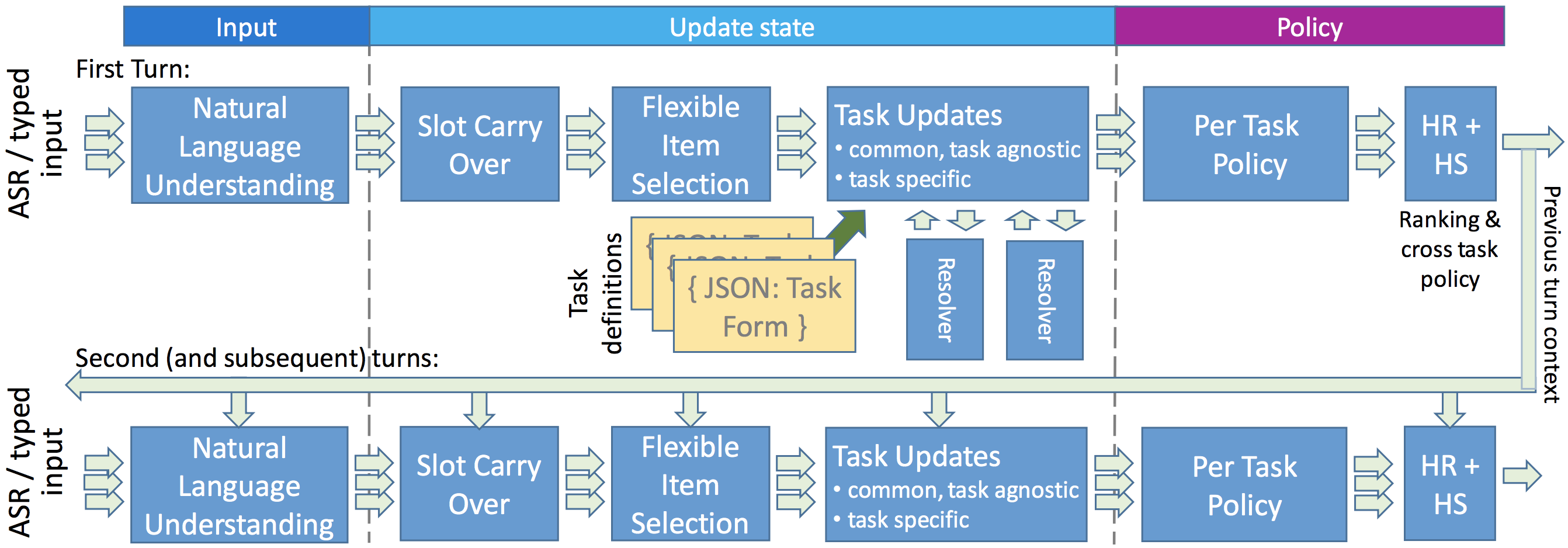}
\caption{Architecture of Task Completion Platform. Figure credit: \citet{crook16task}.} \label{fig:tcp}
\end{figure}

\begin{figure}
\centering
\includegraphics[width=\textwidth]{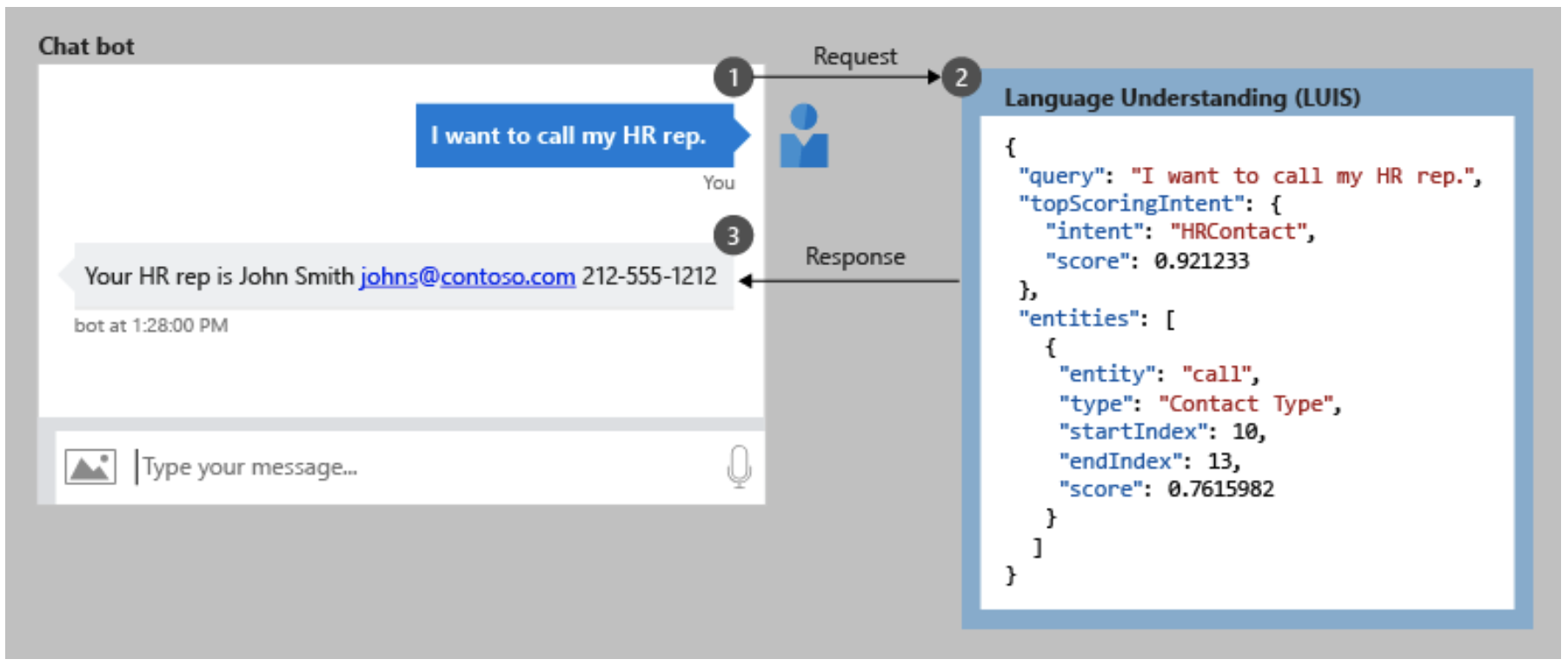}
\caption{Use of LUIS by a client dialogue system. Figure credit: \protect\url{https://docs.microsoft.com/en-us/azure/cognitive-services/LUIS}~.} \label{fig:luis}
\end{figure}

There are also a number of fast-growing tools available to facilitate the development of virtual assistants, including Amazon's Alexa Skills Kit\footnote{\url{https://developer.amazon.com/alexa-skills-kit}}, IBM's Watson Assistant\footnote{\url{https://ibm.biz/wcsdialog}}, and similar offerings from Microsoft and Google, among others.  A comprehensive survey is outside of the scope of this section, and not all information of such tools is publicly available.  Here, we will give a high-level description of a sample of them:
\begin{itemize}
\item{The Task Completion Platform (TCP) of Microsoft~\citep{crook16task} is a platform for creating multi-domain dialogue systems.  As shown in \figref{fig:tcp}, TCP follows a similar structure as in \figref{fig:dialogue-arch}, containing language understanding, state tracking, and a policy.  A useful feature of TCP is a task configuration language, TaskForm, which allows the definitions of individual tasks to be decoupled from the platform's overarching dialogue policy.  TCP is used to power many of the multi-turn dialogues
supported by the Cortana personal assistant.}
\item{Another tool from Microsoft is \textbf{LUIS}, a cloud-based API service for natural language understanding\footnote{\url{https://www.luis.ai}}.  It provides a suite of pre-built domains and intentions, as well as a convenient interface for a non-expert to use machine learning to obtain an NLU model by providing training examples.  Once a developer creates and publishes a LUIS app, the app can be used as a NLU blackbox module by a client dialogue system: the client sends a text utterance to the app, which will return language understanding results in the JSON format, as illustrated in \figref{fig:luis}.}
\item{While LUIS focuses on language understanding, the \textbf{Azure Bot Service}\footnote{\url{https://docs.microsoft.com/en-us/azure/bot-service/?view=azure-bot-service-3.0}} allows developers to build, test, deploy, and manage dialogue systems in one place.  It can take advantages of a suite of intelligent services, including LUIS, image captioning, speech-to-text capabilities, among others.}
\item{\textbf{DialogFlow} is Google's development suite for creating dialogue systems on websites, mobile and IoT devices.\footnote{\url{https://dialogflow.com}}  Similar to the above tools, it provides mechanisms to facilitate development of various modules of a dialogue system, including language understanding and carrying information over multiple turns.  Furthermore, it can deploy a dialogue system as an action that users can invoke through Google Assistant.}
\end{itemize}

%

\section{Chatbots}

There have been publicly-available conversational systems going back many decades \citep{Weizenbaum66eliza,colby75paranoia}.
Those precursors of today's chatbot systems relied heavily on hand-crafted rules, and are very different from the data-driven conversational AI systems discussed in \chref{sec:chitchat}. Nowadays publicly available and commercial chatbot systems are often a combination of statistical methods and hand-crafted components, where statistical methods provide robustness to conversational systems (\eg, via intent classifiers) while rule-based components are often still used in practice, \eg, to handle common chitchat queries (\eg, ``tell me a joke''). 
Examples include personal assistants like Amazon's Alexa, Google Assistant, Facebook M, and Microsoft's Cortana, which in addition to personal assistant skills are able to handle chitchat user inputs. 
Other commercial systems such as XiaoIce,\footnote{\url{https://www.msxiaobing.com}} 
Replika, \citep{fedorenko2017avoiding} 
Zo,\footnote{\url{https://www.zo.ai}} and  Ruuh\footnote{\url{https://www.facebook.com/Ruuh}} focus almost entirely on chitchat.
Since relatively little is publicly known about the internals of main commercial systems (Alexa, Google Assistant, etc.), the rest of this section focuses on commercial systems whose architecture have been at least partially described in some public source.

\begin{figure}[t]
\centering 
\includegraphics[width=1.0\linewidth]{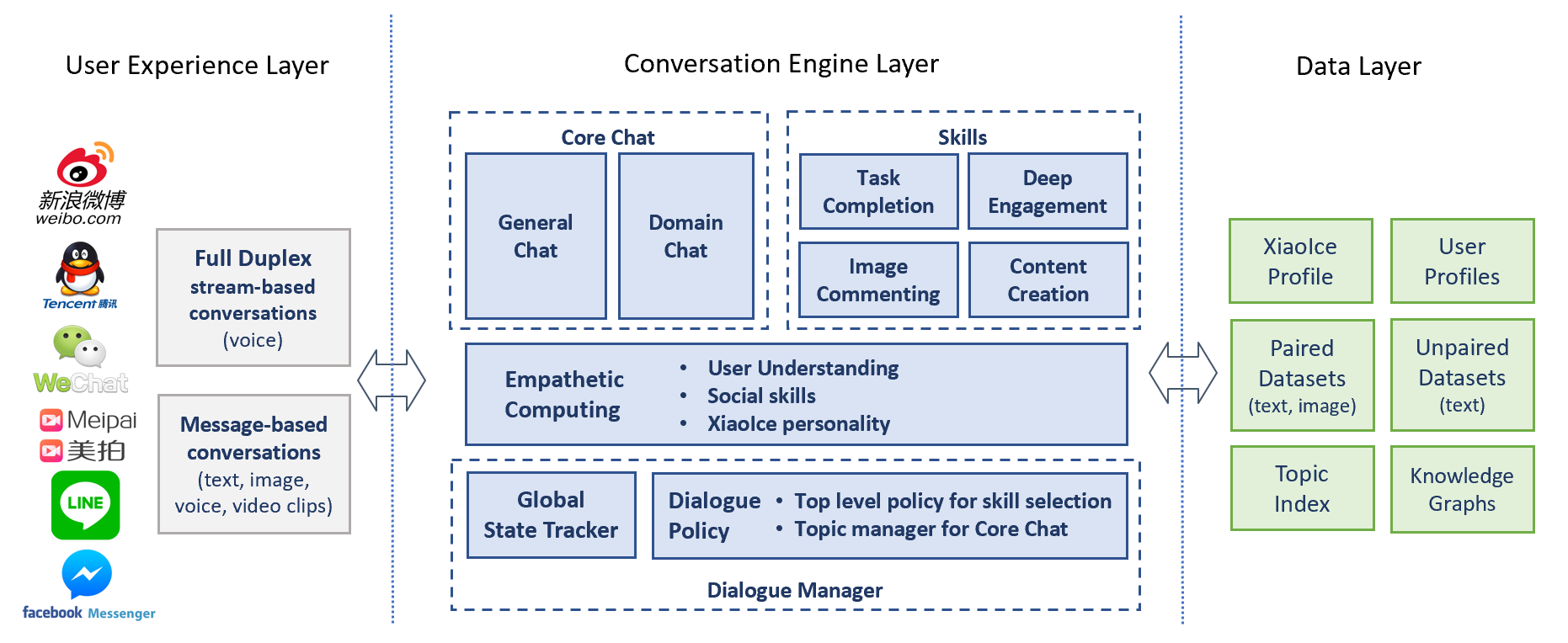}
\caption{XiaoIce system architecture. Figure credit:~\citet{zhou2018design}}
\label{fig:xiaoice-architecture} 
\end{figure}

One of the earliest such systems is {\bf XiaoIce}, which was initially released in 2014. XiaoIce is designed as an AI companion with an emotional connection to satisfy the human need for communication, affection, and social belonging~\citep{zhou2018design}. The overall architecture of XiaoIce is shown in \figref{fig:xiaoice-architecture}. It consists of three layers. 

\begin{figure}[t]
\centering 
\includegraphics[width=0.85\linewidth]{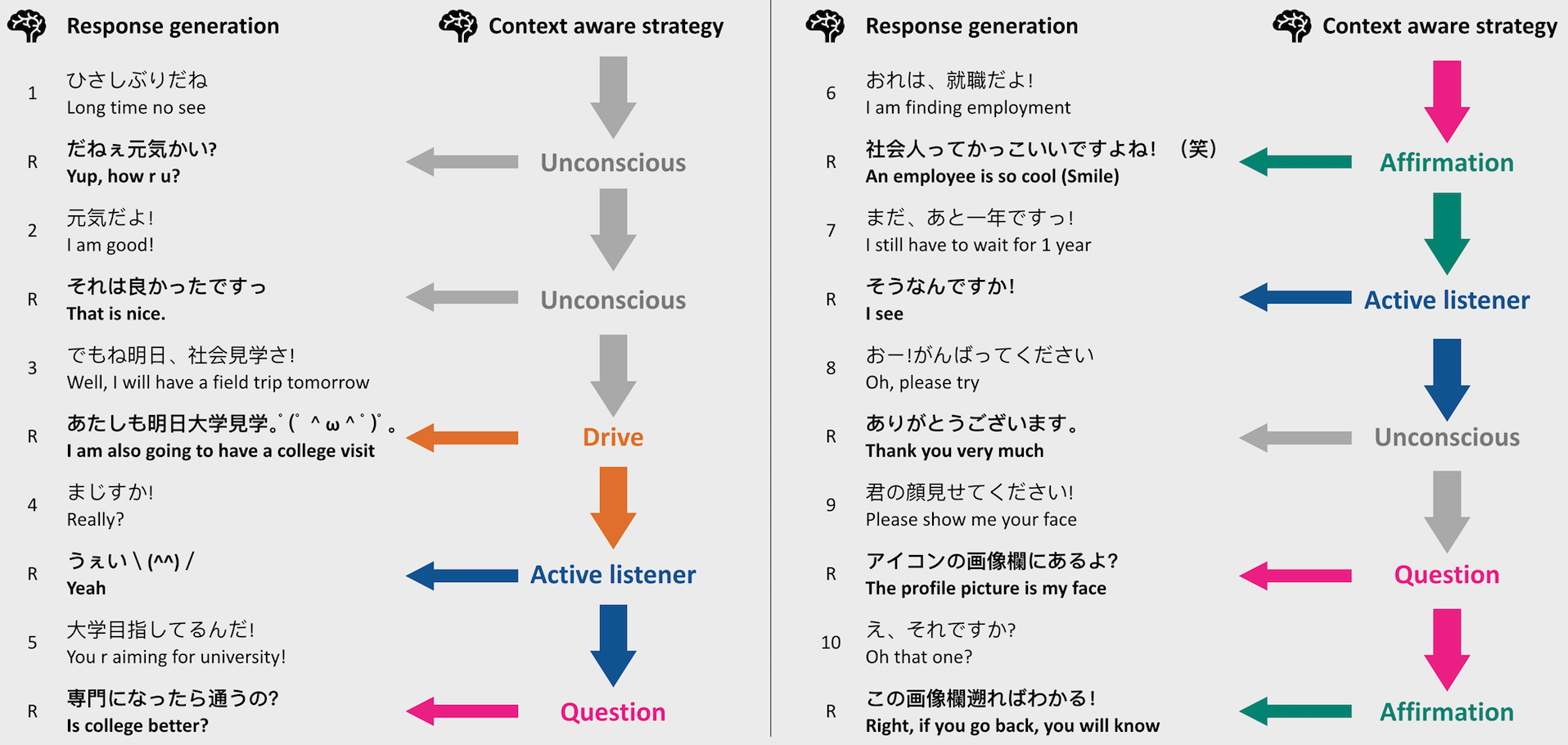}
\vspace{-2mm}
\caption{Conversation between a user and XiaoIce. The empathy model provides a context-aware strategy that can drive the conversation when needed.}
\label{fig:xiaoice} 
\end{figure}

\begin{itemize}
    \item \textbf{User experience layer:} It connects XiaoIce to popular chat platforms (e.g., WeChat, QQ), and deals with conversations in two communication modes. 
    The full-duplex module handles voice-stream-based conversations where both a user and XiaoIce can talk to each other simultaneously. 
    The other module deals with message-based conversations where a user and XiaoIce have to take turns to talk. 
    \item \textbf{Conversation engine layer:} In each dialogue turn, the dialogue state is first updated using the state tracker, and either Core Chat (and a topic) or a dialogue skill is selected by the dialogue policy to generate a response. 
    A unique component of XiaoIce is the empathetic computing module, designed to understand not only the content of the user input (e.g., topic) but also the empathy aspects 
    (e.g., emotion, intent, opinion on topic, and the user's background and general interests), to ensure the generation of an empathetic response that fits XiaoIce's persona.  
    Another central module, Core Chat, combines neural generation techniques (\secref{sec:e2econvo}) and retrieval-based methods~\citep{zhou2018design}. 
    As \figref{fig:xiaoice} show, XiaoIce is capable of generating socially attractive responses (e.g., having a sense of humor, comforting, etc.), and can determine whether to drive the conversation 
    when, \eg, the conversation is somewhat stalled, or whether to perform active listening when the user herself is engaged.\footnote{\url{https://www.leiphone.com/news/201807/rgyKfVsEUdK1BpXf.html}}
    \item \textbf{Data layer:} It consists of a set of databases that store the collected human conversational data (in text pairs or text-image pairs), non-conversational data and knowledge graphs used for Core Chat and skills, and the profiles of XiaoIce and all the registered users for empathetic computing.
\end{itemize}

The {\bf Replika} system \citep{fedorenko2017avoiding} for chitchat combines neural generation and retrieval-based methods, and is able to condition responses on images as in \citep{mostafazadeh2017image}. The neural generation component of Replika is persona-based \citep{li2016persona}, as it is trained to mimic specific characters. While Replika is a company, the Replika system has been open-sourced\footnote{\url{https://github.com/lukalabs/cakechat}} and can thus be used as a benchmark for future research. 

{\bf Alexa Prize systems} \citep{ram2018conversational} are social chatbots that are exposed to real users, and as such anyone with an Alexa device is able to interact with these social bots and give them ratings. This interaction is triggered with the ``Alexa, let's chat'' command, which then triggers a free-form conversation about any topic selected by either the user or the system. These systems featured not only fully data-driven approaches, but also more engineered and modularized approaches. For example, the winning system of the 2017 competition (Sounding Board\footnote{\url{https://sounding-board.github.io}}) contained a chitchat component as well as individual ``miniskills'' enabling the system to handle distinct tasks (\eg, QA) and  topics (\eg, news, sports). Due to the diversity of systems in the Alexa prize, 
it would be impractical to overview these systems in this survey, and instead we refer the interested reader to the Alexa Prize online proceedings \citep{ram2018conversational}.

\chapter{Conclusions and Research Trends}
\label{sec:conclusion}

Conversational AI is a rapidly growing field. This paper surveys neural approaches that were recently developed. Some of them have already been widely used in commercial systems.

\begin{itemize}
\item Dialogue systems for question answering, task completion, chitchat and recommendation etc. can be conceptualized using a unified mathematical framework of optimal decision process. The neural approaches to AI, developed in the last few years, leverage the recent breakthrough in RL and DL to significantly improve the performance of dialogue agents across a wide range of tasks and domains.
\item A number of commercial dialogue systems allow users to easily access various services and information via conversation. Most of these systems use hybrid approaches that combine the strength of symbolic methods and neural models.
\item There are two types of QA agents. KB-QA agents allow users to query large-scale knowledge bases via conversation without composing complicated SQL-like queries. Text-QA agents, equipped with neural MRC models, are becoming more popular than traditional search engines (e.g., Bing and Google) for the query types to which users expect a concise direct answer.
\item Traditional task-oriented systems use handcrafted dialogue manager modules, or shallow machine-learning models to optimize the modules separately. Recently, researchers have begun to explore DL and RL to optimize the system in a more holistic way with less domain knowledge, and to automate the optimization of systems in a changing environment such that they can efficiently adapt to different tasks, domains and user behaviors.
\item Chatbots are important in facilitating smooth and natural interaction between humans and their electronic devices. More recent work focuses on scenarios beyond chitchat, e.g., recommendation. Most state-of-the-art chatbots use fully data-driven and end-to-end generation of conversational responses within the framework of neural machine translation.
\end{itemize}


We have discussed some of the main challenges in conversational AI, common to Question Answering agents, task-oriented dialogue bots and chatbots.

\begin{itemize}
\item {\bf Towards a unified modeling framework for dialogues:}  Chapter 1 presents a unified view where an open-domain dialogue is formulated as an optimal decision process. Although the view provides a useful design principle, it remains to be proved the effectiveness of having a unified modeling framework for system development.  Microsoft XiaoIce, initially designed as a chitchat system based on a retrieval engine, has gradually incorporated many ML components and skills, including QA, task completion and recommendation, using a unified modeling framework based on empathic computing and RL, aiming to maximize user engagement in the long run, measured by expected conversation-turn per session. We plan to present the design and development of XiaoIce in a future publication. \citet{mccann2018natural} presented a platform effort of developing a unified model to handle various tasks including QA, dialogue and chitchat.  
\item {\bf Towards fully end-to-end dialogue systems:}
Recent work combines the benefit of task-oriented dialogue with more end-to-end capabilities. 
The grounded models discussed in \secref{sec:groundedconvo} represent a step towards more goal-oriented conversations, as the ability to interact with the user’s environment is a key requirement for most goal-oriented dialogue systems. Grounded conversation modeling discussed in this paper is still preliminary, and future challenges include enabling API calls in fully data-driven pipelines.

\item {\bf Dealing with heterogeneous data:}
Conversational data is often heterogeneous. For example, chitchat data is plentiful but not directly relevant to goal-oriented systems, and goal-oriented conversational datasets are typically very small. Future research will need to address the challenge of capitalizing on both, for example in a multi-task setup similar to \citet{luan2017multi}. Another research direction is the work of \citet{zhao2017generative}, which brought synergies between chitchat and task-oriented data using a ``data augmentation'' technique. Their resulting system is not only able to handle chitchat, but also more robust to goal-oriented dialogues. Another challenge is to better exploit non-conversational data (e.g., Wikipedia) as part of the training of conversational systems \citep{ghazvininejad2017knowledge}. 
\item {\bf Incorporating EQ (or empathy) into dialogue:} This is useful for both chatbots and QA bots. For example, XiaoIce incorporates an EQ module so as to deliver a more understandable response or recommendation (as in 3.1 of \citep{shum2018xiaoice}). \citet{fung2016towards} embedded an empathy module into a dialogue agent to recognize users' emotion using multimodality, and generate emotion-aware responses.
\item {\bf Scalable training for task-oriented dialogues:} It is important to fast update a dialogue agent to handle a changing environment. For example, \citet{lipton18bbq} proposed an efficient exploration method to tackle a domain extension setting, where new slots can be gradually introduced. \citet{chen2016end} proposed a zero-shot learning for unseen intents so that a dialogue agent trained on one domain can detect unseen intents in a new domain without manually labeled data and without retraining.
\item {\bf Commonsense knowledge} is crucial for any dialogue agents. This is challenging because common sense knowledge is often not explicitly stored in existing knowledge base. Some new datasets are developed to foster the research on common sense reasoning, such as Reading Comprehension with
Commonsense Reasoning Dataset (ReCoRD) \citep{zhang2018record}, Winograd Schema Challenge (WSC) \citep{Morgenstern2015WSC} and Choice Of Plausible Alternatives (COPA) \citep{roemmele2011choice}. 
\item {\bf Model interpretability:} In some cases, a dialogue agent is required not only to give a recommendation or an answer, but also provide explanations. This is very important in e.g., business scenarios, where a user cannot make a business decision without justification. 
\citet{shen2018reinforcewalk,xiong2017deeppath,das2017go} 
combine the interpretability of symbolic approaches and the robustness of neural approaches and develop an inference algorithm on KB that not only improves the accuracy in answering questions but also provides explanations why the answer is generated, \ie, the paths in the KB that leads to the answer node.
\end{itemize}


\bibliographystyle{apalike}
\bibliography{nacai}

\end{document}